\documentclass[
11pt, 
english, 
singlespacing, 
headsepline, 
]{MastersDoctoralThesis} 

\usepackage[utf8]{inputenc} 
\usepackage[T1]{fontenc} 
\usepackage{afterpage}

\usepackage{mathpazo} 
\usepackage{float}
\usepackage{hyperref}
\usepackage{graphicx}

\usepackage[
backend=biber,
style=authoryear-icomp,
maxcitenames =2,
backend = biber,
citestyle=authoryear-comp,
natbib=true,
sorting=nyvt,
doi=false,isbn=false,url=false
]
{biblatex}
\usepackage{comment}
\usepackage{comment}
\usepackage[autostyle=true]{csquotes} 
\thesistitle{Understanding and Analyzing Model Robustness and Knowledge-Transfer in Multilingual Neural Machine Translation using TX-Ray} 
\supervisorOne{Dr. Sharid Loáiciga}
\supervisorAnd{and}
\supervisorTwo{Nils Rethmeier}
\examiner{} 
\degree{Master of Science} 
\author{Vageesh Saxena} 
\addresses{} 

\subject{Biological Sciences} 
\university{\href{https://www.uni-potsdam.de/de/}{University of Potsdam}} 
\department{\href{https://www.uni-potsdam.de/de/}{Department of Computational Linguistics}} 
\group{\href{http://www.ling.uni-potsdam.de/cogsys/}{Cognitive Systems: Language, Learning, and Reasoning}} 
\newcommand{\facname}{Potsdam University}

\AtBeginDocument{
\hypersetup{pdftitle=\ttitle} 
\hypersetup{pdfauthor=\authorname} 
\hypersetup{pdfkeywords=\keywordnames} 
}

\begin{document}
\frontmatter 

\pagestyle{plain} 


\begin{titlepage}
\begin{center}

\vspace*{.06\textheight}
{\scshape\LARGE \univname\par}\vspace{1.5cm} 
\textsc{\Large Master's Thesis}\\[0.5cm] 

\HRule \\[0.4cm] 
{\huge \bfseries \ttitle\par}\vspace{0.4cm} 
\HRule \\[1.5cm] 
 
\begin{minipage}[t]{0.4\textwidth}
\begin{flushleft} \large
\emph{Author:}\\
\href{saxena@uni-potsdam.de}{\authorname} 
\end{flushleft}
\end{minipage}
\begin{minipage}[t]{0.4\textwidth}
\begin{flushright} \large
\emph{Supervisors:} \\
\href{sharid.indhira.loaiciga.sanchez@uni-potsdam.de}{\supnameOne} 
\href{}{\supnameAnd} 
\href{Nils.Rethmeier@dfki.de}{\supnameTwo} 
\end{flushright}
\end{minipage}\\[3cm]
 
\vfill

\large \textit{A thesis submitted in fulfillment of the requirements\\ for the degree of \degreename}\\[0.3cm] 
\textit{in the}\\[0.4cm]
\groupname\\\deptname\\[2cm] 
 
\vfill

{\large October 23, 2020}\\[4cm] 

\vfill
\end{center}
\end{titlepage}


\begin{declaration}
\addchaptertocentry{\authorshipname} 
\noindent I, \authorname, declare that this thesis titled, \enquote{\ttitle} and the work presented in it are my own. I confirm that:

\begin{itemize} 
\item This work was done wholly or mainly while in candidature for a research degree at this University.
\item Where any part of this thesis has previously been submitted for a degree or any other qualification at this University or any other institution, this has been clearly stated.
\item Where I have consulted the published work of others, this is always clearly attributed.
\item Where I have quoted from the work of others, the source is always given. With the exception of such quotations, this thesis is entirely my own work.
\item I have acknowledged all main sources of help.
\item Where the thesis is based on work done by myself jointly with others, I have made clear exactly what was done by others and what I have contributed myself.\\
\end{itemize}
 
\noindent Signed:\\
\rule[0.5em]{25em}{0.5pt} 
 
\noindent Date: 23/10/2020\\
\rule[0.5em]{25em}{0.5pt} 
\end{declaration}

\cleardoublepage


\vspace*{0.2\textheight}

\noindent\enquote{\itshape In science, you can say things that seem crazy, but in the long run, they can turn out to be right. We can get really good evidence, and in the end, the community will come around.}\bigbreak

\hfill Geoffrey Hinton


\begin{abstract}

\addchaptertocentry{\abstractname} 

Compared to the conventional phrase-based statistical machine translation techniques, neural networks have shown promising results in the field of Neural Machine Translation. However, despite their success on massive translation datasets, Multi-lingual Neural Machine Translation in an extremely-low resource setting has not been adequately explored. Through this research, we demonstrate how different languages leverage knowledge-transfer to improve Multi-lingual Neural Machine Translation in an extremely low-resource setting. Utilizing a minimal amount of parallel data to learn useful mappings between different languages, we use the Tatoeba translation challenge dataset from Helsinki NLP [\textbf{\cite{TiedemannThottingal:2020}}] to perform English-German, English-French, and English-Spanish translations. \newline

Unlike most of the conventional approaches that leverage heavy pre-training from source to target language-pair, we pre-train our model to perform English-English translations, where English is the source language for all our translations. We then finetune our pre-trained model to perform English-German, English-French, and English-Spanish translations, using joint multi-task and sequential transfer learning approaches. Through this research, we investigate questions such as :

\begin{itemize}
\item \textbf{RQ1:} \textbf{How can we effectively utilize knowledge-transfer within different languages to improve Multi-lingual Neural Machine Translation in an extremely low-resource language setting?} 

\item \textbf{RQ2:} \textbf{How does selectively pruning neuron-knowledge affects model's generalization and robustness? Additionally, how does pruning affects catastrophic forgetting in our Multi-lingual Neural Machine Translation system?} 

\item \textbf{RQ3:} \textbf{How can we effectively utilize TX-Ray [\cite{rethmeier2019txray}] to interpret, visualize, and quantify knowledge-abstractions and knowledge-transfer, for the trained models in RQ1-2?}
\end{itemize}

We choose BLEU-4 cumulative score as the metrics for evaluating the translation quality of the trained models \footnote{The goal of this research is to reflect on machine learning techniques like Multi-task and Sequential-transfer, aka Continual, learning. A considerable proportion of our work is focused on understanding the knowledge-transfer between various target languages and experimenting with several pruning-based methods. Throughout our research, we made no specific efforts in improving the scores or making the translation quality better than the currently available state-of-the-art networks.}. Our Sequential-transfer learning network outperforms our baselines by the end of our setup, on the extracted corpus of 40k parallel-sentences from the Tatoeba translation challenge dataset. However, pruning neuron-knowledge not only decreases the performance of our sequential-transfer learning setup but also doesn't suggest any improvement towards the system's robustness or generalization. Finally, we demonstrate that pruning neuron-knowledge in an extremely low-resource setting increases catastrophic forgetting in our Multi-lingual Neural Machine Translation system.
\end{abstract}

\begin{acknowledgements}
\addchaptertocentry{\acknowledgementname} 
I would first like to thank my thesis advisors, \textbf{Dr. Sharid Loáiciga} from the University of Potsdam, Germany and \textbf{Nils Rethmeier} from DFKI (German Research Centre for Artificial Intelligence), Berlin, Germany and University of Copenhagen, Denmark. The door to supervisors was always open whenever I ran into a trouble spot or had a question about my research or writing. They consistently allowed this paper to be my work but steered me in the right direction whenever they thought I needed it. \newline

I want to take this opportuning to thank my friend and batchmate \textbf{Jacob Löbkens}, who helped me with the German translation of the abstract. Additionally, I would also like to thank my friends who were involved and supported me with their views, suggestions, and validations for this research project: \textbf{Roshan Rane}, \textbf{Karthik Ajith Kumar}, \textbf{Goncalo Mordido}, and \textbf{Burak Özmen}. \newline

I would also like to acknowledge the Department of \textbf{Computational Linguistics and Cognitive Systems: Language, Learning, and Reasoning} at the University of Potsdam, Germany, for providing me such an excellent opportunity. Additionally, I would like to thank \textbf{DFKI, Berlin} for providing me with such a fantastic atmosphere and essential resources for completing this research.\newline

Finally, I must express my very profound gratitude to my parents, family, and my best friends \textbf{Ami Panchal} and \textbf{Saurabh Mathur} for providing me with unfailing support and continuous encouragement throughout my years of study and through the process of researching and writing this thesis. This accomplishment would not have been possible without them. Thank you. \newline

Vageesh Saxena

\end{acknowledgements}


\tableofcontents 

\listoffigures 

\listoftables 


\begin{abbreviations}{ll} 
\textbf{TX-Ray} & \textbf{T}ransfer e\textbf{X}plainability as p\textbf{R}eference of \textbf{A}ctivations anal\textbf{Y}sis\\
\textbf{ML} & \textbf{M}achine \textbf{L}earning\\
\textbf{AI} & \textbf{A}rtificial \textbf{I}ntelligence\\
\textbf{XAI} & e\textbf{X}plainable \textbf{A}rtificial \textbf{I}ntelligence\\
\textbf{NN} & \textbf{N}eural \textbf{N}etwork\\
\textbf{ANN} & \textbf{A}rtifical \textbf{N}eural \textbf{N}etwork\\
\textbf{DNN} & \textbf{D}eep \textbf{N}eural \textbf{N}etwork\\
\textbf{RNN} & \textbf{R}ecurrent \textbf{N}eural \textbf{N}etwork\\
\textbf{CNN} & \textbf{C}onvolutional \textbf{N}eural \textbf{N}etwork\\
\textbf{LSTM} & \textbf{L}ong \textbf{S}hort-\textbf{T}erm \textbf{M}emory\\
\textbf{GRU} & \textbf{G}ated \textbf{R}ecurrent \textbf{U}nit\\
\textbf{A-BGRU} & \textbf{A}ttention-based \textbf{B}idirectional \textbf{G}ated \textbf{R}ecurrent Units\\
\textbf{NMT} & \textbf{N}eural \textbf{M}achine \textbf{T}ranslation\\
\textbf{M-NMT} & \textbf{M}ultilingual \textbf{N}eural \textbf{M}achine \textbf{T}ranslation\\
\textbf{En} & \textbf{En}glish Language\\
\textbf{De} & German Language\\
\textbf{Fr} & \textbf{Fr}ench Language\\
\textbf{Es} & Spanish Language\\

\end{abbreviations}


\mainmatter 

\pagestyle{thesis} 


\pdfminorversion=4

\chapter{Introduction} 

\label{Chapter1} 



\section{Motivation}

Artificial neural networks are a series of interconnected neurons within multiple layers that mimics sensory processing of the brain [\textbf{\cite{10.5555/1207109}}]. This subfield of machine learning, based on training deep(multi-layered) artificial neural networks is called Deep Learning or Deep neural learning. These neural networks acquire their power and ability to learn from data through an algorithm called backpropagation. The role of this algorithm is to optimize the network output by iteratively adapting the weight for each neuron(aka hidden units), thereby optimizing the learning process for that batch of data. However, this performance comes at a considerable price. Due to backpropagation, neural networks are sensitive to new information, and they often overwrite the learned knowledge. In other words, when something new is learned, it wipes out the old information. Nevertheless, the real world consist of an infinite number of novel scenarios and training a model on such data is nearly impossible.
\newline

The ability of neural networks to transfer knowledge used on new scenarios is commonly known as transfer learning. Transfer Learning is a traditional approach where a previously trained neural network is reinitialized as the starting point for a new neural network that learns an additional secondary task. In other words, learning from the first model is used as pretrained-weights to the second model. The conventional supervised learning paradigm, on the other hand, breaks down if we do not have enough labelled data for the end-task or domain we care about, for a reliable model. Transfer learning enables us to deal with such scenarios by leveraging the already existing labelled data on any related task or domain. We try to transfer this knowledge from the source task in the source domain and apply it to our problem of interest in the target domain. Unlike classical machine learning setups, this transfer of knowledge from one domain to another through limited data is essential to semi-supervised learning under domain shift. 
\newline

While transfer learning aids in generalizing to the scenarios that are not encountered during training, it partially configures the model to prioritize on the second end-task. Inspired by the idea behind learning in the biological human brain, lifelong learning is a methodology focused on the ability of cortical and subcortical circuits to add new information on the fly. Multi-task training is one of the recognized techniques that allows us to train neural networks across many tasks. However, this is only possible if all tasks are available at the time of primary training. In other words, for each new task, we would generally need to retrain our model on all the tasks again. In the real world, however, we would like a model that can gradually leverage from its experience. For that reason, we need to facilitate a model that can learn continuously without any drastic forgetting. This field of machine learning is known as continual learning.
\newline

The increment in the number of parameters and tasks not only increases the computational complexity of the model but also causes the problem of catastrophic forgetting [\textbf{\cite{nguyen2019understanding}}]. Despite the demonstrated promises, optimization for these methodologies has been a challenge. In a transfer or continual learning setup, it is often seen that not all neurons are effectively utilized during training. \newline

A neural network performs inference by representing knowledge on multiple feature levels. However, despite its simple anatomy, it doesn't have the crafted structure of sparse connectivity like a brain. Our understanding of these neural networks is in terms of probability densities and ensemble distributions. Like mentioned in [\textbf{\cite{rethmeier2019txray}}], while state-of-the-art(SOTA) explainability (XAI) methods in NLP, concentrate on supervised, per-instance or probing task evaluation [\textbf{\cite{belinkov-glass-2019-analysis, arras-etal-2017-explaining, CSI19}}], it is inadequate to interpret and quantify knowledge transfer during (un-) supervised training. Through our method, we express knowledge-representations as the neuron-feature(token) distribution over many instances. Such a technique helps us to interpret and quantify knowledge-transfer amongst different languages in our continual-learning based multi-lingual neural machine translation setup. Nonetheless, such an analysis can find:
\begin{itemize}
    \item Task-relevant neuron knowledge; that appears to overfit for a specific end-task.
    \item Task-irrelevant neuron knowledge; that reduces overfitting, which helps to describe components that generalize or specialize knowledge-transfer.
    \item Noise sensitive neuron knowledge; that is prone to the input noise, which helps with better generalization, making the model more robust.
\end{itemize}

Inspired by [\textbf{\cite{rethmeier2019txray}}], we express knowledge representations as an interpretable token-activation distribution. In this work, we demonstrate the effects of selective pruning 
on model performance in sequential-transfer learning/continual setup. 
Furthermore, we also provide a visual exploration to analyze neuron-knowledge change, expressing transfer of knowledge between other languages, for the task of multi-lingual neural machine translation.

\section{Research Questions}
Through this work, we investigate on questions such as:

\begin{itemize}

\item \textbf{RQ1:} \textbf{How can we effectively utilize knowledge-transfer within different languages to improve Multilingual Neural Machine Translation, in an extremely low-resource language setting?} Through this research question, we demonstrate how Multilingual Neural Machine Translation leverage knowledge-transfer within different languages (like English, German, French, and Spanish), in an extremely low-resource setting. Additionally, we also investigate how languages with similar roots affect this multilingual-training process. 


\item \textbf{RQ2:} \textbf{How does selectively pruning neuron-knowledge affects model's generalization and robustness? Additionally, how does pruning affects catastrophic forgetting in neural networks?}; Following the research from our original paper [\textbf{\cite{rethmeier2019txray}}], we extend the scope of TX-Ray by examining the effects of selective pruning neuron-knowledge on multilingual translation quality. In this experiment, we selectively prune neuron-knowledge from the max, most-n, and least-n active neurons and examine their effects on Catastrophic forgetting. 

\item \textbf{RQ3:} \textbf{How can we interpret, visualize, and quantify knowledge-abstractions for the trained models in RQ1-2?} Through this research question, we :
\begin{itemize}
    \item \textbf{(A)}: Visualize knowledge-abstractions to understand the evolution of positive and negative knowledge while training, for different translation pairs (namely English-English, English-German, English-French, and English-Spanish). Additionally, we also examine the effects of transfer and pruning on the knowledge-abstractions and translation quality.
    \item \textbf{(B)}: Employ TX-Ray [\textbf{\cite{rethmeier2019txray}}]; an interpretable framework that utilizes visualizations as a means to express knowledge-transfer and different linguistic features learnt by the trained models.
\end{itemize}
\end{itemize}
\pdfminorversion=4

\chapter{Related Research} 

\label{Chapter2} 



\section{Deep Learning and Neural Machine Translation}
\begin{figure}[h]
\centering
\includegraphics[width=\textwidth, keepaspectratio]{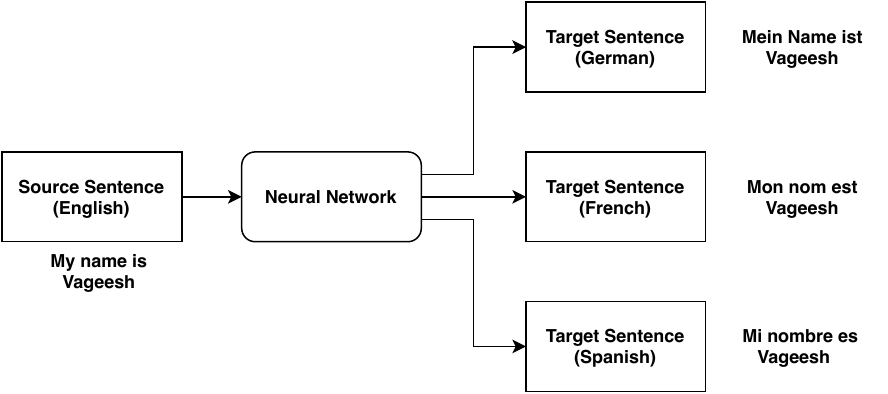}
\caption{Multilingual Neural Machine Translation; Source language, English is translated to multiple Target languages like German, French, and Spanish.}
\label{fig:machine_translation}
\end{figure}

Over the years, Deep Neural Networks (DNNs) have achieved exceptional accomplishment by learning numerous intricate tasks in the fields of Computer Vision and Natural Language Processing(NLP). In this work, we utilize the learning capabilities of such state-of-the-art networks to perform the task of Machine Translation in an extremely low-resource setting, aka Neural Machine Translation(NMT). Neural Machine Translation is a subfield of computational linguistics focused on translating one language to another, through the means of a Neural Network. During the training, a sequence of words(tokens) is converted from the source language to one or many target languages. In case of latter, such a task is widely known as Multilingual Neural Machine Translation (Figure \ref{fig:machine_translation}).\footnote{For demonstration purposes, the translations are taken from DeepL, a free online neural machine translation service.}\newline

Neural Machine Translation operates on sequential data over several time-steps. Translating one language to another language consists of a sequence of words of variable length, both from source language as well as the target language. However, deep learning architectures like Feed-Forward Neural Networks, aka Multilayer perceptron networks(MLP) and Convolutional Neural Networks(CNNs) need a fixed input vector(size) to produce a sized fix output.  In general, practices like padding and pad-packing are widely known in the field of NLP to tackle such problems. However, as demonstrated by [\textbf{\cite{Dwarampudi2019EffectsOP}}], neither pre-padding nor post-padding, help these networks as much as they do for Recurrent Neural networks(RNNs). Not only do RNNs have loops(memory units) to persist such sequential information, but they also recurrently perform the same task for every element in the sequence. Furthermore, the output elements in RNNs are dependent on previous elements or states. Therefore, RNNs can have one or more inputs as well as one or more outputs. \newline

With the overwhelming results obtained by [\textbf{\cite{johnson-etal-2017-googles}}] and [\textbf{\cite{aharoni2019massively}}] on massive data, Multilingual Neural Machine Translation(MNMT) for the low-resource language setting has become a trending research area [\textbf{\cite{gu2018universal}} and \textbf{\cite{lakew2019multilingual}}]. Improving translations in Multilingual Neural Machine Translation setup, for an extremely low-resource language setting is nevertheless even more challenging. In practice, neural methods require a massive amount of parallel data to learn effectively between languages. Since Multilingual Neural Machine Translation encodes word representations from multiple languages in a shared semantic space, it allows the network to utilize the knowledge-transfer between different languages [\textbf{\cite{lakew2019multilingual}}] positively. Through this work, we intend to visualize, interpret and quantify this transfer of knowledge between different languages like English, German, French, and Spanish in an extremely-low resource language setting (\autoref{Chapter3}). \newline

\section{Multi-task and Continual Learning}
Multitask training techniques [\textbf{\cite{johnson2017google}}] have shown promising results in the field of Multilingual Neural Machine Translation. The de-facto approach here is to not only share the input vocabulary across all the tasks but also introduce an artificial token at the beginning of the input sentence to identify the target language. The intention here is to enable the knowledge-transfer between all the target languages while training on parallel batches. Whereas in transfer learning, the standard approach is to pre-train a sequence encoder and fine-tune it to a set of end-tasks. This fine-tuning is performed either by freezing the weights or by directly fine-tuning the pretrained model [\textbf{\cite{rethmeier2019txray}}]. Following the work of Sebastian Ruder in [\textbf{\cite{ruder2019neural}}], we design a neural machine translation system that uses sequential-transfer learning to perform Multilingual Neural Machine Translation. Sequential transfer learning is a setting where the training between the source and target tasks is committed sequentially. Therefore, unlike multitask learning, instead of optimizing learning jointly, it is optimized sequentially. The goal of sequential-transfer learning, as part of continual learning, is to transfer information from the pre-trained model to several domains, in a sequential fashion. In other words, a sequential-transfer setup is just a multi-step transfer learning setup. \newline 

\begin{figure}[h]
\centering
\includegraphics[width=\textwidth, keepaspectratio]{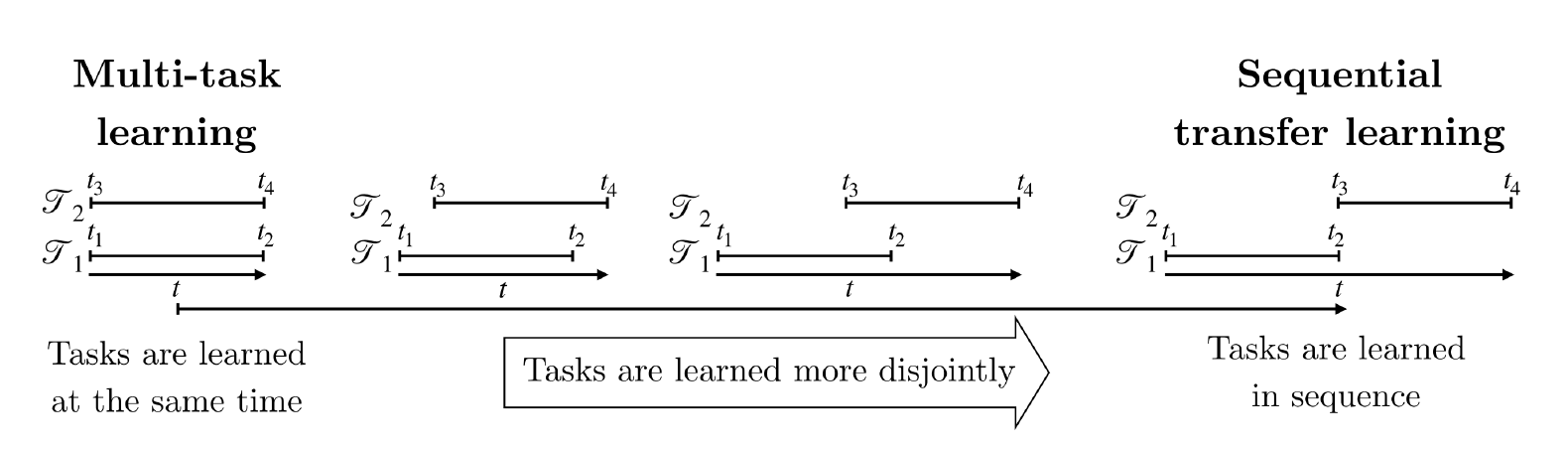}
\caption{Difference between Sequential-transfer learning and multi-task learning [\textbf{\cite{ruder2019neural}}].}
\label{fig:sequential-learning}
\end{figure}

Despite the differences, sequential transfer learning (or continual learning) and Multi-task learning are closely related (Figure  \ref{fig:sequential-learning}). They both utilize the transfer of knowledge from one domain to improve the performance in others. However, based on the research [\textbf{\cite{ruder2019neural}}], sequential-transfer learning-based approaches are helpful in scenarios where:
\newpage
\begin{itemize}
    \item Training data for all the tasks is not available at the get-go.
    \item The source task contains much more information that the target task.
    \item The transfer of knowledge is beneficial for multiple end-tasks (like M-NMT).
\end{itemize}

\section{TX-Ray : Transfer eXplainability as pReference of Activations analYsis}
Despite the tremendous success of sequential-transfer learning in natural language processing, neural-learning remains a deep mystery. The networks internal working is shielded from human eyes, concealed in layers of computations, making it hard to diagnose errors, activations, and biases. Understanding them will not only provide transparency in deep learning but will also help in further improving the performance. In this research, we attempt to interpret the transfer of knowledge through various visualizations, from one domain to another in the setting of Multilingual Neural Machine Translation. However, To accomplish that, there is a need to construct deep abstractions and instantiate them in rich interfaces. As mentioned in [\textbf{\cite{rethmeier2019txray}}], methods such as [\textbf{\cite{carter2019activation}} and \textbf{\cite{journals/corr/YosinskiCNFL15}}] focus on decision-understanding [\textbf{\cite{CSI19}}] by analyzing supervised probing tasks [\textbf{\cite{belinkov-glass-2019-analysis}}] using either performance metrics [\textbf{\cite{Senteval,Glue}}] or laborious per-instance explainability [\textbf{\cite{arrasACL19,2019-errudite}}]. These approaches analyze input-output relations for decision-understanding, whereas methods [\textbf{\cite{CSI19}}] like [\textbf{\cite{DeepEyes18, carter2019activation}}] visualize interpretable model abstractions learned via supervision. \newline 

\begin{figure}
\centering
\includegraphics[width=\textwidth, keepaspectratio]{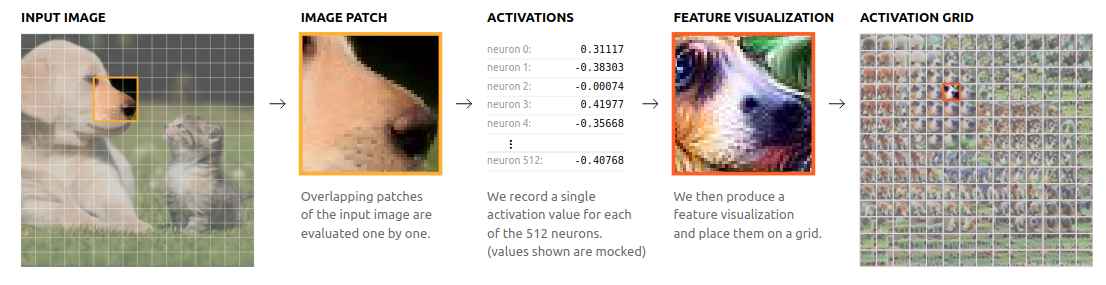}
\caption{Activation Atlas: Grid of feature visualizations(one for each patch) showing how the network sees and understand different parts of the input image. \cite{carter2019activation}}
\label{fig:activation-atlas}
\end{figure}

With the tremendous success of transfer learning, NLP relies heavily on pretrained data. The field currently lacks methods that allow one to explore, analyse, and quantify knowledge transfer mechanisms for pretraining or fine-tuning stages. Inspired by [\textbf{\cite{olah2018the}}], that uses techniques like feature visualisation, attribution, and dimensionality reduction, we use TX-Ray [\textbf{\cite{rethmeier2019txray}}]; an interpretability framework that evaluates the knowledge transfer in a transfer-learning setup. Our framework not only helps us visualise what the network learns but also explain how it developed its understanding while keeping the amount of information to human-scale. From analysing recent model and explainability methods [\textbf{\cite{CSI19,  belinkov-glass-2019-analysis, ExplainingExplanations}}], two kinds of approaches emerge: supervised model-understanding (MU) and decision-understanding (DU). While DU visualises interactions between the inputs and outputs space to understand model decisions, MU visualises model abstraction to understand the learned knowledge. Since both DU and MU heavily focus supervision analysis, understanding transfer or continual learning in un- and supervised models remain open challenges. Supervised-DU techniques use probing tasks to understand language properties like syntax and semantics [\textbf{\cite{Senteval, Glue, Decathlon, DiagnosticClassifiers}}]. Nonetheless, DU also uses per-sample supervised analysis [\textbf{\cite{arrasACL19,ExplainingExplanations}}] for model decision [\textbf{\cite{belinkov-glass-2019-analysis}}] by highlighting prediction-based decision i.e. relevant input words per instance [\textbf{\cite{arras-etal-2017-explaining}}]. While model understanding-MU techniques like Activation Atlas(Figure \ref{fig:activation-atlas}), or summit [\textbf{\cite{carter2019activation, olah2018the, hohman2019summit}}] enable exploration of supervised model knowledge in vision, NLP methods like [\textbf{\cite{SEQ2SEQVIS, RetainVis}}] compare models using many per-sample explanations. These methods produce a high cognitive load, showing many details, which makes it harder to understand overarching learning phenomena. \newline

\begin{figure}[h]
\centering
\includegraphics[width=\textwidth, keepaspectratio]{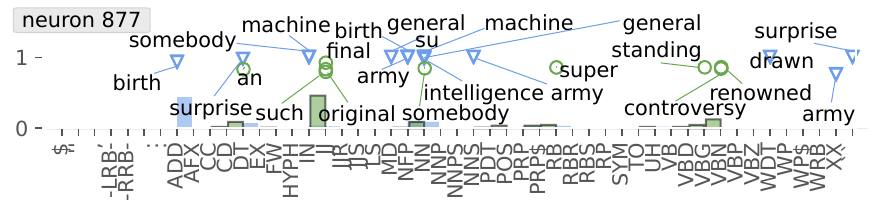}
\caption{TX-Ray: Neuron: 877, demonstrating transfer of knowledge from the zero-shot(blue) task to the supervised(green) task. \cite{rethmeier2019txray}}
\label{fig:tx-ray}
\end{figure}

Following the inspiration from [\textbf{\cite{carter2019activation}}], we designed TX-Ray; an interpretable framework, that visualizes different linguistic features learnt by Deep Neural Networks, trained on text data [\textbf{\cite{rethmeier2019txray}}]. During our research, we created a platform to implement visualizations as a means of understanding the transfer of knowledge from the pre-trained/zero-shot task to the supervised task(Figure \ref{fig:tx-ray}). To make it accessible, we visualized this knowledge between the shared neurons for the two tasks, by collecting the max-activations for every input-sample in test data. Through our work, we show how self-supervision training helps in learning linguistic features like discrete word function (like parts-of-speech (POS)), without teaching it to our model explicitly. We then used Hellinger distance as a measure to quantify the knowledge-transfer between these shared neurons. Later in the experiments, we also utilized various pruning based approaches to support our interpretations by showing how pruning different parts of our network help with model generalization and robustness.\newline

In this work, we extend the capabilities of our framework to explore the transfer of knowledge between various target languages, for a Multilingual Neural Machine Translation setup trained in both Multi-task, and Sequential-transfer, aka continual, learning setting. Similar to the experiments performed in our original works [\textbf{\cite{rethmeier2019txray}}], we also show the effects of various pruning based approaches on the model performance, in terms of generalization and robustness.

\section{Catastrophic forgetting in Continual Learning}
\label{sec:catastrophic_forgetting}
The ability to learn in a sequential-transfer learning fashion is critical in natural language processing. Not only it helps to solve the problem of limited labelled data but the ability to transfer knowledge from previous experience also enables the network to learn and generalize better. While continual learning shows the promising results towards model generalization in natural language processing, it suffers from the problem of catastrophic forgetting. Forgetting in neural networks or humans is defined as the inability to extract information from memory earlier. Although forgetting in humans comes in many forms, the process is normal, adaptive, and a crucial aspect of the learning. Neurons in neural networks allow a similar mechanism of forgetting information while trained continually, to solve specialized end-tasks. While this helps in improving the performance of a specific end-task, is catastrophic from the generalized learning in such systems. Catastrophic Forgetting in Continual learning is a phenomenon due to which the network forgets crucial information while updating its memory to improve the performance on a specific task [\textbf{\cite{nguyen2019understanding}}]. \newline

Several approaches like Elastic Weight Consolidation(EWC), Incremental Moment Matching (IMM), and Neural Pruning (CLNP)  [\textbf{\cite{kirkpatrick2017overcoming, lee2017overcoming, golkar2019continual}}] have shown promising results towards improving training in continual learning setup. These approaches focus on methods that enable selective freezing of the weights of the network to minimize catastrophic forgetting. However, to keep things simple, we choose to manually freeze the weights of our pre-trained network. Similar to the experiments performed by [\textbf{\cite{nguyen2020dissecting}}], we aim to improve transfer in the neural network through the means of interpretability techniques and visualizations, we implement our continual learning system by simply freezing the weights of our pre-trained network at every step [\textbf{\cite{soseliafreezing}}]. Furthermore, in this work, we experiment with various approaches by pruning specialized, generalized, and noise-sensitive knowledge to show its effects on model performance. \newline

\section{BLEU-4 as the Evaluation Metric}
Inspired by the research in [\textbf{\cite{papineni2002bleu}}], we choose Bilingual Evaluation Understudy Score, or in short BLEU, as our evaluation metrics. The metric calculates matching n-grams between the source and target sequences, where n represents the number of tokens/words compared in pairs, regardless of their order. In this work, we propose using BLEU-4 cumulative score that calculates the individual n-grams, weighted in all orders between 1 to n via geometric mean. In BLEU-4, the weights are uniformly distributed, i.e. 1/4 or 25\%, between each of the 1-gram, 2-gram, 3-gram and 4-gram scores. \newline

Even though BLEU-4 score is not perfect, it is widely recognized as an easy, inexpensive, and language-independent technique for evaluating predictions in neural machine translation. An ideal BLEU-4 score lies between 0.0 and 1.0, where 1.0 represents a perfect match and 0.0 represents perfect mismatch.

\section{Network Architectures}
\label{sec:architectures}
\subsection{Sequence-to-Sequence learning}
Over the years, sequence to sequence (seq2seq) models has shown promising results in the field of Neural Machine Translation [\textbf{\cite{sutskever2014sequence}}]. Usually, seq2seq models are encoder-decoder models, that use a recurrent neural network (RNN)-based architecture to encode a source(input) sentence into a fixed-length vector otherwise known as context vector. This abstract representation of a given input sentence is decoded by a second RNN to produce the target(output) sentence, by generating one word(token) at a time. \newline

Exploiting pre-trained Transformers(like BERT) have recently gained their popularity in the field of Natural Language Processing. Various researchers [\textbf{\cite{clinchant2019use}}, \cite{zhu2020incorporating}, and \textbf{\textbf{\cite{klein2017opennmt}}}] have demonstrated extraordinary results by fine-tuning such pre-trained models for the task of Neural Machine Translation. However, Since one of our focus is in training a Multilingual Neural Machine Translation system in an extremely low-resource setting(i.e. with small training data and a uni-layered neural network model), we train a Long Short-Term Memory (LSTM), a Gated Recurrent Unit (GRU), and an Attention-based Bidirectional-Gated Recurrent Unit (A-GRU) sequential models to determine the architecture that performs the best for machine translation on our corpora [\textbf{\cite{TiedemannThottingal:2020}}]. We briefly introduce each of these next:

\subsection{Long Short-Term Memory (LSTM)}
Following the research by [\textbf{\cite{sutskever2014sequence}}], we train an LSTM-based neural architecture to perform translation from one source language ($x$) to the target language ($y$). We first append start of sequence (sos) and end of sequence (eos) tokens to the source (English) sentence. This helps our model to understand the content between the start and end of every sentence in the dataset.

\subsubsection{LSTM Encoder}
\begin{figure}[h]
\centering
\includegraphics[width=12cm, height=12cm]{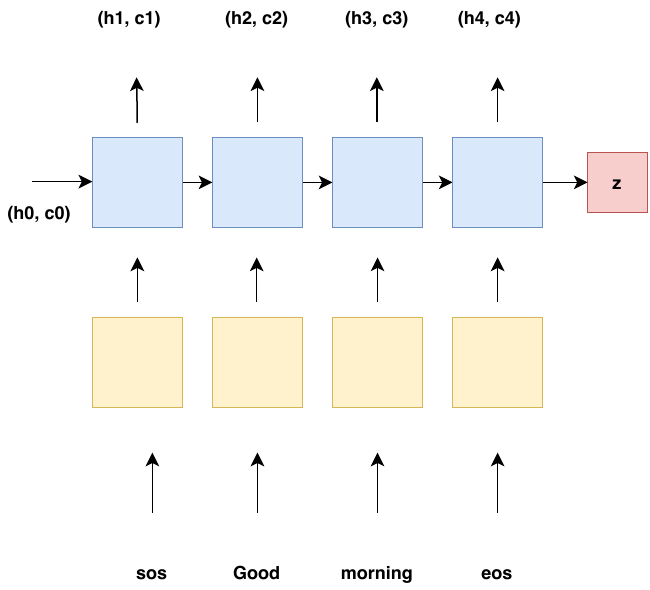}
\caption{The displayed figure represents an LSTM-Encoder for Seq2Seq based Neural Machine Translation system. The source sentence is passed along with the initial hidden and cell states through the left side of the network, to extract the input features into the context vector.}\label{fig:lstm_encoder_architecture}
\end{figure}

The new input, "sos good morning eos", is then passed from the embedding layer (yellow) to the encoder unit (blue) of our network. At each time-step of the LSTM Encoder (Figure \ref{fig:lstm_encoder_architecture}), the input to the encoder of the network is the embedding of the current token ($e(x_{t})$) and hidden state ($h_{t-1}$) from the previous time-step. Together, LSTM Encoder can be represented as a function of $e(x_{t})$ and $h_{t-1}$ as:

$$h_t = \text{LSTMEncoder}(e(x_{t}), h_{t-1})$$

Where $h_{t}$ represent the intermediatory hidden state, which is the vector representation of the sentence so far. Once the final token (eos) from the source sentence has been passed to the LSTM unit through the embedding layer, we use this final hidden state as the context vector ($z$), i.e. the vector representation for the entire source sentence. \newline

\subsubsection{LSTM Decoder}
\begin{figure}[h]
\centering
\includegraphics[width=8cm, height=13cm]{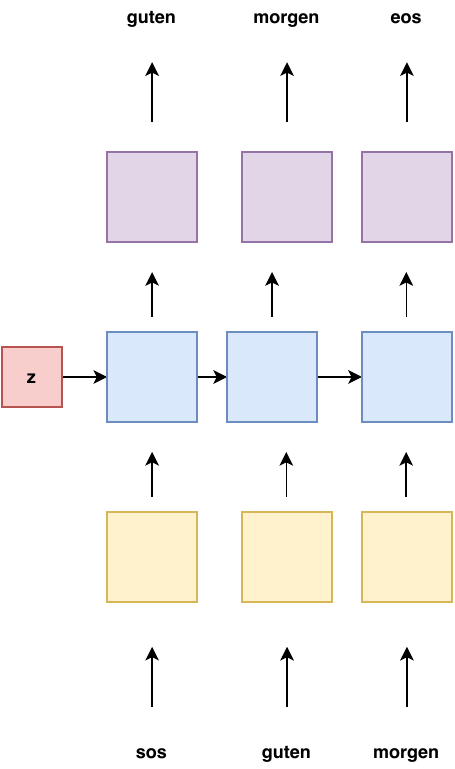}
\caption{The displayed figure represents an LSTM-Decoder for Seq2Seq based Neural Machine Translation system. The context vector extracted by LSTM-Encoder is utilized by the network to generate the target sentence.}\label{fig:lstm_decoder_architecture}
\end{figure}

Once we have the final context vector ($z$), we pass it to the LSTM Decoder (Figure \ref{fig:lstm_decoder_architecture}) to obtain our appended target sentence, "sos guten morgen eos". Now, at each time-step of the LSTM Decoder (blue), we have the embedding of the current token ($d(y_{t})$) and the hidden state ($s_{t-1}$) from the previous time-step. The initial hidden state ($s_{0}$) of the decoder is assigned as the context vector ($z$) obtained from the network's encoder. Please note that although the input/source embedding layer and the output/target embedding layer are both shown in yellow in Figure \ref{fig:lstm_decoder_architecture}, they are two different embedding layers with their own initialized parameters. Similar to the LSTM Encoder, LSTM Decoder can be expressed as a function of $d(y_{t})$ and $s_{t-1}$ as:

$$s_t = \text{LSTMDecoder}(d(y_{t}), s_{t-1})$$

Here, $s_{t}$ represents the intermediatory hidden state, which is the vector representation of the sentence generated so far. The final hidden state is then passed through a linear layer(purple) to predict the next token in the sequence($\hat{y}_t$), in the supervised fashion. 

$$\hat{y}_t = f(s_t)$$

Following the same approach, the decoder generates a token one after another at each time step. During the training of our network, we utilize teacher forcing, a methodology that uses a proportion of actual/ground truth next token in the sequence($y_t$), instead of the token predicted by our decoder($\hat{y}_{t-1}$). During inference, we generate tokens until the model outputs an eos token or after a certain amount of words have been generated. Once the predicted target sentence is achieved, $\hat{Y} = \{ \hat{y}_1, \hat{y}_2, ..., \hat{y}_T \}$, we compare it against our actual target sentence, $Y = \{ y_1, y_2, ..., y_T \}$, and calculate the loss to update the parameters of our model.

\subsection{Gated-Recurrent Units (GRU)}
The Decoder in the LSTM-based architecture described earlier, has a significant downside. Since the hidden states need to contain information from the entire source sequence along with all the tokens decoded so far, the LSTM decoder compresses tons of information into these hidden states. Following the research by [\textbf{\cite{cho2014learning}}], we train a GRU based architecture, to achieve improved test perplexity whilst only using a  layer in both the encoder and the decoder units. Researchers have demonstrated that both LSTMs and GRU's performs similarly, however, GRUs are found to be faster than the LSTMs and they abstract information better [\textbf{\cite{chung2014empirical}}]. \newline

\subsubsection{GRU Encoder}
\begin{figure}[h]
\centering
\includegraphics[width=12cm, height=11cm]{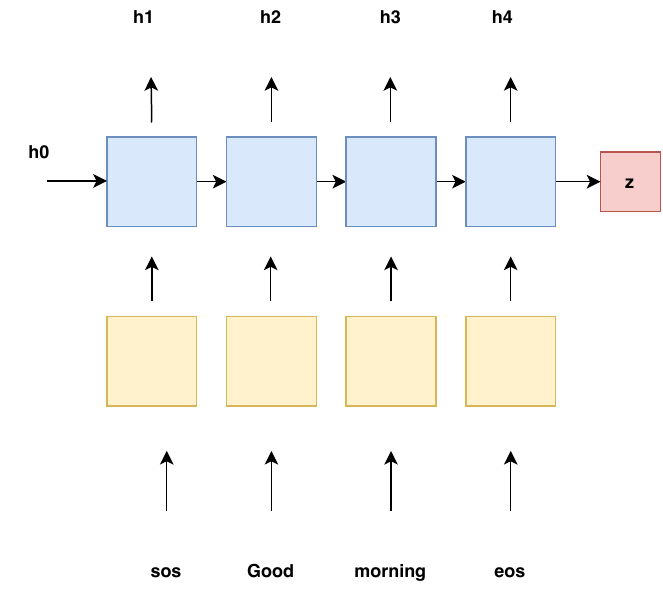}
\caption{he displayed figure represents an GRU-Encoder for Seq2Seq based Neural Machine Translation system. The source sentence is passed along with the initial hidden states through the left side of the network, to extract the input features into the context vector.}\label{fig:gru_encoder_architecture}
\end{figure}

The architecture implemented for GRU Encoder (Figure \ref{fig:gru_encoder_architecture}) is similar to the LSTM Encoder described previously, except the fact that the LSTM units are swapped with the GRUs. Similar to LSTMs, GRUs have several gating mechanisms that control the information flow. However, GRUs only requires and returns hidden states since there are no cell states. Similar to the LSTM Encoder, the GRU Encoder takes in a sequence of inputs, $X = \{x_1, x_2, ... , x_T\}$, and passes it through the embedding layer. A context vector, $z=h_T$,  is then extracted after computing the hidden states, $H = \{h_1, h_2, ..., h_T\}$, obtained from the inputs. In other words, the operations inside the GRU Encoder can be represented through the following mathematical equation:

$$h_t = \text{EncoderGRU}(e(x_t), h_{t-1})$$

\subsubsection{GRU Decoder}
\begin{figure}[h]
\centering
\includegraphics[width=9cm, height=12.5cm]{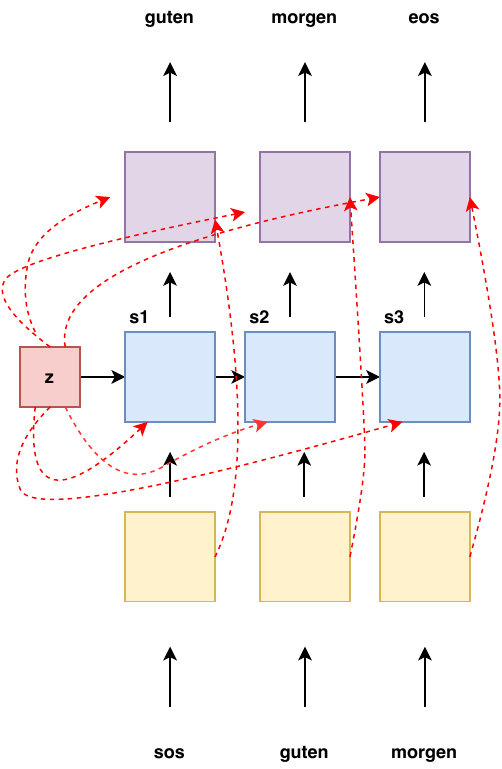}
\caption{The displayed figure represents an GRU-Decoder for Seq2Seq based Neural Machine Translation system. The context vector extracted by LSTM-Encoder is utilized by the network to generate the target sentence.}\label{fig:gru_decoder_architecture}
\end{figure}
In the Decoder (Figure \ref{fig:gru_decoder_architecture}), however, instead of taking just the embedding target token $d(y_t)$ and previous hidden state $s_{t-1}$ as inputs, GRU units also takes the context vector $z$ (dotted-red lines). Note that this context vector from the GRU Encoder is independent of the time-step is re-used as the same, for any time-step in the GRU Decoder. In the end, we pass the embedding of the current token, $d(y_t)$, along with the context vector, $z$, to the linear layer for predicting the next token in the supervised fashion.
 
$$s_t = \text{GRUDecoder}(d(y_t), s_{t-1}, z)$$
$$\hat{y}_{t+1} = f(d(y_t), s_t, z)$$

Since we feed the context vector with the initial hidden state, $s_0$, while generating the first token, there are two identical context vectors inside the GRU Decoder.  Therefore, the decoder hidden states, $s_t$, no longer need to contain information about the source sequence. This means, it only needs to contain the information about what tokens have been generated so far. Additionally, introducing $y_t$ to the final linear layer makes it possible for the decoder to generate tokens without getting it from hidden states. This as a result, reduces the problem of information compression, we faced with LSTM architecture.

\subsection{Attention-based Bidirectional Gated Recurrent Units (A-BGRU)}
Despite the better information compression achieved through GRUs, the context vector obtained through the encoder still needs to contain the entire information from the source sentence. To overcome this problem, we train an attention-based bidirectional GRU model to perform our task. The attention model of our networks calculates an attention vector, $a$, which represents the length of the source sentence. Since every element of this attention vector lies in between 0 and 1 (so that the entire vector space sums to 1), we then calculate a weighted sum of the hidden states, $H$, to get a weighted source vector while decoding, $w$. We use this as an input to our networks Decoder and linear layer to make the prediction.

$$w = \sum_{i}a_ih_i$$

\begin{figure}[h]
\centering
\includegraphics[width=\textwidth, keepaspectratio]{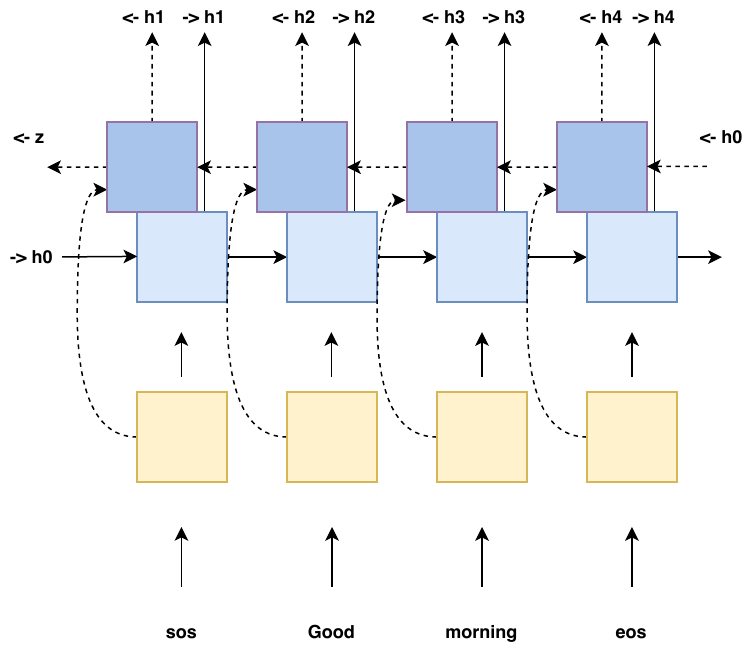}
\caption{The displayed figure represents Bidirectional GRU Encoder for Neural Machine Translation. Since, we are using bidirectional GRUs in the Encoder, the direction of the hidden states, $h_t$, and context vector, $z$, represent the direction of GRU. Therefore, $\rightarrow$ represents hidden states and context vector extracted from the forward GRU units whereas $\leftarrow$ represents hidden states and context vector extracted from the backward GRU units.}\label{fig:bgru_encoder_architecture}
\end{figure}

\subsubsection{Bidirectional GRU Encoder}
Following the research by [\textbf{\cite{bahdanau2014neural}}], we implement our Encoder (Figure \ref{fig:bgru_encoder_architecture}) using a layer of Bidirectional GRU unit. This enables the forward GRU to embedded the source sentence from left to right(light blue) and the backward GRU to embedded the same source sentence from right to left(dark blue). Therefore, just like before, the GRU can be formulated for both the directions by the below-mentioned equations. The direction of the arrows represents the direction of GRU Encoder. While the $\rightarrow$ represents forward direction, the $\leftarrow$ represents backward direction. Therefore, unlike before, we receive two context vectors, $\rightarrow$, and, $\leftarrow$, from the forward and backward directions respectively. The hidden state thus obtained from the GRU Encoder is the concatenation of the forward and backward hidden states, i.e. $h_1 = [h_1^\rightarrow; h_{T}^\leftarrow]$, $h_2 = [h_2^\rightarrow; h_{T-1}^\leftarrow]$ and therefore, the encoder hidden states are represented as $H=\{ h_1, h_2, ..., h_T\}$.  

$$ h_t^\rightarrow = \text{GRUEncoder}^\rightarrow(e(x_t)^\rightarrow, h_{t-1}^\rightarrow) $$
$$ h_t^\leftarrow = \text{GRUEncoder}^\leftarrow(e(x_t)^\leftarrow, h_{t-1}^\leftarrow) $$

\subsubsection{Attention Layer}
\begin{figure}[h]
\centering
\includegraphics[width=\textwidth, keepaspectratio]{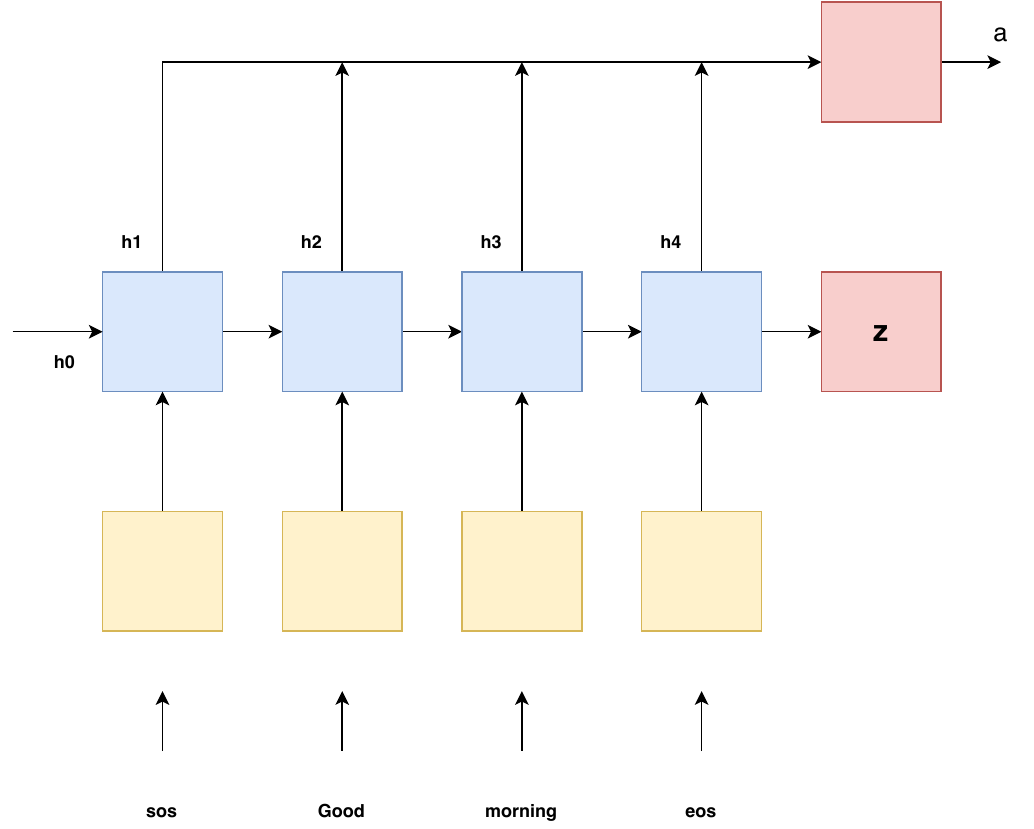}
\caption{The displayed figure represents represents the attention layer for the calculation of the attention vector, where $s_{t-1} = s_0 = z$. The blue blocks represent the concatenated hidden states from both forward and backward directions, whereas the attention is computed within the pink blocks.}\label{fig:attention_layer}
\end{figure}

The attention layer (Figure \ref{fig:attention_layer}) uses the previous hidden states of the decoder, $s_{t-1}$, along with the stacked forward and backwards hidden states from the encoder, $H$. An attention vector, $a_t$, equal to the length of the source sentence is then generated, where every element of the vector is between 0 and 1, and the entire vector sums to 1. This attention vector represents the features(token/words) in the source sentence that the model should emphasize on, to correctly predict the next token, $\hat{y}_{t+1}$. \newline

To determine the attention vector, we first calculate the energy between the previous decoder and the encoder hidden states. Since the encoder hidden states are sequences of N tensors, we duplicate the previous decoder hidden state T times. These hidden states are then concatenated and passed through a linear layer and tanh activation function.

$$E_t = \tanh(\text{Attention}(s_{t-1}, H))$$

The attention vector is finally obtained by concatenating the weighted sum of the energy across all encoder hidden states to the Energy vector obtained. The weights show the significance of each token in the form of attention over the source sentence. The randomly initialized parameters of the attention are learned along with the other parameters of the model during backpropagation. At last, the attention vector achieved is passed through a linear layer to ensure that all elements are between 0 and 1 and the vector sums to 1.

$$a_t = \text{softmax}(\hat{a_t})$$

\subsubsection{GRU Decoder}
\begin{figure}[h]
\centering
\includegraphics[width=\textwidth, keepaspectratio]{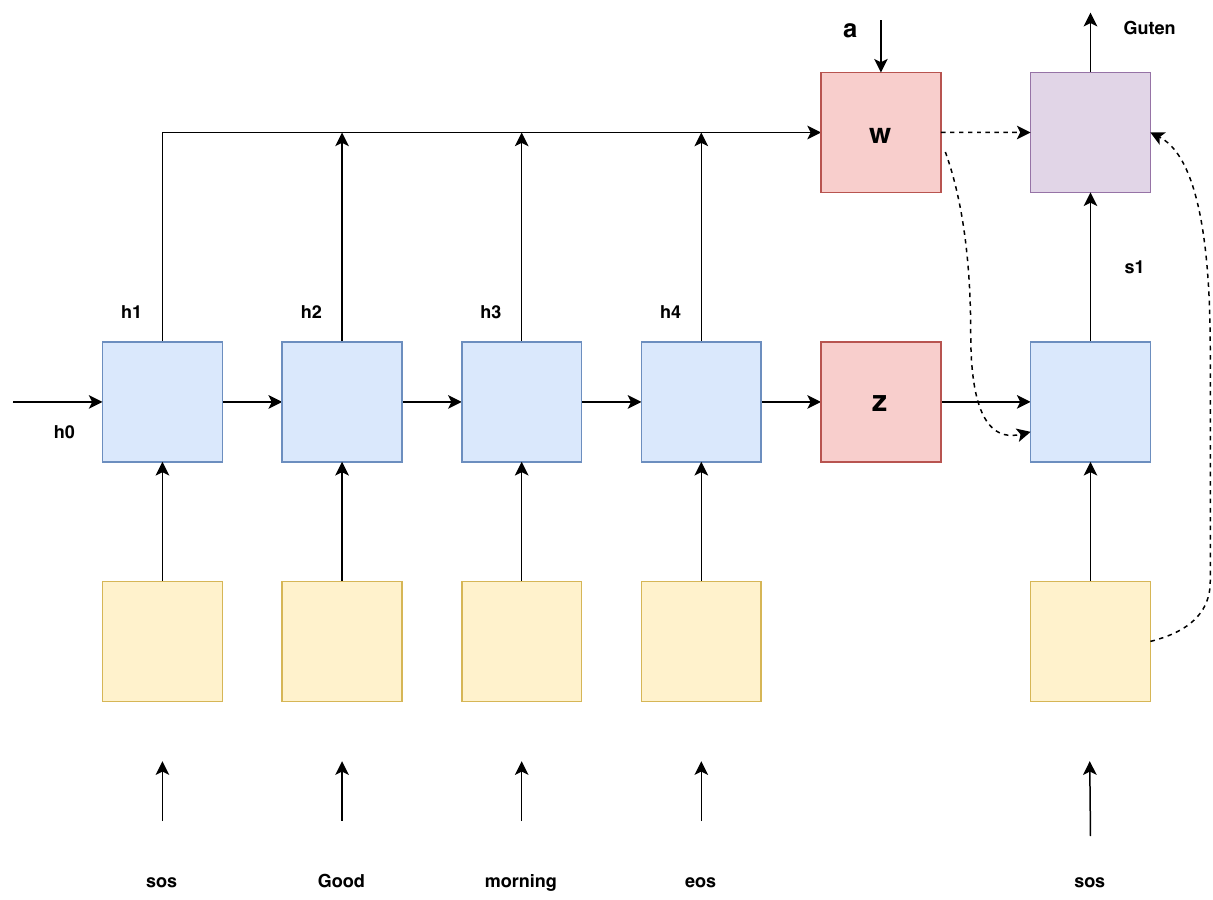}
\caption{The blue blocks on the left represent the concatenated hidden states from both forward and backward directions of GRU Encoder, whereas the attention is computed within the pink blocks. The blue box on the right represents GRU Decoder that outputs $s_{t}$. The purple block shows the linear layer, $f$, that outputs the next predicted token, $\hat{y}_{t+1}$, and the orange block shows the calculation of the weighted sum over $H$ by $a_t$.}\label{fig:bgru_decoder_architecture}
\end{figure}

The Decoder utilizes attention layer that takes the previous hidden states, $s_{t-1}$, along with the encoded hidden states, $H$, to return an attention vector, $a_t$. This attention vector, $a_t$, is concatenated with the weighted sum of the encoder hidden states, $H$, to create a weighted source vector, $w_t$. As shown in Figure \ref{fig:bgru_decoder_architecture}, the embedded input, $d(y_t)$, and the previous decoded hidden state, $s_{t-1}$,  together along with the concatenation of $d(y_t)$ and $w_t$ are passed into the GRU Decoder. At last, $d(y_t)$, $w_t$, and $s_t$  is passed through the linear layer, $f$, to generate the predicted next token, $\hat{y}_{t+1}$, for a target sequence in a supervised fashion. 

$$w_t = a_t H$$
$$s_t = \text{GRUDecoder}(d(y_t), w_t, s_{t-1})$$
$$\hat{y}_{t+1} = f(d(y_t), w_t, s_t)$$

Note that, since our decoder is unidirectional, it is only possible to send one context vector. Unlike in the original research [\textbf{\cite{bahdanau2014neural}}], that uses the backward hidden states from the GRU Encoder, we concatenate the forward and backward context vectors using the tanh activation function, before passing them through the linear layer.

$$z=\tanh(g(h_T^\rightarrow, h_T^\leftarrow)) = \tanh(g(z^\rightarrow, z^\leftarrow)) = s_0$$
 
\pdfminorversion=4

\chapter{Dataset} 

\label{Chapter3} 
Since one of the goals of this work is to train a generalized Multilingual Neural Machine Translation System, we train our network on the dataset provided by the Tatoeba Translation Challenge [\textbf{\cite{TiedemannThottingal:2020}}]. The dataset covers over 500 languages for the task of Machine Translation in real-world scenarios, under realistically low-resource settings. The translation challenge includes training data from OPUS parallel corpus [\textbf{\cite{TiedemannN04}}] and test data from the Tatoeba corpus via the aligned dataset in OPUS. In this work, we show the performance of our network only on English-German, English-French, and English-Spanish datasets from [\textbf{\cite{TiedemannThottingal:2020}}]. Additionally, since the source language for our system is always consistent i.e. English, we pre-train our network on English-German dataset from Tatoeba, with the source and target languages both being English. For more details, check sec. 

\section{Data-insights}
\begin{figure}[H]
\centering
\includegraphics[width=\textwidth, keepaspectratio]{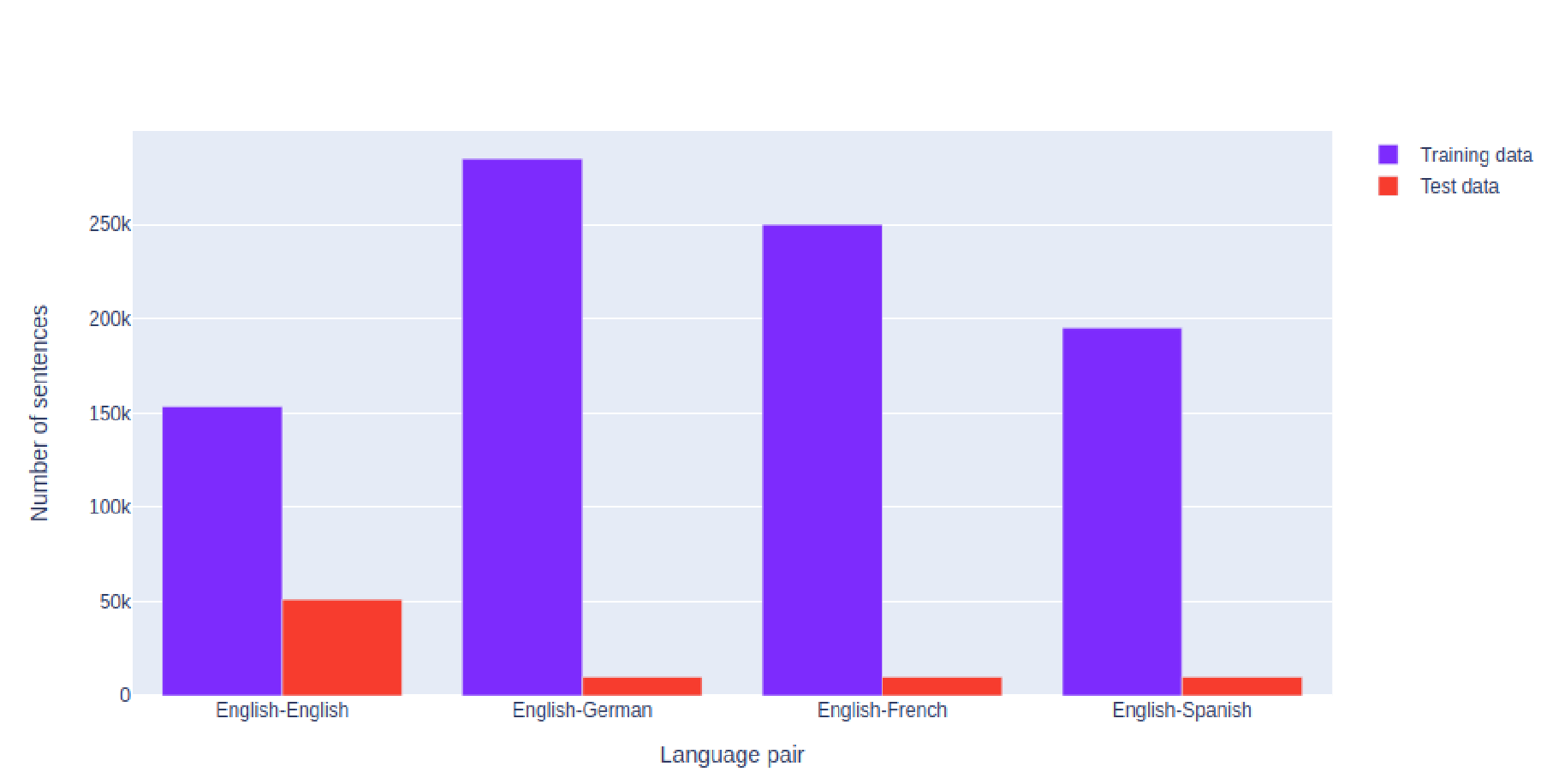}
\caption{Statistics on number of instances for training and test data, in the original Tatoeba Translation Challenge dataset.}
\label{fig:raw_data}
\end{figure}

As shown in the Fig. \ref{fig:raw_data}, the original Tatoeba Translation Challenge datasets contains 284768, 250098, and 194977 bilingual sentences for English-German, English-French, and English-Spanish in the training dataset respectively. Whereas, there are only 10000 bilingual sentences in each test dataset. For the pre-training(i.e. English-English), we have used the \href{manythings.org}{Tatoeba English-German bilingual corpus}, which has 153430 training sentences and 51144 test sentences. \newline

\begin{figure}[H]
\centering
\includegraphics[width=\textwidth, keepaspectratio]{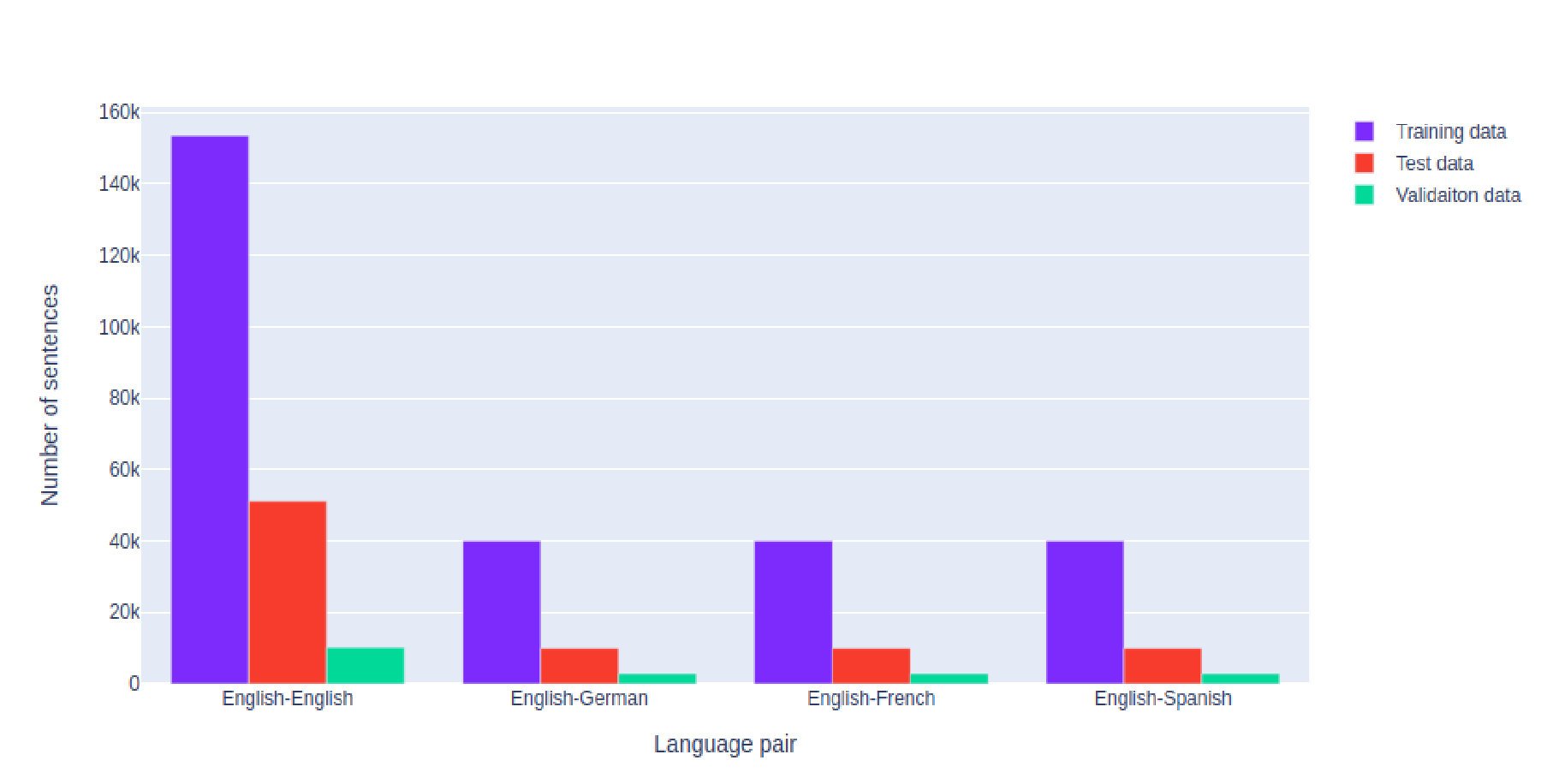}
\caption{Statistics on number of instances for training and test data, in the trimmed Tatoeba Translation Challenge dataset.}
\label{fig:processed_data}
\end{figure}

Since not all the sentences in all these corpora are parallelly-alligned, we decided to trim the original Tatoeba Translation Challenge datasets to obtain 40000, 2770, and 10000 bilingual sentences(Fig. \ref{fig:processed_data}) each in training, validation, and test datasets. Since one of the objectives of our work is to quantify and interpret knowledge-transfer between different languages, we decided to keep the datasets parallelly aligned and of the same size. For pre-training, we argue that since the source and target languages are the same, no parallel data is required. Ideally, one can always take sentences from the Wikipedia corpus and use it for pre-training. Therefore, we decided to use 153430 training sentences and 51144 test sentences the Tatoeba (manythings.org) English-German bilingual corpus, for the pre-training purpose.\newline

\section{Pre-processing and data cleaning}
To keep things simple, we have avoided heavy data pre-processing/cleaning. Not only we removed all the special symbols and characters to keep the data consistent, but spelling checks and proper contractions (like you’re -> you are, won’t -> will not) are performed for the English language in every corpus. Also, we decided not to use any pre-trained embeddings during our analysis. For tokenization purposes, we have used pre-trained Spacy tokenizers.

\pdfminorversion=4

\chapter{Experiments} 

\label{Chapter4} 

\section{Architectural details}
\begin{table}[h!]
  \begin{center}
    \caption{\textbf{Hyperparameter setting} : Hyperparameter values used accross different architectures}
    \label{tab:hyperparameters}
    \begin{tabular}{l|c|c|r} 
      \textbf{Hyperparameters} & \textbf{LSTM} & \textbf{GRU} & \textbf{A-BGRU}\\
      \hline
      Embedding size & 300 & 300 & 300\\
      Layers & 1 & 1 & 1\\
      Hidden units & 512 & 512 & 512\\
      Epochs (with Early stopping) & 50 & 50 & 50\\
      Dropout & 0.5 & 0.5 & 0.5\\
      Learning rate & 0.001 & 0.001 & 0.001\\
      Batch size & 40 & 40 & 40\\
      Regularizer & L2 & L2 & L2\\
      Gradient clipping rate & 5 & 5 & 5\\
    \end{tabular}
  \end{center}
\end{table}

Since the objective of the experiments mentioned above is to find the most suitable architecture for Neural Machine Translation in a low-resource setting, we decided to retain our hyperparameters setting almost identical across all the architectures. \newline

The implementation of the networks has been carried out in Pytorch. For optimization purpose, we have used Adam; a method of Stochastic Optimization with L2 penalty. Since our end task is to perform machine translation, we presume it as an N-class classification problem, where N represents the unique number of tokens/classes present in the target vocabulary. Therefore, we use CrossEntropy as our loss function that contains a combination of the logarithmic function of Softmax, and negative log-likelihood to solve a classification problem of N classes. \newline

We obtain better results by training a multi-layered architecture. However, due to computational and low-resource setting reasons, we have focused our research by training and comparing uni-layered models only. During our experiments, we manually set the padding index to 0 and make sure that our loss function remains unaffected from the padded tokens. We achieve this by ignoring this padding index, i.e. 0 while computing the loss function. We use a standard split of 70-30 to divide the pretrained data, i.e. Tatoeba English-English data [\textbf{\cite{TiedemannThottingal:2020}}] into training and test dataset.

\section{Baselines}
\label{sec:baselines}
Since we only train our models on the 40K parallel sentences shared between English-German, English-French, and English-Spanish corpora in Tatoeba Translation Challenge [\textbf{\cite{TiedemannThottingal:2020}}], we don't have a SOTA baseline. Therefore, we find it essential to set up our baselines first. 

\subsection{End-to-End training}
\label{sec:end_to_end}
\begin{figure}[h]
\centering
\includegraphics[width=\textwidth, keepaspectratio]{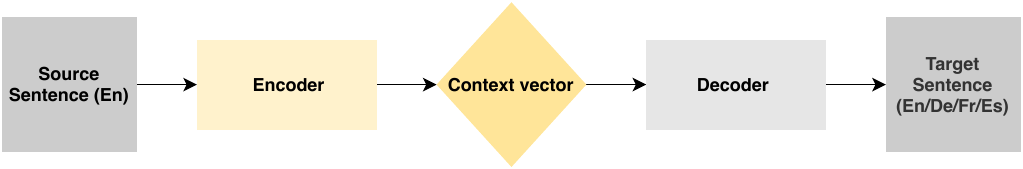}
\caption{\textbf{End-to-End training} : The English source sequences is fed from the left side of the network. The Encoder extracts the input representations to generate a context vector, which is fed to the right side of the network, where the Decoder generates a English, German, French , or Spanish target sequence.}\label{fig:end2end}
\end{figure}

For the end-2-end baseline (Figure \ref{fig:end2end}), we train four different networks to perform English-English, and English-German, English-French, and English-Spanish translations. We take the training and test corpora for these networks from English-German dataset in [\textbf{\cite{TiedemannN04}}], and English-German, English-French, and English-Spanish from [\textbf{\cite{TiedemannThottingal:2020}}], respectively. \newline

We generate these results to have a rough estimation of how the network performs on our extracted parallel-dataset, for direct source-target translation. Please note that since we do not apply any transfer in our end-to-end networks, we find it unnecessary to share the vocabulary across all the different datasets. Also, for English-English translations,  we take the source sequence from the from English-German dataset in [\textbf{\cite{TiedemannN04}}] and copy the same as the target sequence. Further details and explanations regarding the same are discussed in section \ref{sec:1-hop}.   

\subsection{1-hop Transfer}
\label{sec:1-hop}
\subsubsection{Step 1: Pre-training}
\begin{figure}[h]
\centering
\includegraphics[width=\textwidth, keepaspectratio]{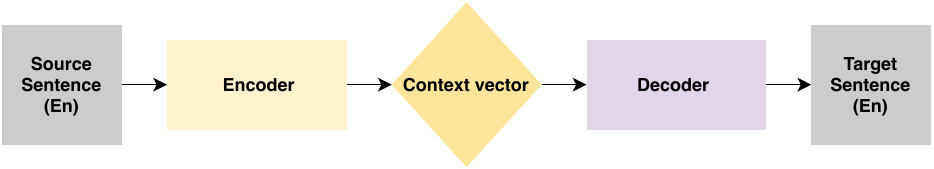}
\caption{\textbf{Pre-training} : The English source sequences is fed from the left side of the network. The Encoder extracts the input representations to generate a context vector, which is to the right side of the network, where the Decoder generates an English (grey-color) target sequence.}\label{fig:pretraining_architecture}
\end{figure}

While End-to-End baseline shows how well our model performs for direct source-target translations, 1-hop transfer shows the effect of the direct transfer, from our pre-trained knowledge on the fine-tuned language. Since we perform all the translations from the English language, we argue that the model must develop an understanding of how to represent the source language perfectly. Therefore, we pre-train our Encoder-Decoder model to perform translations from English to the English language from English-German dataset in [\textbf{\cite{TiedemannN04}}] first (Figure \ref{fig:pretraining_architecture}). \newline

While most of the low-resource research in neural machine translation focuses on heavy pre-training [\textbf{\cite{johnson-etal-2017-googles}} and \textbf{\cite{aharoni2019massively}}] from English-French on WMT corpora, we argue to utilize English sentences from any known dataset and duplicating it for target sequences. This increases the potential of the pre-training phase, as one can ideally utilize transfer from the entire Wikitext corpus by using it similarly to our English-English pre-training.

\subsubsection{Step 2: 1-hop transfer}
\begin{figure}[h]
\centering
\includegraphics[width=\textwidth, keepaspectratio]{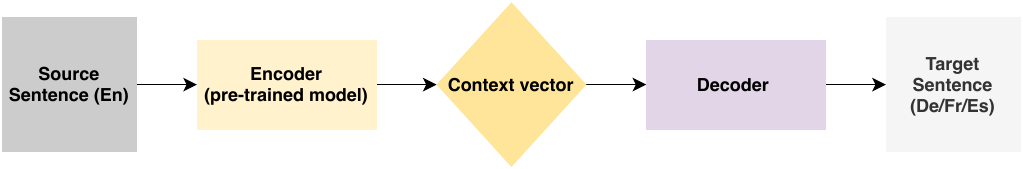}
\caption{\textbf{1-hop transfer} : The figure demonstrates the fine-tuning of three different models on the pre-trained English-English model to perform English-German, English-French, and English-Spanish translations. As can be seen, the Encoder from the pre-trained model is shared during fine-tuning while training in transfer learning fashion.}\label{fig:1-hop-transfer}
\end{figure}

For the 1-hop transfer, we take the English-English dataset from English-German dataset in [\textbf{\cite{TiedemannN04}}] and share its input vocabulary with English-German, English-French, and English-Spanish dataset in [\textbf{\cite{TiedemannThottingal:2020}}]. As shown in Figure \ref{fig:1-hop-transfer}, we then transfer the knowledge from English-English dataset to English-German, English-French, and English-Spanish datasets by using the pre-trained model from step 1 and separately fine-tuning it for English-German, English-French, and English-Spanish translation models respectively. \newline

As mentioned earlier, to keep things simple and minimize catastrophic forgetting, we choose to freeze the weights of our pre-trained model before applying it for fine-tuning. Since the end-task for pre-trained and fine-tuned models are different, i.e. translation to different languages, we only share the Encoder from our pre-trained model during fine-tuning. 

\section{RQ1: Utilizing Knowledge-transfer between different languages}
\label{sec:knowledge_transfer_training}
To understand the influence of knowledge-transfer from one/many languages to another, we train our network using both joint multi-task and sequential transfer learning approaches. In this section, we describe each one of them separately:

\subsection{Joint Multi-task transfer Learning}
Unlike sequential-transfer learning (described in section \ref{sec:sequential_transfer}) where the transfer of knowledge is only in the forward direction, we allow the knowledge-transfer to flow from all the target languages. The idea is to compare and observe how the knowledge-flow from all the target languages destroys and build knowledge on the top of each other in a Multilingual-Neural Machine Translation system.

\subsubsection{Step 1: Pre-training}
Similar to the training performed in Figure \ref{fig:pretraining_architecture}, we perform pre-training on English-German dataset from [\textbf{\cite{TiedemannN04}}]. As mentioned earlier, we duplicate the source sequences to target sequences and train a neural machine translation model to perform English-English translations for pre-training. The idea here is to let again the model have an understanding of how to represent the source language before applying transfer-learning.

\subsubsection{Step 2: Multi-task transfer}
To enable the knowledge-transfer flow from every target language, we first combine the training and test datasets from English-German, English-French, and English-Spanish corpora in [\textbf{\cite{TiedemannThottingal:2020}}]. Like earlier, we share the input vocabulary from our extracted English-English dataset with the new combined dataset. The idea is to generate multilingual translations by fine-tuning on our pre-trained English-English translation model. Therefore, we call this training as the joint multi-task transfer training, as all the end-tasks are combined and provided at the get-go. \newline

\begin{figure}[h]
\centering
\includegraphics[width=7cm, height=12cm]{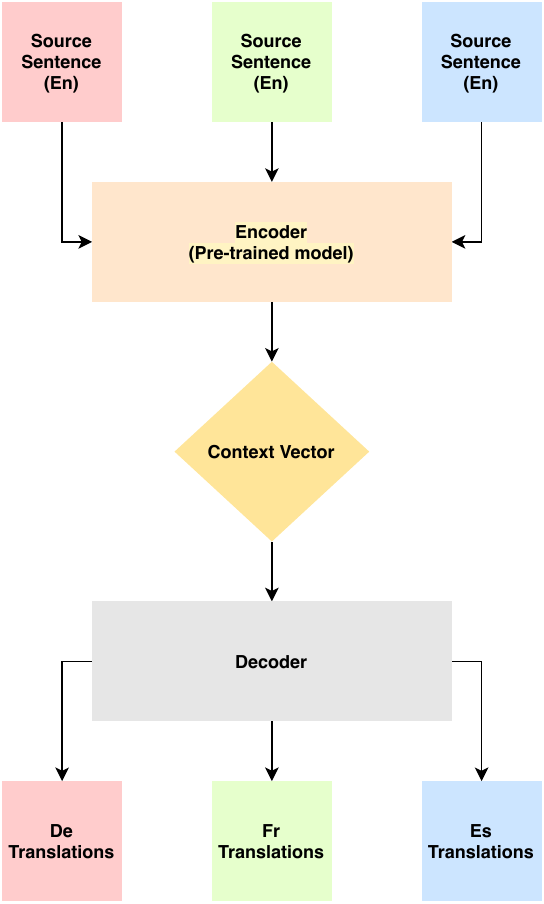}
\caption{\textbf{Joint muti-task training} : The figure demonstrates the methodology used for joint multi-task transfer learning. We combine source sentences from English-German (red colour), English-French (green colour), and English-Spanish (blue colour) datasets are send them to the Encoder of our pre-trained network. A context vector is then generated, which is utilized by the Decode of our network to perform multilingual translation.}\label{fig:parallel_transfer_architecture}
\end{figure}

As shown in Figure \ref{fig:parallel_transfer_architecture}, the Encoder from the pre-trained English-English translation model is shared while fine-tuning our model on the new combined dataset. We argue that since our objective is to understand how knowledge-transfer from multiple languages destroys and build to model learning, we optimize our loss together for multilingual translations. Note that this is completely different from the conventional multitask training, where all the losses are optimized separately to find a common local minima.

\subsubsection{Step 3: Testing}
\label{sec:parallel_testing}
\begin{figure}[h]
\centering
\includegraphics[width=7.5cm]{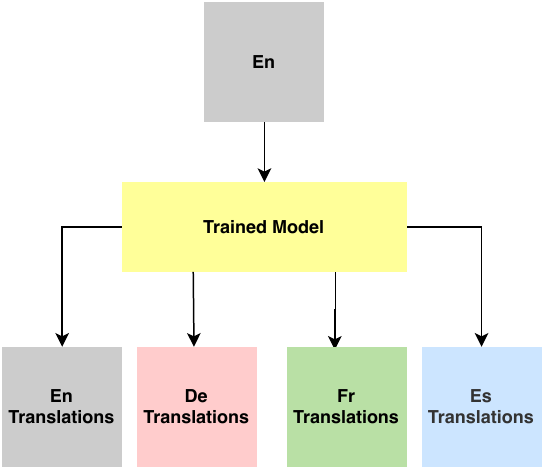}
\caption{\textbf{Testing multi-task model} : The figure demonstrates testing procedures followed for evaluating our trained model. In the first case, the combined source sentences from the test data of English-German, English-French, and English-Spanish datasets are sent to the trained model. The evaluation is then performed based on the BLEU-4 cumulative score on the combined dataset. In the second case, all of them is evaluated separately. }\label{fig:parallel_transfer_test}
\end{figure}

Since we save our model on the basis of combined validation loss, we test our trained model in two different circumstances (Figure \ref{fig:parallel_transfer_test}). In the first case, we try to obtain the combined BLEU-4 score on the combined test dataset from the English-German (red colour), English-French (green colour), and English-Spanish (blue colour) dataset in [\textbf{\cite{TiedemannThottingal:2020}}]. This shows our models capability to perform multilingual translations. In the second case, we evaluate all the three datasets separately. We argue that this helps us to determine how different languages influences model learning differently. \newline

We also compare model-learning through knowledge-transfer between our parallelly trained transfer learning model and sequential-transfer learning model described in the next section. We hypothesize, that since we are conducting our experiment for low-resource setting, the flow of knowledge-transfer from all the languages should destroy networks performance. We further elaborate our hypothesis and conclusions in the result and visualization sections. 

\subsection{Sequential Transfer Learning}
\label{sec:sequential_transfer}

In our sequential-transfer learning setup, we transfer knowledge only in the forward direction. In other words, the fine-tuned model does not affect the knowledge-abstraction of the pre-trained model at any stage of training. At each point of transfer, we freeze the weights of our pre-trained model to minimize the effects of catastrophic forgetting.  This type of training can also be related to continual learning in NLP. Similar to the theory of lifelong learning, new tasks are added at every stage of training one after another. 

\subsubsection{Step 1: Pre-training}
As mentioned earlier, we perform similar experiments as demonstrated in Figure \ref{fig:pretraining_architecture}, for pretraining our English-English Translation model. 

\subsubsection{Step 2: En-De Transfer}
\begin{figure}[h]
\centering
\includegraphics[width=\textwidth, keepaspectratio]{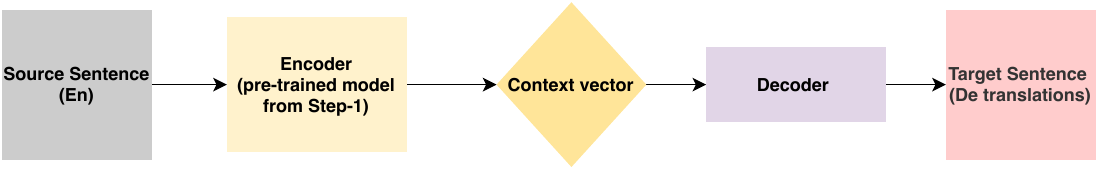}
\caption{\textbf{En-De transfer} : The figure demonstrates the fine-tuning of the trained model from the previous step, in transfer learning fashion to perform English-German (red colour) translations. }\label{fig:de_transfer}
\end{figure}

In the second step, we transfer the knowledge from the extracted English-English corpus (from English-German dataset) in [\textbf{\cite{TiedemannN04}}], to the English-German corpus in [\textbf{\cite{TiedemannThottingal:2020}}]. Like we mentioned earlier, we freeze the weights of the Encoder from the pre-trained model in step 1, and perform fine-tuning on the new corpus to achieve English-German translations (Figure \ref{fig:de_transfer}). 

\subsubsection{Step 3: De-Fr Transfer}
\begin{figure}[h]
\centering
\includegraphics[width=\textwidth, keepaspectratio]{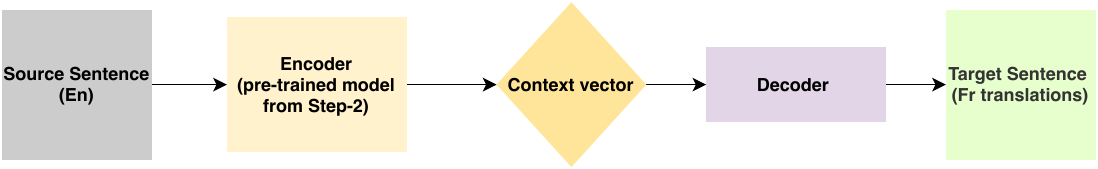}
\caption{\textbf{En-Fr transfer} : The figure demonstrates the fine-tuning of the trained model from the previous step, in sequential-transfer learning fashion to perform English-French (green colour) translations. }\label{fig:fr_transfer}
\end{figure}

In the third step, we transfer the knowledge from the extracted English-English corpus (from English-German dataset) in [\textbf{\cite{TiedemannN04}}] and English-German corpus in [\textbf{\cite{TiedemannThottingal:2020}}], to the English-French corpus in [\textbf{\cite{TiedemannThottingal:2020}}]. Once again, we freeze the weights of the Encoder for the previously trained model in step 2 and fine-tune it on the new corpus to achieve English-French translations (Figure \ref{fig:fr_transfer}). 

\subsubsection{Step 4: Fr-Es Transfer}
\begin{figure}[h]
\centering
\includegraphics[width=\textwidth, keepaspectratio]{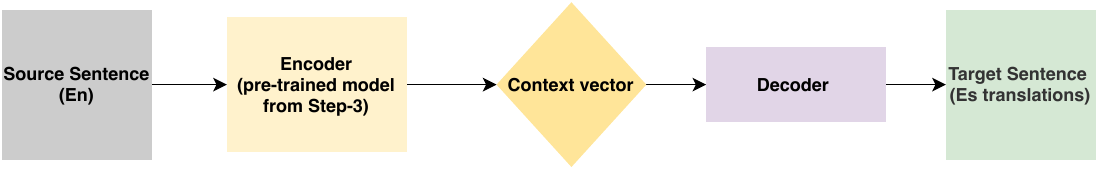}
\caption{\textbf{En-Fr transfer} : The figure demonstrates the fine-tuning of the trained model from the previous step, in sequential-transfer learning fashion to perform English-Spanish (blue colour) translations.}\label{fig:es_transfer}
\end{figure}

For the final step, we transfer the knowledge from the extracted English-English corpus (from English-German dataset) in \cite{TiedemannN04}, English-German, and English-French corpora in [\textbf{\cite{TiedemannThottingal:2020}}], to the English-Spanish corpus in [\textbf{\cite{TiedemannThottingal:2020}}]. Once again, we use the trained model in step 3 as the pre-trained model for this step and fine-tune it on the new corpus to achieve English-Spanish translations (Figure \ref{fig:es_transfer}). 

Note we choose the order of the dataset to be trained, based on languages that have similar roots. We argue that since German is closer to English than French or Spanish, the transfer of knowledge should be better from English to German. Similarly, since  French and Spanish have the same roots, we expect higher transfer between them.

\subsubsection{Step 5: Testing}
Unlike the testing performed in section \ref{sec:parallel_testing}, we evaluate each model individually at every training steps on the corresponding test dataset. Since we freeze the weights of our trained model at every step, we argue that this helps us to obtain the multilingual translation without the influence of fine-tuned knowledge over the pre-trained model.

\section{RQ2: Selective Pruning} 
\label{sec:pruning}
As described in section \ref{sec:catastrophic_forgetting}, deep neural networks are widely known to experience catastrophic forgetting. Since we are training our multilingual neural machine translation system in an extremely low-resource setting, i.e. with 1 layer and 512 neurons, the effects of catastrophic forgetting would be even severe. In this section, we run a few experiments to gain insights on whether pruning selective neurons-knowledge can help our system to perform better. \newline

Similar to the max-activation based activation pruning approach performed in [\textbf{\cite{rethmeier2019txray}}], we test the effects of pruning max, most n, and least n mass-activations on our multilingual neural machine translation setup. We define mass activations as the element-wise sum of all the activations, collected over the entire test dataset. While pruning approaches like Lottery ticket [\textbf{\cite{frankle2018lottery}}] have shown promising results by reducing the number of training parameters without compromising the performance of a network, applying it to a low-resource setting does not make much sense. Because, in a low-resource setting, the neurons are packed with crucial information and pruning the entire neurons increases catastrophic forgetting. Therefore, instead of pruning neuron as the whole, we decided to only prune the neuron-knowledge in pre-trained models, by turning the magnitude of selected neurons in the weight matrix to 0. We argue that pruning the most and least reactive neurons from the mass activation matrix show us effects similar to pruning specialized and negative neuron-knowledge. \newline

\subsection{Pruning dead neurons from the mass max-activation matrix}
\label{sec:max-activation}
\begin{figure}[h]
\centering
\includegraphics[width=\textwidth, keepaspectratio]{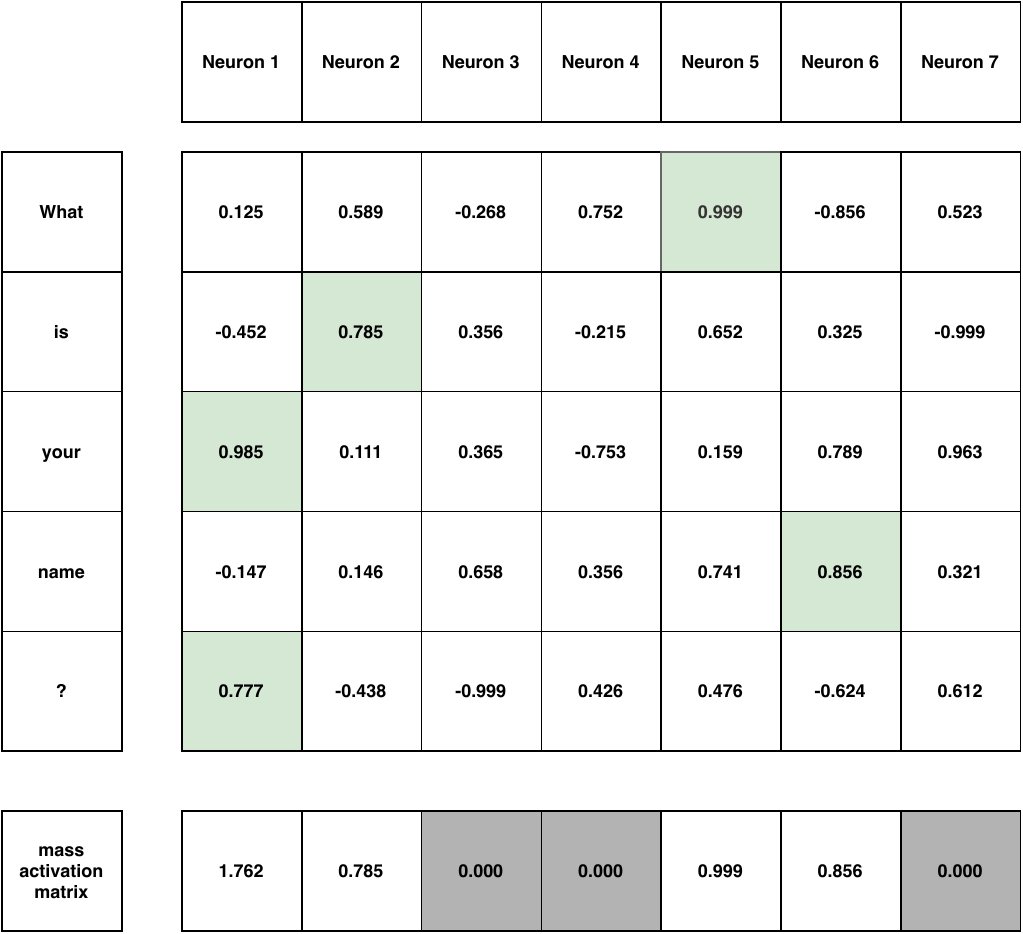}
\caption{\textbf{Pruning dead-neurons from the max mass-activation matrix} : The figure demonstrates an example of the mass max-activation matrix, collected over every sample in the test dataset. The light blue boxes represent the max-activation for every input feature in the activation matrix, whereas the grey boxes represent the dead neurons, i.e. the neurons that were never activated or got deactivated after transfer.}\label{fig:max_activation}
\end{figure}

Inspired by the max-activation experiment performed in [\textbf{\cite{rethmeier2019txray}}], we perform similar experiments for our multilingual neural machine translation system. Once the model has been trained, we feed input sequences from the test dataset through the network, on evaluate mode. While passing a sample from the source sequence, the Encoder generates a context vector in the form of hidden and cell states. We extract these hidden states from every token in the test sequence and collect them together. \newline

To properly understand the procedure in detail, let's consider a small example where the test dataset has a single source sequence (Figure \ref{fig:max_activation}), i.e. "What is your name? ". In the first step, we send the sequence to the Encoder of the trained model in evaluate mode to generate a context vector. By the end of the sequence, we will receive an output of shape 5x7 where 5 is the number of input features, aka tokens in the sequence, and 7 is the number of neurons in the trained network. These activations lie anywhere in-between from -1.000 to +1.000 in magnitude. The more negative the activation, the negatively it affects the performance of our system to generate accurate translations. We define max-activated neurons as the most activated neurons (in magnitude), for every token in the sequence. For every neuron in our network, we then add these max activations (light blue - boxes) to obtain the mass-activation matrix. The mass activation matrix represents the knowledge content for every neuron, collected over the entire test dataset. Note that the neurons that were never activated or got deactivated after application of transfer have a mass activation of 0.000 (grey boxes). We call these neurons as the dead neurons. \newline

At every stage of training, different neurons get activated or deactivated depending upon the knowledge-transfer. In this experiment, we prune these dead neurons from the pre-trained model at every step, before fine-tuning our model on a new dataset. We perform this at every step of training until we reach our final end-task, i.e. English-Spanish translations.  

\subsection{Pruning most-n activated neuron-knowledge from the mass-activation matrix}
\label{sec:most-activation}
\begin{figure}[h]
\centering
\includegraphics[width=\textwidth, keepaspectratio]{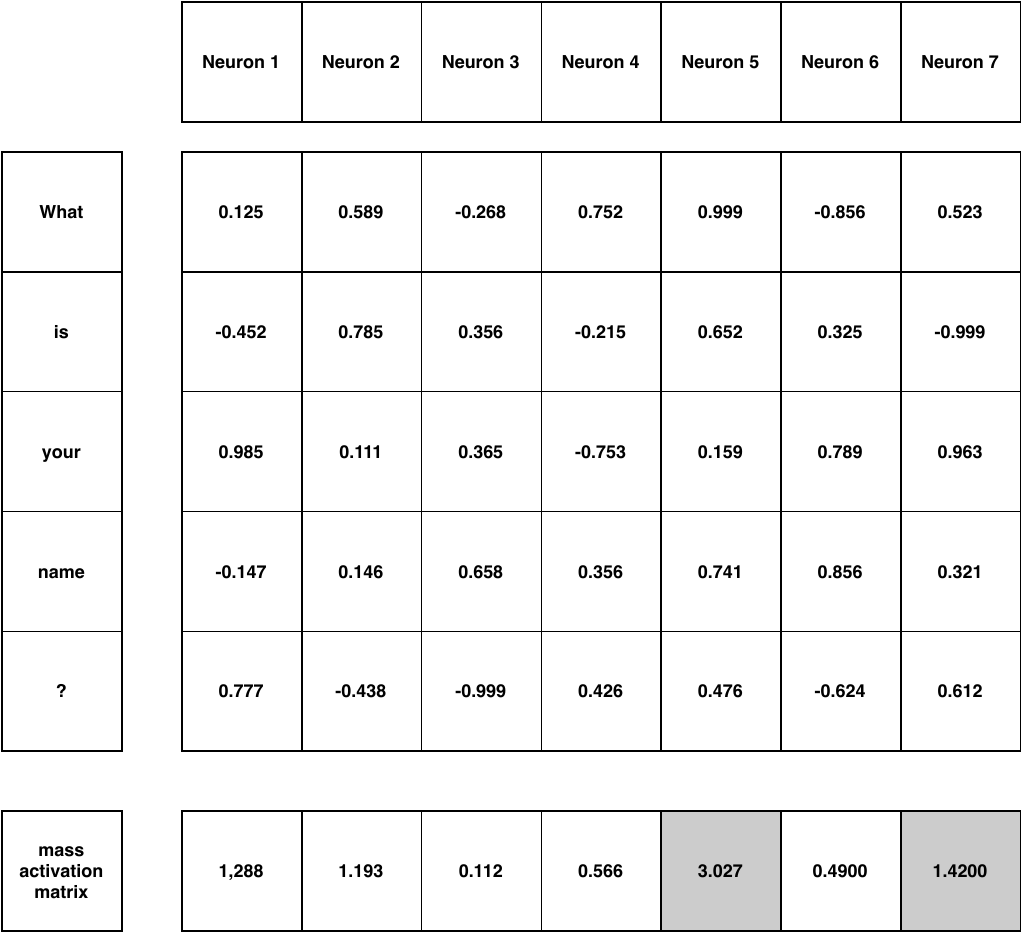}
\caption{\textbf{Pruning neurons with the most activated neuron knowledge} : The figure demonstrates an example of the most n-activated neuron in mass activation matrix, collected over every sample in the test dataset. The grey boxes represent the most activated n neurons, for n being 2. The shaded neurons represent the specialized knowledge, represented by the network for a given end-task.}\label{fig:most_activation}
\end{figure}

Similar to the approach mentioned in section \ref{sec:max-activation}, we collect the activation data for every sample in the test dataset. We argue that despite the max-activation representing the influence of a token/word on a certain neuron, it doesn't reveal the influence entire sequence, otherwise known as context on the same neuron. Therefore, instead of just acknowledging the max-activation like earlier, we consider the entire activation matrix this time. \newline

Considering the same example as earlier (Figure \ref{fig:most_activation}), we collect the activation data for the sequence, "What is your name?". However, instead of looking through the activation matrix for max-activation, we consider the entire activation matrix and perform summation over the amplitude of activation for every neuron. We call this 1-D array as the mass-activation matrix in our further experiments. \newline

Depending upon the value of n (2 in the example provided), we choose n neurons with the most activation (grey box) in the mass activation matrix and prune the knowledge within them, from the pre-trained model before transfer. We perform this at every step of training until we reach our final end-task, i.e. English-Spanish translations. We argue that by pruning the most activated neurons, we are removing the specialized knowledge from every task. Since we are only pruning the neuron-knowledge and not the entire neuron, this method does not affect the performance of our pre-trained model. 

\subsection{Pruning least-n activated neuron-knowledge from the mass-activation matrix}
\label{sec:least_n}
\begin{figure}[h]
\centering
\includegraphics[width=\textwidth, keepaspectratio]{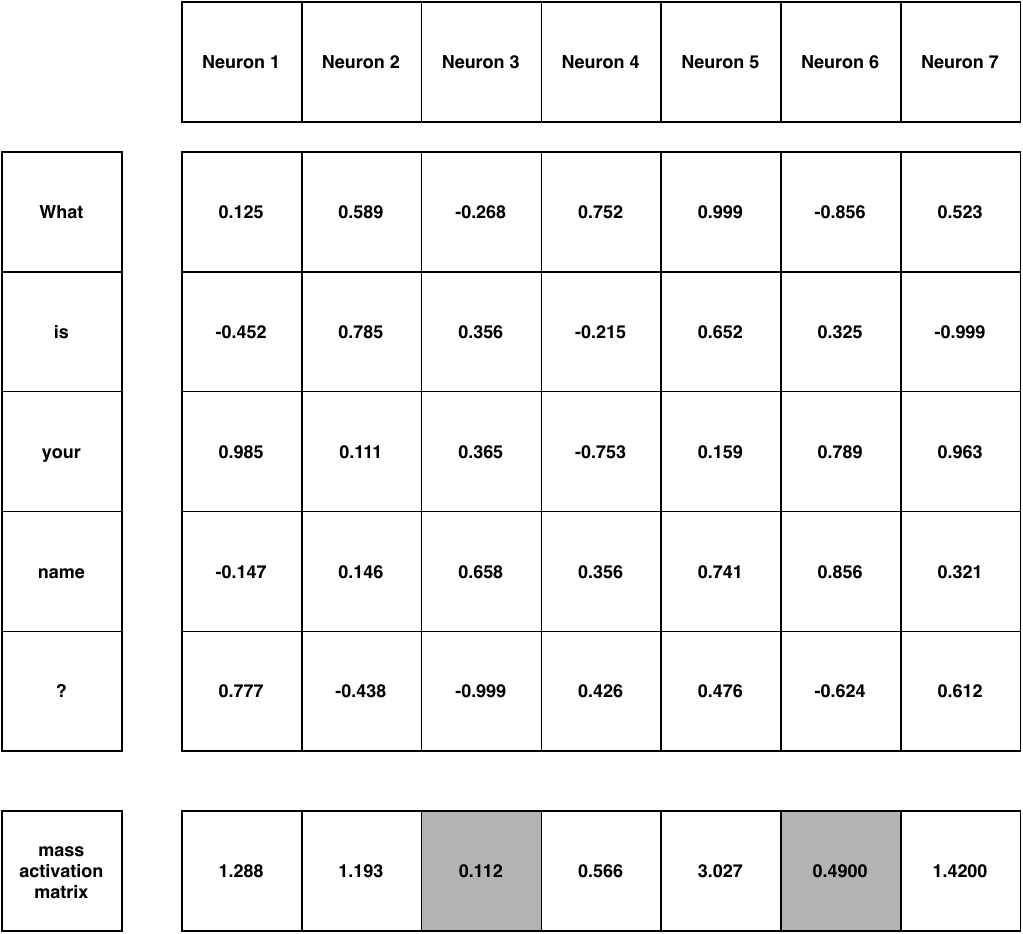}
\caption{\textbf{Pruning neurons with the least activated neuron knowledge} :The figure demonstrates an example of the least n-activated neuron in mass activation matrix, collected over every sample in the test dataset. The grey boxes represent the least activated n neurons, for n being 2. The shaded neurons represent the least-reactive knowledge, represented by the network for a given end-task.}\label{fig:least_activation}
\end{figure}

Considering the same example once again (Figure \ref{fig:least_activation}), we collect the activation data for the sequence, "What is your name?" in a similar way as demonstrated in section \ref{sec:most-activation}. However, instead of looking through the activation matrix for n most activated neuron, we prune n least activated neurons this time. \newline

Depending upon the value of n (2 in the example provided), we choose n neurons with the least activation (grey box) in the mass activation matrix and prune the knowledge within them, from the pre-trained model before transfer. Similar to the section \ref{sec:most-activation}, we perform this at every step of training until we reach our final end-task, i.e. English-Spanish translations. We argue that by pruning the least activated neurons, we are removing the knowledge from neurons that were least reactive for the given task. This should help transfer to effectively distribute new knowledge in the pruned neurons while keeping the effects of catastrophic forgetting to a bare minimum. 

\pdfminorversion=4

\chapter{Results} 

\label{Chapter5} 

\section{Network Architectures}
To have a baseline model, we train all the architectures mentioned in section \ref{sec:architectures} to perform English-English, and English-German, English-French, and English-Spanish translations from [\textbf{\cite{TiedemannThottingal:2020}}] and [\textbf{\cite{TiedemannN04}}] datasets respectively. The results in table \ref{tab:end2end}, help us in identifying the best performing model for all our datasets. The details about the hyperparameters used while training is mentioned in table \ref{tab:hyperparameters}.

\begin{table}[h!]
  \begin{center}
    \caption{\textbf{BLEU-4 cumulative scores for End-to-End architectures: } Performance of Long Short-Term Memory (LSTM), Gated-Recurrent Units (GRU), Attention-based Bidirectional Gated Recurrent Units (A-BGRU) neural architectures on English-English (En-En), English-German (En-De), English-French (En-Fr), and English-Spanish (En-Es) test datasets. }
    \label{tab:end2end}
    \begin{tabular}{l|c|c|c|r} 
      \textbf{Architectures} & \textbf{En-En} & \textbf{En-De} & \textbf{En-Fr} & \textbf{En-Es}\\
      \hline
      LSTM & 0.9191 & 0.1356 & 0.1642 & 0.1471\\
      GRU & 0.9264 & 0.1432 & 0.1653 & 0.1578\\
      A-BGRU & 0.9837 & 0.2013 & 0.2318 & 0.2263\\
    \end{tabular}
  \end{center}
\end{table}

All the architectures in table \ref{tab:end2end} are trained in end-to-end fashion described in section \ref{sec:end_to_end}. As can be observed in the table \ref{tab:end2end}, the Attention-based Bidirectional Gated Recurrent Units (A-BGRU) architecture performs best on all our datasets. Therefore, for all of our further research, we only use A-BGU as our standard architecture. 

\section{Baseline Scores}
\label{results:baseline}
As mentioned in section \ref{sec:baselines}, we perform end-to-end and 1-hop transfer training to establish a benchmark for the transfer models described in section \ref{sec:knowledge_transfer_training}. To give a quick recap, we perform translations in end-to-end, i.e., between the source and target languages directly (Figure \ref{fig:end2end}) without any transfer. Whereas, in 1-hop transfer training, we first pre-train our translation model to perform English-English translations. We then fine-tune this pre-trained model to perform English-German, English-French, or English-Spanish translations in transfer-learning fashion (Figure \ref{fig:pretraining_architecture} and \ref{fig:1-hop-transfer}). \newline

\begin{table}[h!]
  \begin{center}
    \caption{\textbf{BLEU-4 cumulative scores for baselines architectures} : Performance of End-to-End and 1-hop transfer baselines, evaluated on English-English (En-En), English-German (En-De), English-French (En-Fr), and English-Spanish (En-Es) test datasets. }
    \label{tab:baselines}
    \begin{tabular}{l|c|c|c|r} 
      \textbf{Baselines} & \textbf{En-En} & \textbf{En-De} & \textbf{En-Fr} & \textbf{En-Es}\\
      \hline
      End-to-End & 0.9837 & 0.2013 & 0.2318 & 0.2263\\
      1-hop transfer & - & 0.2133 & 0.2256 & 0.2181\\
    \end{tabular}
  \end{center}
\end{table}

As can be observed, the results of end-to-end training are better than 1-hop-transfer training for French and Spanish datasets. We argue that this is due to the significantly smaller size of our fine-tuning datasets. Whereas, for the German dataset, our 1-hop baseline beats the end-to-end score. This is due to English and German language sharing the same (Germanic) roots. However, due to experimenting on an extremely low-resource dataset, our pre-trained model is not properly able to adapt the knowledge from the fine-tuning dataset. Note we train the end-to-end baselines without sharing the vocabulary between the datasets. So, we always expected a slight difference in the performances between our two baselines. Regardless, we have used both the baselines mentioned as separate benchmarks to compare results in our further research.  

\section{RQ1: Knowledge-Transfer between different languages}
\label{sec:transfer_scores}
As described in section \ref{sec:knowledge_transfer_training}, we train our multilingual neural machine translation system in two different fashions, i.e. Multi-task and Sequential. In Joint multi-task transfer learning, we first pre-train our model to perform English-English translations (Figure \ref{fig:pretraining_architecture}). In the second step, we fine-tune the pre-trained model on the combined training datasets of English-German, English-French, and English-Spanish corpora (Figure \ref{fig:parallel_transfer_architecture}). For evaluation, we test the trained model on the combined and individual test dataset separately (table \ref{tab:parallel_and_sequential}). \newline

\begin{table}[h!]
  \begin{center}
    \caption{\textbf{BLEU-4 cumulative scores for joint multi-task and sequential-transfer learning setup} : Performance evaluated on English-English (En-En), combined (En-De + En-Fr + En-Es), and individual English-German (En-De), English-French (En-Fr), and English-Spanish (En-Es) test datasets.}
    \label{tab:parallel_and_sequential}
    \begin{tabular}{l|c|c|c|c|r} 
    \textbf{} & \textbf{} & \textbf{En-De + En-Fr} & \textbf{} & \textbf{} & \textbf{}\\
      \textbf{Transfer} & \textbf{En-En} & \textbf{+ En-Es} & \textbf{En-De} & \textbf{En-Fr} & \textbf{En-Es}\\
      \hline
      Multi-task & 0.9837 & 0.0784 & 0.0881 & 0.0903 & 0.0457 \\
      Sequential & 0.9837 & - & 0.2133 & 0.2055 & 0.2308 \\
    \end{tabular}
  \end{center}
\end{table}

Similar to the multi-task transfer learning setup, in sequential-transfer learning, we pre-train our model to perform English-English translations (Figure \ref{fig:pretraining_architecture}). However, instead of combining the datasets and fine-tuning on it, we apply knowledge-transfer one after another, in a sequential manner (Figure \ref{fig:de_transfer}, \ref{fig:fr_transfer}, and \ref{fig:es_transfer}). \newline

The results in table \ref{tab:parallel_and_sequential} demonstrates the performance of models trained in multi-task and sequential transfer-learning fashion. We argue that despite the understanding of source language from the pre-trained model, it is too challenging for the multi-task transfer model to perform multilingual translations solely on 40k sentences, shared between English-German, English-French, and English-Spanish corpora. Additionally, the knowledge-transfer from every language together fluctuates the gradient too much, making the learning even harder. On the other hand, freezing weights of the pre-trained model at every step, like in sequential-transfer learning setup, enables our model to utilize knowledge-transfer better. As can be seen, the performance of our sequential-transfer learning for English-German and English-French is initially lower than our baselines mentioned in table \ref{tab:baselines}. However, by the end of our setup, i.e. for English-Spanish, we see a significant increase in the performance. We argue that unlike German and French, French and Spanish have similar roots, and, therefore, the increase in performance is due to better transfer between French to Spanish. \newline

We also train four additional models in sequential transfer learning fashion, to examine the order of transfer that provides us with the best performance. Note that we pre-train all of our models on English-English translations for two major reasons :
\begin{itemize}
    \item We argue that it essential for our multilingual translation system to have an understanding of source language.
    \item In practise, pre-training is always performed on the largest available dataset. 
\end{itemize}

\begin{table}[h!]
  \begin{center}
    \caption{\textbf{BLEU-4 cumulative scores for sequential-transfer learning setup} : Performance of networks trained in different language-transfer orders, evaluated on English-English (En-En), combined (En-De + En-Fr + En-Es), and individual English-German (En-De), English-French (En-Fr), and English-Spanish (En-Es) test datasets.}
    \label{tab:transfer_order}
    \begin{tabular}{l|c|c|c|r} 
    \textbf{} & \textbf{} \textbf{} & \textbf{} & \textbf{}\\
      \textbf{Transfer Order} & \textbf{En-En} & \textbf{En-De} & \textbf{En-Fr} & \textbf{En-Es}\\
      \hline
      En-De-Fr-Es & 0.9837 & 0.2133 & 0.2055 & 0.2308 \\
      En-De-Es-Fr & 0.9837 & 0.2133 & 0.1846 & 0.2002 \\
      En-Fr-Es-De & 0.9837 & 0.1932 & 0.2256 & 0.2019 \\
      En-Es-Fr-De & 0.9837 & 0.1932 & 0.2005 & 0.2181\\
    \end{tabular}
  \end{center}
\end{table}

Note that the Tatoeba Challenge Dataset, [\textbf{\cite{TiedemannThottingal:2020}}], has training and test instances from OPUS and Tatoeba corpus respectively. Unlike most of current machine translation research, we evaluate our model over 10k sentences from a different corpus. Therefore, despite the smaller test to training dataset ratio (10k/40k, i.e. 1/4), our model shows promising results in terms of generalization. As can be seen in table \ref{tab:transfer_order}, we achieve the best performance with En-De-Fr-Es language-transfer order. Therefore, for all our further research, we only perform experiments on En-De-Fr-Es sequential-transfer learning setup. 

\section{RQ2: Effects of Selective Pruning}
\label{results:pruning}
Since we train our network in a low-resource setting, i.e. with one layer and 512 hidden units, the catastrophic forgetting gets more and more severe with every transfer. Therefore, we perform various pruning experiments to examine the effects of catastrophic forgetting on our neural machine translation system. As mentioned in section \ref{sec:pruning}, we perform mass-activation based neuron-knowledge pruning using three different approaches:

\subsection{Pruning dead neurons from the max mass-activation matrix}

Following the experiment in section \ref{sec:max-activation}, we prune the dead neurons from the max mass-activation matrix (Figure \ref{fig:max_activation}). Inspired by the experiments performed by [\textbf{\cite{rethmeier2019txray}}], we expected that pruning dead neurons, i.e. the neurons that were never activated or got deactivated in the mass-activation matrix, should improve generalization and reduce the effects of noise. The state-of-the-art (SOTA) results in table \ref{tab:max_activation_results} represents the performance of our sequential-transfer learning setup without pruning, whereas the pruning results represent the performance of the same setup after pruning, based on the experiments performed in section \ref{sec:max-activation}. Since we freeze the weights of the pre-trained model at every stage of the training,  pruning neuron-knowledge does not affect the performance of the pre-trained model at any stage.  Also, by pruning neuron-knowledge, we only deactivate the knowledge (weights) in a particular neuron. Therefore, the connections to other neurons and their weights do not concern us. As can be seen, there is a decrease in performance for English-German, English-French, and English-Spanish translations. \newline

\begin{table}[h!]
  \begin{center}
    \caption{\textbf{BLEU-4 cumulative scores after pruning dead neurons} : Performance of sequential-transfer learning setup after pruning dead neurons, based on max-mass activation matrix.The evaluation is carried on English-English (En-En), English-German (En-De), English-French (En-Fr), and English-Spanish (En-Es) test datasets.}
    \label{tab:max_activation_results}
    \begin{tabular}{l|c|c|c|r} 
      \textbf{} & \textbf{En-En} & \textbf{En-De} & \textbf{En-Fr} & \textbf{En-Es}\\
      \hline
      SOTA & 0.9837 & 0.2133 & 0.2055 & 0.2308 \\
      pruning dead-neurons & 0.9837 & 0.0535 & 0.0638 & 0.0865 \\
    \end{tabular}
  \end{center}
\end{table}

We argue that despite our reasonable expectation, we were only performing binary text classification in [\textbf{\cite{rethmeier2019txray}}]. By pruning neurons, we forced our network to generate sparsed connections. However, in our multilingual neural machine translation system, each end-task is comprised of an N-class classification problem, where N represents the vocabulary of a particular target language. Sequentially applying transfer for different end-tasks only makes the catastrophic forgetting worse. In other words, we conclude that pruning based on max mass-activation does not apply to an N-class classification problem. Furthermore, the continuous decrease in the performance suggests the loss in knowledge-transfer at every stage of training. 

\subsection{Pruning most-$n$ activated neuron-knowledge from the mass-activation matrix}
\label{res:mostn}
As suggested in section \ref{sec:most-activation}, instead of pruning on max mass-activations, we now prune n-neurons with most activated neuron-knowledge (aka activation potential) from the mass activation matrix (Figure \ref{fig:most_activation}). In other words, instead of considering just the max-activations, we compute the mass activation matrix by index wise summing all the activation potentials for every neuron, collected over the entire test datasets. 

\begin{table}[h!]
  \begin{center}
    \caption{\textbf{BLEU-4 cumulative scores after pruning 1\%, 5\%, and 10\% most activated neurons} : Performance of sequential-transfer learning setup after pruning 1\%, 5\%, and 10\% most activated neurons, based on mass activation matrix. The evaluation is carried on English-English (En-En), English-German (En-De), English-French (En-Fr), and English-Spanish (En-Es) test datasets.}
    \label{tab:mostn_activation_results}
    \begin{tabular}{l|c|c|c|c|r} 
      \textbf{Pruning \%} & \textbf{No. of neurons} & \textbf{En-En} & \textbf{En-De} & \textbf{En-Fr} & \textbf{En-Es}\\
      \hline
      0\% (SOTA) & 0 & 0.9837 & 0.2133 & 0.2055 & 0.2308 \\
      1\% & 5 & - & 0.1915 & 0.1924 & 0.1808 \\
      5\% & 25 & - & 0.1918 & 0.2146 & 0.1464 \\
      10\% & 51 & - & 0.1096 & 0.0960 & 0.0452 \\
    \end{tabular}
  \end{center}
\end{table}

Table \ref{tab:mostn_activation_results} shows the effects of pruning 1\%, 5\%, and 10\% pruning rate, i.e. 5, 25, and 51 neurons with most-n activated neuron-knowledge, on the system's performance. We then compare these results against our SOTA network to investigate the system's performance. Contrary to our hypothesis, none of the pruned networks beats the SOTA performance. As can be seen, we achieve the best performance by pruning 1\%, i.e. five neurons. \newline

The 0\% pruned network represents our state-of-the-art (SOTA) sequential-transfer learning network discussed in section \ref{sec:sequential_transfer}. We take the performance of this network as our new baseline for this experiment and compare it against the pruned networks. By pruning the neurons with most reactive neuron-knowledge, we are removing the specialized knowledge, and therefore, we expect this to increase generalization and reduce catastrophic forgetting. However, in comparison to the baseline, almost all the pruned networks show a consistent drop in performance between English-Spanish transfer (i.e. English-English -> English-German -> German-French -> French-Spanish). Since English and German, and French and Spanish have similar roots, we argue that continuously pruning the most activated neurons causes loss of crucial information which not only reduces generalized knowledge but also negatively affects the knowledge-transfer between these languages. \newline

Since German and French have different roots, pruning most-reactive neuron-knowledge from the 1\% and 5\% pruned-networks erase the specialized knowledge from the German language. As a result, there is an improvement in performance between English-German and English-French transfer. Pruning such specialized-knowledge not only improves the model's robustness but also helps in better generalization. Once again, since French and Spanish have common roots, the \% drop in performance between English-French and English-Spanish transfer suggests a loss in specialized knowledge between French to Spanish. However, for 10\% network, pruning knowledge from most activated 51 neurons almost takes most of the essential information required for the transfer. As can be seen in Table \ref{tab:mostn_activation_results}, continuous pruning of knowledge from these 51 neurons only results in a consistent drop in the performance of the network. In summary, we conclude that pruning most activated/specialized neuron-knowledge affects the performance of the system positively for languages with different roots, whereas pruning it for languages with similar roots affects it negatively.  

\subsection{Pruning least-\textbf{$n$} activated neuron-knowledge from the mass-activation matrix}
Contrary to the experiment performed in the section above \ref{res:mostn}, instead of pruning the most-$n$ activated neurons, we prune neuron-knowledge from the least-\textbf{$n$} activated neurons in this experiment (Figure \ref{fig:least_activation}). Once again, we compute the mass-activation matrix as suggested previously and look for neurons that constitute the least amount of neuron-knowledge content (aka activation potential). We argue that these neurons have the least influence on our system's performance, and therefore, pruning knowledge from them should increase model robustness and decrease catastrophic forgetting by enabling our model to learn new information, essential for the next steps. \newline

\begin{table}[h!]
  \begin{center}
    \caption{\textbf{BLEU-4 cumulative scores after pruning 1\%, 5\%, and 10\% least activated neurons} : Performance of sequential-transfer learning setup after pruning 1\%, 5\%, and 10\% least activated neurons, based on mass activation matrix. The evaluation is carried on English-English (En-En), English-German (En-De), English-French (En-Fr), and English-Spanish (En-Es) test datasets.}
    \label{tab:leastn_activation_results}
    \begin{tabular}{l|c|c|c|c|r} 
      \textbf{Pruning \%} & \textbf{No. of neurons} & \textbf{En-En} & \textbf{En-De} & \textbf{En-Fr} & \textbf{En-Es}\\
      \hline
      0\% (SOTA) & 0 & 0.9837 & 0.2133 & 0.2055 & 0.2308 \\
      1\% & 5 & - & 0.1915 & 0.1838 & 0.1983 \\
      5\% & 25 & - & 0.1935 & 0.1983 & 0.1641 \\
      10\% & 51 & - & 0.1993 & 0.1732 & 0.1432 \\
    \end{tabular}
  \end{center}
\end{table}

Table \ref{tab:leastn_activation_results} shows the effects of pruning neurons with least-\textbf{$n$} activated/reactive neuron-knowledge, on the system's performance. Once again, we prune three different networks for our experiment, with 1\%, 5\%, and 10\% pruning rate, i.e. 5, 25, and 51 neurons, respectively. We then compare these results against our SOTA network to examine our hypothesis. As can be seen, none of the pruned networks beats SOTA performance. The best results, however, are obtained by pruning 1\%, i.e. five neurons. Similar to section \ref{res:mostn}, the system's performance decreases as we increase the pruning \%. \newline

As can be seen, the English-German performance as we increase the pruning from 1\% to 10\%. Additionally, there is an increase in performance for French-Spanish transfer in the 1\% pruned model, which suggests an improvement in model robustness.  However, it decreases for the 5\% and the 10\% network. The continuous decrement in the performance of 5\% and 10\% network suggests that pruning any further only increases catastrophic forgetting that erases crucial information, required for better transfer at every step of training.  As indicated in section \ref{sec:transfer_scores}, the \% decrease in the performance from German to French is due to German and French languages having different roots. Despite the continuous decline in the performance, it is interesting to witness that our sequentially-transferred network is less sensitive to pruning least reactive than the most reactive neuron-knowledge. \newline

\pdfminorversion=4

\chapter{Inferences} 
\label{Chapter6} 

While we understand that activations inside the hidden layers of neural networks range between negative and positive values, there has been no research that shows the impact of these activations on the network's performance. In this chapter, we utilize various visualization techniques to understand the change in knowledge-abstractions, aka mass activation for every individual neuron, at different stages of our research. While the provided empirical results in sections \ref{results:baseline}, \ref{sec:transfer_scores}, and \ref{results:pruning} demonstrates the increase/decrease in the model's performance, they fail to explain how the neuron-knowledge was evolved, transferred, or erased during the training. Through this experiment, we examine the change in neuron-activations and its effects on the network's performance. 

\section{RQ3(A): Analyzing knowledge-abstractions and translation quality}
\label{sec:knowledge-abstraction}
Inspired by the research [\textbf{\cite{rethmeier2019txray}}], we use similar data accumulation technique as demonstrated in our original paper. We pass the source sequences, i.e. English sentences, to the Encoder of our Bidirectional Attention-based GRU network. As described in section \ref{sec:pruning}, the Encoder then generates the context vector in the form of hidden states, aka activation matrix. For a given source sequence with 5 tokens, we get a context vector of size (5x512), where 512 are the number of hidden units/neurons in our network. \newline

Following the same approach, we create a database from every input sequence in the test dataset. Additionally, we parse, tag, annotate, and record POS features for every token in each input sequence using Spacy [\textbf{\cite{spacy2}}]. Inspired by the visualizations generated in [\textbf{\cite{rethmeier2019txray}}], we generate neuron-distributions to quantify and compare knowledge-transfer within different models in section \ref{sec:neuron-feature-distributions}. \newline

\begin{figure}[h]
\centering
\includegraphics[width=\textwidth, keepaspectratio]{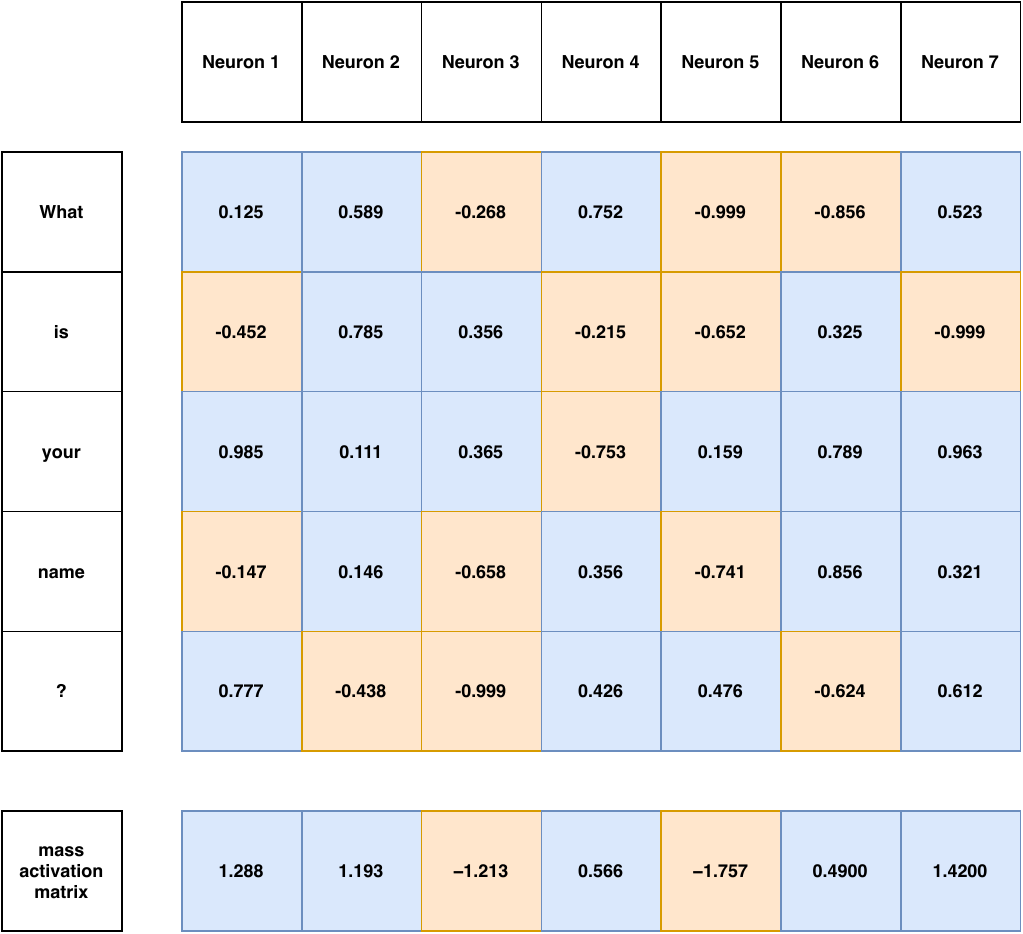}
\caption{\textbf{Positive and Negative activations} : Calculation of mass activation matrix from individual hidden states collected over the test datasets. All the positive and negative activations in the activation matrices are marked with blue and red colour respectively. Similarly, the neurons labelled with blue colour in the mass activation matrix are the positive neurons, whereas the labelled with red colour are negative neurons.  }\label{fig:mass_activation_viz}
\end{figure}

Figure \ref{fig:mass_activation_viz} demonstrates the calculation of mass activation matrix from individual hidden states, collected over the test dataset. Acknowledging the activation matrix as a mathematical vector and depending upon whether the activation potential of a neuron is positive or negative, we add or subtract the activations for each neuron in the activation matrices. We call the computed activation matrix as the mass activation matrix. We calculate the overall positive-knowledge by adding all the positive activations in the mass activation matrix. Similarly, the negative-knowledge is the sum of all the negative activations in the mass activation matrix. In the mass activation matrix, we identify positive neurons (blue colour) as the neurons with a positive activation potential, whereas the negative neurons (red colour) with negative activation potential. Additionally, we represent the overall knowledge as the sum of all the positive and negative activations in the mass activation matrix. Through this research question, we examine the effects of these positive and negative neurons on model performance and catastrophic forgetting. Starting on now, we treat these activation potentials from the mass activation matrix as positive and negative knowledge-abstractions, accordingly. \footnote{The scope of this research is limited to understanding the effects of knowledge-transfer and pruning, on Multi-lingual Neural Machine Translation. Nevertheless, we also show how the evolution of positive and negative knowledge-representations, during the training of differnt End-to-End, 1-hop, Multi-task, and Sequential-transfer architectures in appendix \ref{AppendixA}. } \newline

\subsection{Analyzing knowledge-abstractions for the transfer models}
\subsubsection{En-De Transfer}

\begin{figure}[h] 
12

\centering
\includegraphics[width=\textwidth, keepaspectratio]{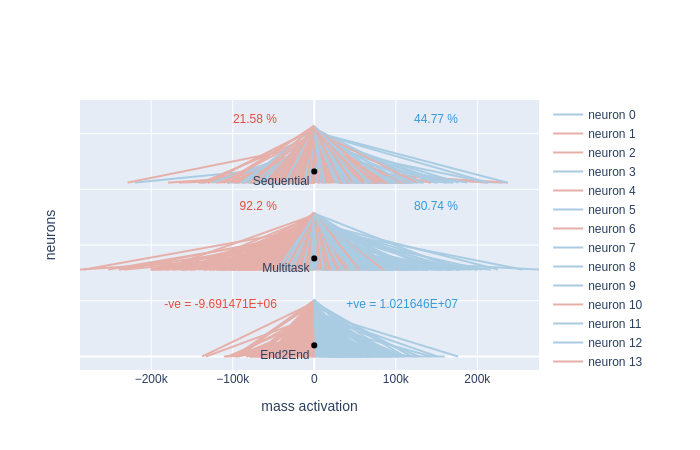}
\caption{\textbf{Positive and Negative knowledge-abstractions for English to German translations} : Positive (blue) and Negative (red) knowledge-abstractions for English to German translations in End2End, multi-task, and Sequential networks}\label{fig:en2de_comp}
\end{figure}

Through this experiment, we examine the change in positive and negative knowledge-abstractions between our end-to-end, 1-hop, multi-task, and sequential trained models. Figure \ref{fig:en2de_comp} shows the mass-activation based knowledge-abstraction plot for English to German translations, where we represent the positive and negative knowledge-abstractions with blue and red colour lines. On the y-axis, we have the 512 neurons, aka hidden units. On the x-axis, we have the positive or negative mass activation potential corresponding to an individual neuron on the y-axis. The colour of the neuron on the side labels represents whether the knowledge in that neuron is positive or negative. When plotted separately, this helps us to identify neurons that changed the most. \footnote{Please note that at no point we claim that negative-knowledge decreases and positive-knowledge increase the performance of a neural network. We solely make our interpretations based on the empirical results obtained in chapter \ref{Chapter5}, knowledge-abstraction plots, and translation quality. Please note that we present one good translation, i.e. Example 1, and one bad translation, i.e. Example 2, for all the cases.}  

\begin{table}[h!]
  \begin{center}
    \caption{Positive, Negative, and Overall-knowledge content in English to German translation models}
    \label{tab:en2de_comp}
    \begin{tabular}{l|c|c|r} 
     & \textbf{Positive} & \textbf{Negative} & \textbf{Overall} \\ 
     & \textbf{knowledge} & \textbf{knowledge} & \textbf{knowledge} \\ 
      \hline
     End-to-End & 1.021645E+07 & -9.697141E+06 & 5.249897E+05 \\  
     Multi-task & 1.846483E+07 & -1.862721E+07 & -1.623772E+05 \\  
     Sequential & 1.478999E+07 & -1.178285E+07 & 3.007141E+06
    \end{tabular}
  \end{center}
\end{table}

As can be seen in table \ref{tab:en2de_comp} and figure \ref{fig:en2de_comp}, the overall positive and negative-knowledge for our trained End2End model is 1.021645E+07 and -9.697141E+06, respectively. When added together, this sums up to an overall-knowledge of 5.249897E+05. For the multi-task and sequentially trained network, we only show the \%  increase or decrease in positive and negative-knowledge compared to our End2End model. There is an 80.74\% increase in positive-knowledge and 92.2\% in the negative-knowledge in the multi-task-transfer network. When added together, this sums up to a negative knowledge of -1.623772E+05. Whereas, there is a 44.77\% increase in positive-knowledge and 21.58\% in the negative-knowledge with an overall-knowledge of 3.007141E+06 in the sequential-transfer network.\newline

\begin{figure}[h]
\centering
\includegraphics[width=\textwidth, keepaspectratio]{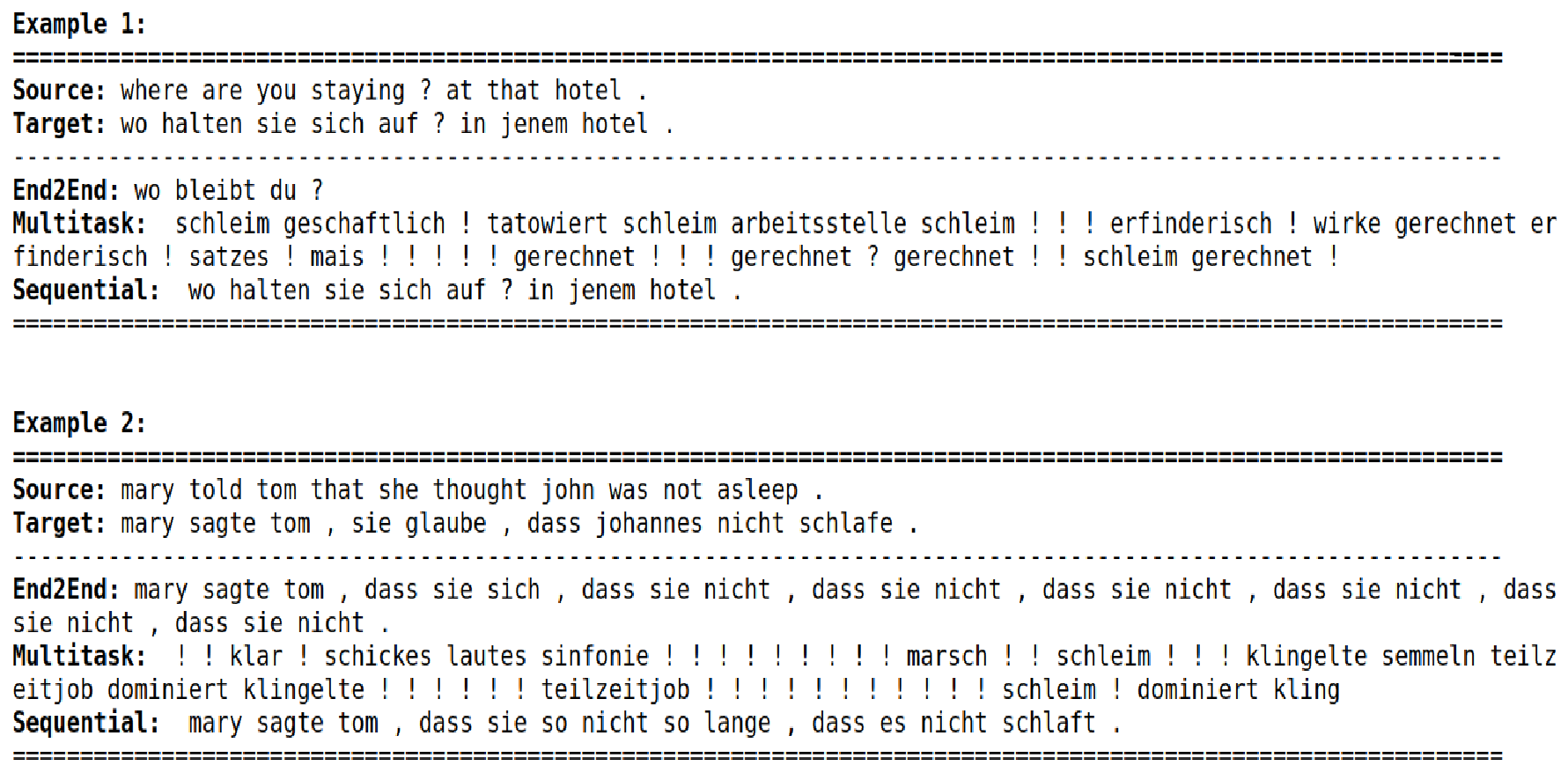}
\caption{English to German translations for End2End, multi-task, and Sequential networks}\label{fig:en2de_comp_translation}
\end{figure}

From tables \ref{tab:baselines} and \ref{tab:parallel_and_sequential}, we observe that the performance decreases in the case of multi-task transfer for English-German translations. At the same time, there is an increase in performance for the sequential-transfer network. We argue that this is due to a massive increase (92.2\%) of negative knowledge in the multi-task network. On the other hand, there is a slight increase in English to German translation quality for the sequential network. We argue that due to a steady increase in both negative and positive knowledge of the sequential network. Figure \ref{fig:en2de_comp_translation} shows the generated English-German translations for end-to-end, multi-task, and sequential network. As can be expected, the translation quality becomes poor for multi-task network whereas, it gets better for the sequential network. \newline

\subsubsection{De-Fr Transfer}
\begin{table}[h!]
  \begin{center}
    \caption{Positive, Negative, and Overall knowledge content in English to French translation models}
    \label{tab:en2fr_comp}
    \begin{tabular}{l|c|c|r} 
     & \textbf{Positive} & \textbf{Negative} & \textbf{Overall} \\ 
     & \textbf{knowledge} & \textbf{knowledge} & \textbf{knowledge} \\ 
      \hline
     End-to-End & 1.157816E+07 & -1.152872E+07 & 4.944777E+04 \\  
     1-hop & 1.562637E+07 & -1.342490E+07 & 2.201469E+06 \\ 
     Multi-task & 1.949042E+07 & -1.976563E+07 & -2.752049E+05 \\  
     Sequential & 1.242264E+07 & -1.384221E+07 & -1.419572E+06
    \end{tabular}
  \end{center}
\end{table}

\begin{figure}[h]
\centering
\includegraphics[width=\textwidth, keepaspectratio]{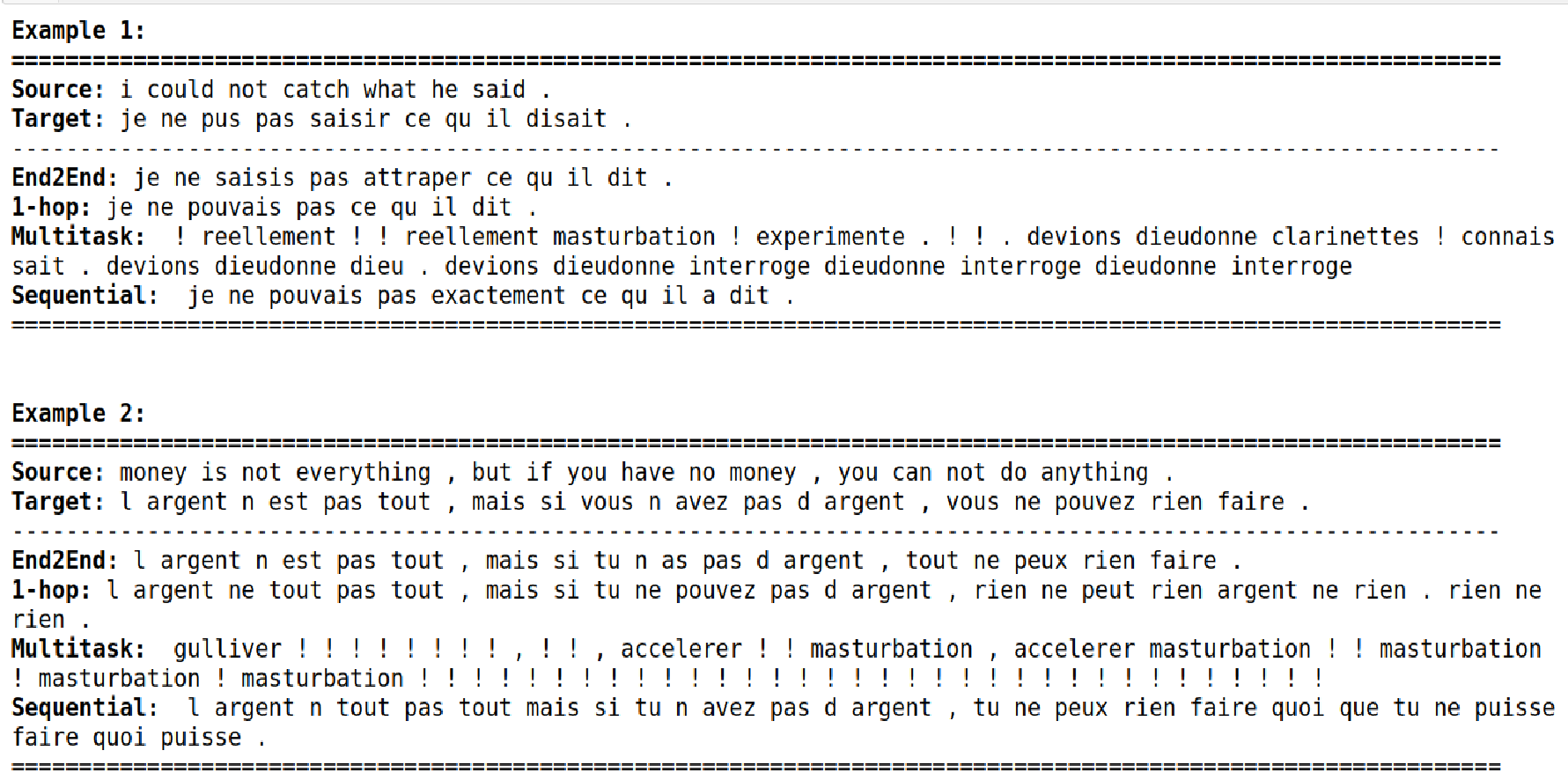}
\caption{English to French translations for End2End, 1-hop, multi-task, and Sequential networks}\label{fig:en2fr_comp_translation}
\end{figure}

\begin{figure}[h]
\centering
\includegraphics[width=\textwidth, keepaspectratio]{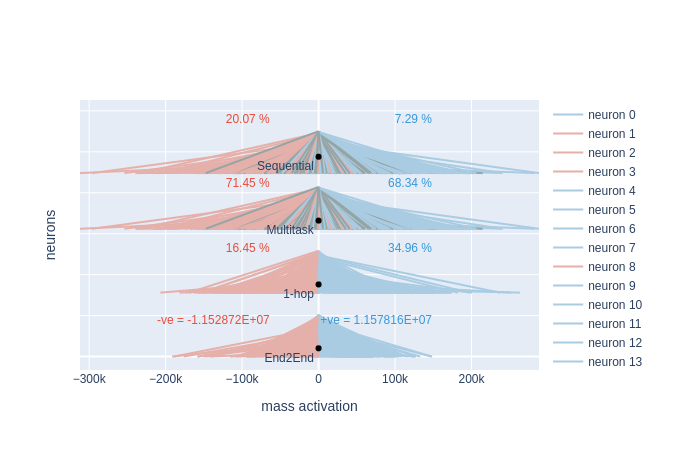}
\caption{\textbf{Positive and Negative knowledge-abstractions for English to French translations} : Positive   (blue)   and   Negative   (red)   knowledge-abstractions for English to French translations in End2End, 1-hop, multi-task,and Sequential networks}\label{fig:en2fr_comp}
\end{figure}

Similarly, table \ref{tab:en2fr_comp}, figure \ref{fig:en2fr_comp}, and \ref{fig:en2fr_comp_translation} shows the impact of positive and negative-knowledge on English to French translations. The grey lines indicate the neurons for which there is an increase in positive or negative-knowledge from the end-to-end baseline, but not 1-hop baseline. As mentioned in section \ref{sec:1-hop}, we use both 1-hop and End2End as our baselines for English-French and English-Spanish translations. As can be seen, there is a 34\% and 16.45\% increase in positive and negative-knowledge in the 1-hop baseline. Similarly, there is an increase of 68.34\% and 7.29\% in positive-knowledge for the multi-task and sequential networks. Whereas, there is an increase of 71.45\% and 20.07\% in negative-knowledge for the multi-task and sequential networks. \newline

For both the multi-task and sequential model, there is a massive increase in negative-knowledge compared to the positive-knowledge. Therefore, the overall knowledge decreases by a factor of -6.56\% and -29.70\% for Sequential and multi-task network, respectively. Once again, we argue that this decrease of knowledge in our sequential network is due to a low transfer from English-German to English-French model. From tables \ref{tab:baselines} and \ref{tab:parallel_and_sequential}, we observe that the performance decreases in the case of multi-task and sequential-transfer networks for English-French translations. Also, there is an increase in performance for the 1-hop network. We argue that the massive increase in positive-knowledge increase the overall-knowledge in the 1-hop network. In contrast, the massive increase in the negative-knowledge decrease in overall-knowledge in our multi-task and sequential networks. Note that there is a decrease in overall knowledge-transfer in the sequential network, from English-German (3.007141E+06) to English-French(-1.419572E+06). This is due a decrease in positive-knowledge transfer (from 1.478999E+07 to 1.242264E+07) and increase in negative-knowledge transfer (from -1.178285E+07  to -1.384221E+07). Once again, we argue that this decrease of knowledge-transfer in our sequential network is due to a low transfer from English-German to English-French model. \newline

Figure \ref{fig:en2fr_comp_translation} shows the generated English-French translations for end-to-end, multi-task, and sequential network. Similar to the performance of the networks, we achieve the best translation quality from the 1-hop baseline. Although we find the translation quality from the sequential-network better than our end-to-end baseline, the 1-hop translations are better. Unsurprisingly, the translation quality becomes worse after applying the multi-task transfer. \newline

\subsubsection{Fr-Es Transfer}
\begin{figure}[H]
\centering
\includegraphics[width=\textwidth, keepaspectratio]{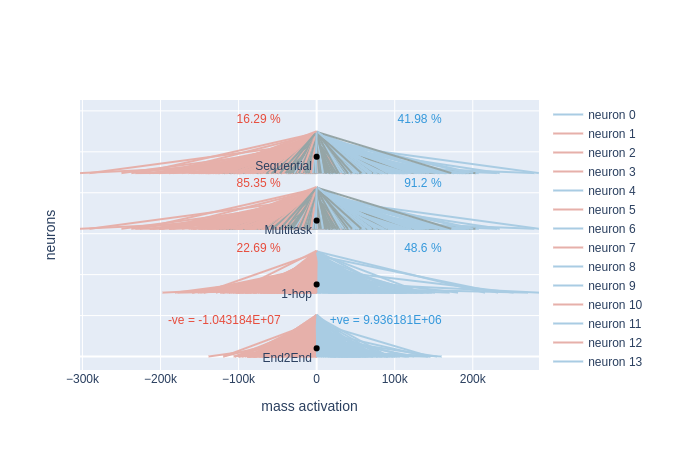}
\caption{\textbf{Positive and Negative knowledge-abstractions for English to Spanish translations}: Positive (blue) and Negative (red) knowledge-abstractions for English to Spanish translations for End2End, 1-hop, multi-task,and Sequential networks}\label{fig:en2es_comp}
\end{figure}

\begin{figure}[h]
\centering
\includegraphics[width=\textwidth, keepaspectratio]{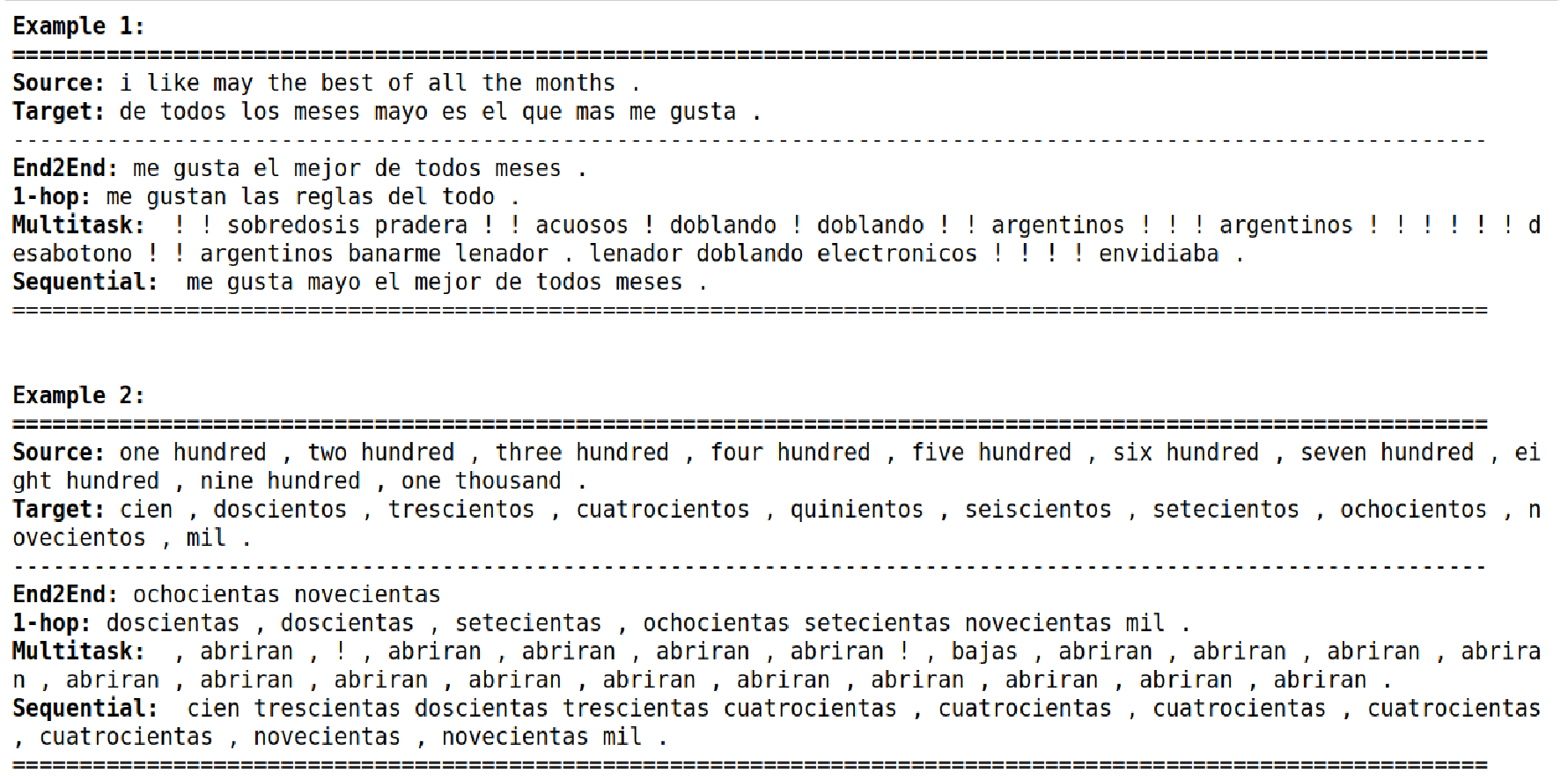}
\caption{English to Spanish translations for End2End, 1-hop, multi-task, and Sequential models}\label{fig:en2es_comp_translation}
\end{figure}

Table \ref{tab:en2es_comp}, figure \ref{fig:en2es_comp}, and \ref{fig:en2es_comp_translation} shows the impact of positive and negative-knowledge on English to Spanish translations. Once again, the grey neurons indicate an increase in positive or negative-knowledge from the end-to-end baseline, but not 1-hop baseline.  As can be seen, there is a 48.6\% and 22.69\% increase in positive and negative-knowledge in the 1-hop baseline. Similarly, there is an increase of 91.2\% and 41.98\% in positive-knowledge for the multi-task and sequential networks. Whereas, there is an increase of 85.35\% and 16.29\% in negative-knowledge for the multi-task and sequential networks, respectively. \newline

\begin{table}[h!]
  \begin{center}
    \caption{Positive, Negative, and Overall knowledge content in English to Spanish translation models}
    \label{tab:en2es_comp}
    \begin{tabular}{l|c|c|r} 
     & \textbf{Positive} & \textbf{Negative} & \textbf{Overall} \\ 
     & \textbf{knowledge} & \textbf{knowledge} & \textbf{knowledge} \\ 
      \hline
     End-to-End & 9.936181E+06 & -1.043184E+07 & -4.956634E+05 \\  
     1-hop & 1.476538E+07 & -1.279895E+07 & 1.966426E+06 \\ 
     Multi-task & 1.899803E+07 & -1.933551E+07 & -3.374808E+05 \\  
     Sequential & 1.410704E+07 & -1.213143E+07 & 1.975610E+06
    \end{tabular}
  \end{center}
\end{table}

As can be seen in table \ref{tab:en2es_comp}, the overall-knowledge in our end-to-end network is negative. This means the magnitude of negative-knowledge in English-Spanish translations is greater than the magnitude of the positive-knowledge. Therefore, despite the massive increase in positive-knowledge in the multi-task network, the overall-knowledge is still negative. Moreover, there is an increase of 0.46\% in overall knowledge-transfer when compared to our 1-hop model. Note that there is an increase in overall knowledge-transfer in the sequential network, from English-French  (-1.419572E+06) to English-Spanish (1.975610E+06). This is due an increase in positive-knowledge transfer (1.242264E+07 to 1.410704E+07) and decrease in negative-knowledge transfer (-1.384221E+07 to -1.213143E+07). Once again, we argue that this increase of knowledge in our sequential network is due to a high transfer from English-French to English-Spanish model. \newline

From tables \ref{tab:baselines} and \ref{tab:parallel_and_sequential}, we observe that compared to our end-to-end and 1-hop baselines, there is an increase in performance of the sequential-transfer networks for English-Spanish translations. Figure \ref{fig:en2es_comp_translation} shows the generated English-Spanish translations for end-to-end, multi-task, and sequential network. As can be seen, the translation quality increases with the performance of the network and therefore, we obtain the best translations through our sequential network.  

\subsection{Analyzing knowledge-abstractions for the pruned models}
Since we only prune neuron-knowledge from our sequential transfer learning setup, instead of using the end-to-end baseline, we compare our pruned results with the sequential-transfer setup. In section \ref{sec:knowledge-abstraction}, we define and show the effects of positive and negative knowledge on the model's performance. Together with the results shown in tables \ref{tab:mostn_activation_results} and \ref{tab:leastn_activation_results}, we attempt to understand the impacts of pruning this positive and negative neural-knowledge on model's performance and translation quality. Once again, we present one good translation, i.e. Example 1, and one bad translation, i.e. Example 2, for all the cases.

\subsubsection{En-De Transfer}
\begin{table}[h!]
  \begin{center}
    \caption{Pruned Positive, Negative, and Overall neuron-knowledge for the most-n activated neurons in English-German translation models}
    \label{tab:en2de_most_comp}
    \begin{tabular}{l|c|c|r} 
     & \textbf{Positive} & \textbf{Negative} & \textbf{Overall} \\ 
     & \textbf{knowledge} & \textbf{knowledge} & \textbf{knowledge} \\ 
      \hline
     Sequential & 1.478999E+07 & -1.178285E+07 & 3.007141E+06 \\  
     1\%-most pruned & 1.403722E+07 & -1.136065E+07 & 2.676574E+06 \\ 
     5\%-most pruned & 1.400047E+07 & -1.114715E+07 & 2.853314E+06 \\  
     10\%-most pruned & 1.343739E+07 & -1.189430E+07 & 1.543088E+06
    \end{tabular}
  \end{center}
\end{table}

\begin{figure}[H]
\centering
\includegraphics[width=\textwidth, keepaspectratio]{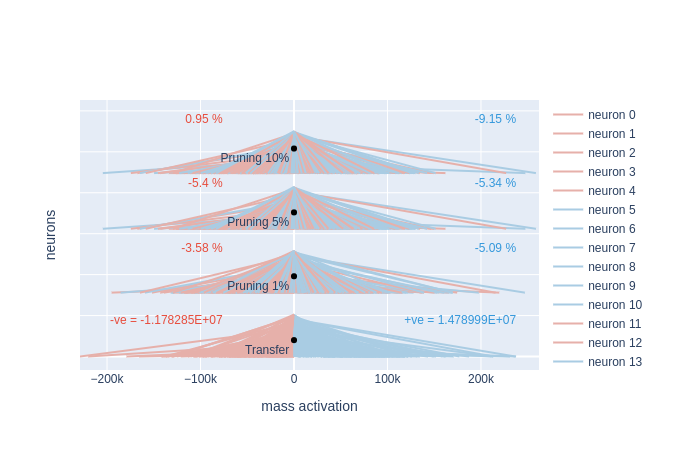}
\caption{\textbf{Positive and Negative knowledge-abstractions for English to German translations, after pruning of the most-n activated neuron-knowledge}: Positive (blue) and Negative (red) knowledge-abstractions for English to German translations in Sequential, 1\%-most pruned, 5\%-most pruned,and 10\%-most pruned networks.}\label{fig:en2de_most}
\end{figure}

\begin{figure}[H]
\centering
\includegraphics[width=\textwidth, keepaspectratio]{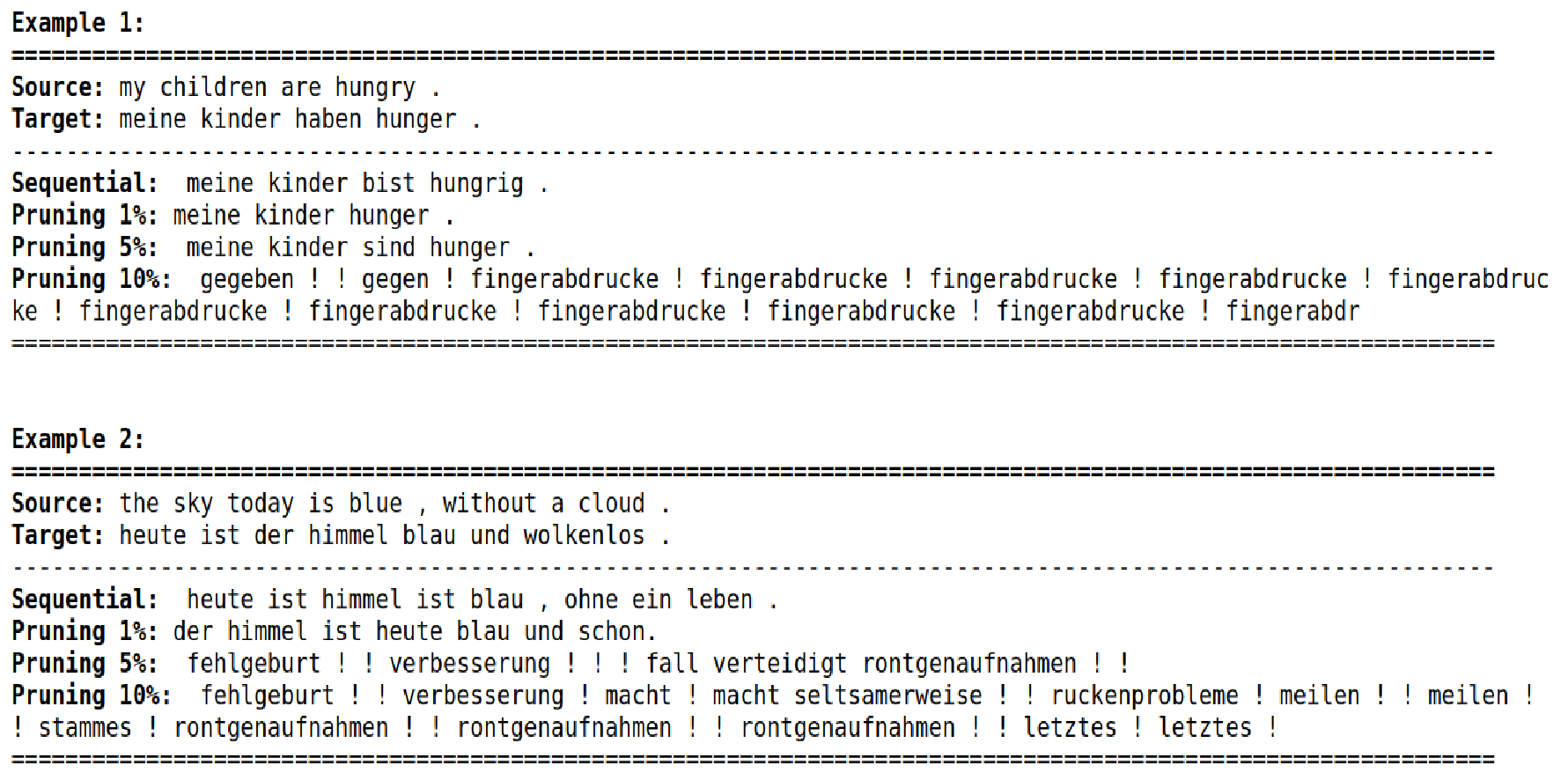}
\caption{English to German translations for Sequential, 1\%-most pruned, 5\%-most pruned,and 10\%-most pruned networks.}\label{fig:en2de_most_translation}
\end{figure}

Figure \ref{fig:en2de_most} shows the impact of pruning neuron-knowledge from the most activated $n$ neurons, on positive and negative knowledge abstractions, for English to German translations. As can be seen in table \ref{tab:en2de_most_comp}, there is a consistent decrease in positive-knowledge as we increase the pruning from 1\% to 10\%. On the other hand, while there is a decrease in negative-knowledge for 1\% and 5\%, it increases by 0.95\% for the 10\% pruned-network. \newline

As described in table \ref{tab:mostn_activation_results}, there is a slight increase in the performance for English-German translations from 1\% to 5\% pruned network. We argue that the increase in performance is the result of an increase in the overall-knowledge, from 2.676574E+06 in the 1\% network to 2.853314E+06 in the 5\% pruned-network. Figure \ref{fig:en2de_most_translation} shows the impact of pruning neuron-knowledge from the most activated neurons, on the English to German translation quality. As can be seen, the translation quality decreases as we increase the pruning \%. While the translations do improve from 1\% to 5\%, they are still not better than the sequential-transfer baseline. \newline

\begin{table}[h!]
  \begin{center}
    \caption{Pruned Positive, Negative, and Overall neuron-knowledge for the least activated neurons in English-German translation models}
    \label{tab:en2de_least}
    \begin{tabular}{l|c|c|r} 
     & \textbf{Positive} & \textbf{Negative} & \textbf{Overall} \\ 
     & \textbf{knowledge} & \textbf{knowledge} & \textbf{knowledge} \\ 
      \hline
     Sequential & 1.478999E+07 & -1.178285E+07 & 3.007141E+06 \\  
     1\%-least pruned & 1.111770E+07 & -1.059133E+07 & 5.263659E+05 \\ 
     5\%-least pruned & 1.225276E+07 & -9.397045E+06 & 2.855712E+06 \\  
     10\%-least pruned & 1.234896E+07 & -8.667709E+06 & 3.681250E+06
    \end{tabular}
  \end{center}
\end{table}

\begin{figure}[H]
\centering
\includegraphics[width=\textwidth, keepaspectratio]{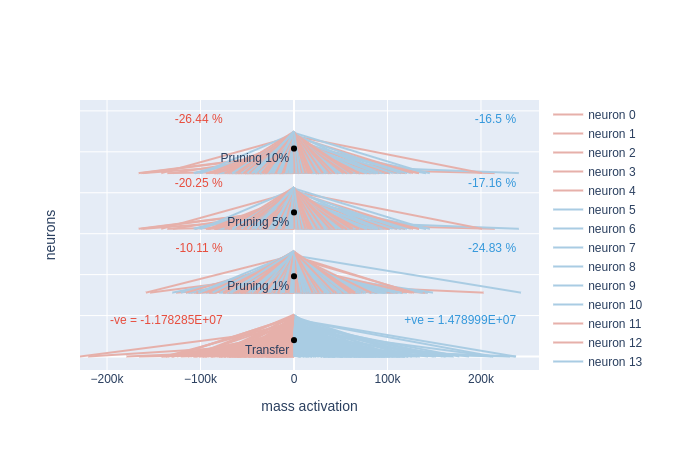}
\caption{\textbf{Positive and Negative knowledge-abstractions for English to German translations, after pruning of the least-n activated neuron-knowledge}: Positive (blue) and Negative (red) knowledge-abstractions for English to German translations in Sequential, 1\%, 5\%,and 10\%-least pruned networks.}\label{fig:en2de_least}
\end{figure}

Similarly, figure \ref{fig:en2de_least} shows the impact of pruning neuron-knowledge from the least-$n$ activated neurons, on positive and negative knowledge abstractions, for English to German translations. As can be seen, while there is a decrement in the decrease of positive-knowledge, the negative-knowledge tends to decrease more and more as we increase the pruning \%. We report similar observations through table \ref{tab:leastn_activation_results}, where the performance of our English-German translation model increases as we increase the pruning \%. As shown in table \ref{fig:en2de_least}, this is due to a consistent increase in overall-knowledge. \newline 

\begin{figure}[h]
\centering
\includegraphics[width=\textwidth, keepaspectratio]{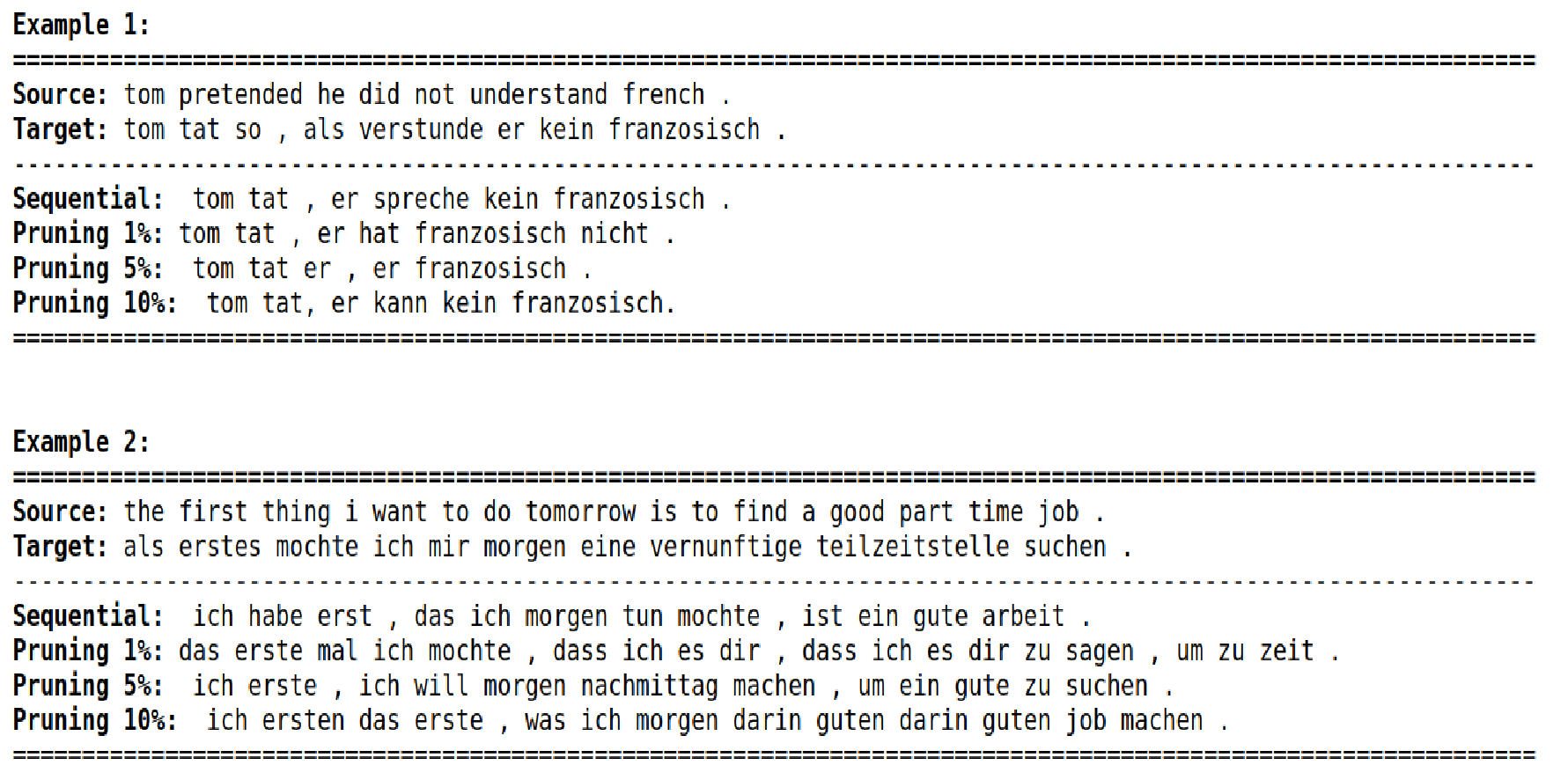}
\caption{English to German translations for Sequential, 1\%-least pruned, 5\%-least pruned,and 10\%-least pruned networks.}\label{fig:en2de_least_translations}
\end{figure}

Figure \ref{fig:en2de_least_translations} shows the translations obtained after pruning 1\%, 5\%, and 10\% neurons with least activate neuron-knowledge. As can be seen, the translation quality in Example 1 and Example 2 increase as we increase the pruning \% from 1\% to 10\%. Similar to the performance measured in table \ref{tab:leastn_activation_results}, none of the pruned translations is better than the sequential-transfer baseline.  \newline

\newpage
\subsubsection{De-Fr Transfer}
\begin{figure}[H]
\centering
\includegraphics[width=\textwidth, keepaspectratio]{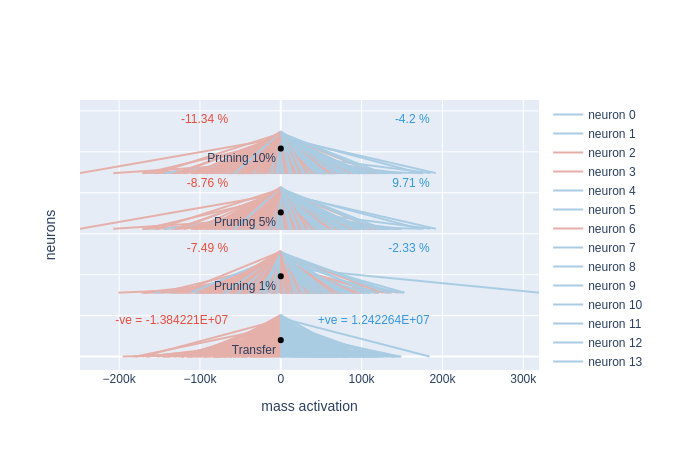}
\caption{\textbf{Positive and Negative knowledge-abstractions for English to French translations, after pruning of the most-n activated neuron-knowledge}: Positive (blue) and Negative (red) knowledge-abstractions for English to French translations in Sequential, 1\%, 5\%,and 10\%-most pruned networks.}\label{fig:en2fr_most}
\end{figure}

Figure \ref{fig:en2fr_most} and table \ref{tab:en2fr_most} shows the impact of pruning neuron-knowledge from the most-$n$ activated neurons, on positive and negative knowledge-abstractions, for English to French translations. As can be seen in 1\% pruned-network, there is a decrease in positive and negative-knowledge by a factor of 2.33\% and 7.49\%. For the 5\% pruned network, there is an increase of 9.71\% in the positive-knowledge and a decrease of 8.76\% in the negative-knowledge. Once again, for the 10\% network, there is a decrease in positive and negative-knowledge by a factor of 4.2\% and 11.34\%. \newline

\begin{table}[H]
  \begin{center}
    \caption{Pruned Positive, Negative, and Overall neuron-knowledge for the most activated neurons in English-French translation models}
    \label{tab:en2fr_most}
    \begin{tabular}{l|c|c|r} 
     & \textbf{Positive} & \textbf{Negative} & \textbf{Overall} \\ 
     & \textbf{knowledge} & \textbf{knowledge} & \textbf{knowledge} \\ 
      \hline
     Sequential & 1.242264e+07 & -1.384221e+07 & -1.419572e+06 \\  
     1\%-most pruned & 1.213377e+07 & -1.280577e+07 & -6.719998e+05\\ 
     5\%-most pruned & 1.362876e+07 & -1.263013e+07 & 9.986315e+05 \\  
     10\%-most pruned & 1.190130e+07 & -1.227286e+07 & -3.715609e+05
    \end{tabular}
  \end{center}
\end{table}

\begin{figure}[h]
\centering
\includegraphics[width=\textwidth, keepaspectratio]{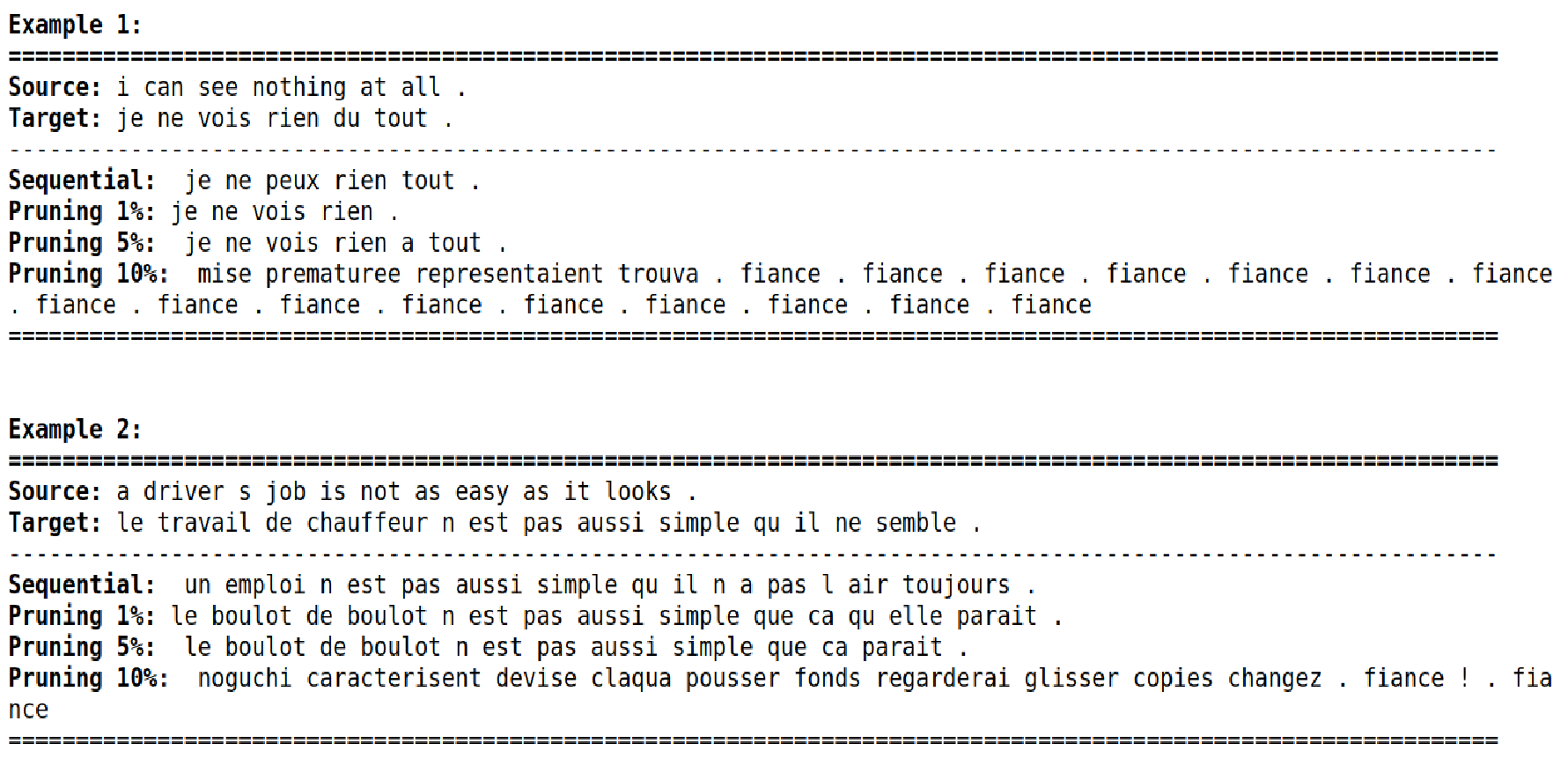}
\caption{English to French translations for Sequential, 1\%-most pruned, 5\%-most pruned,and 10\%-most pruned networks.}\label{fig:en2fr_most_translations}
\end{figure}

We argue that the increase in the overall-knowledge in the 5\% pruned network improves the performance of the model (table \ref{tab:mostn_activation_results}). Figure \ref{fig:en2fr_most_translations} shows the impact of positive and negative-knowledge on English to French translations. As can be seen, the translation quality slightly increases as we increase the pruning from 1\% to 5\%. Like we mentioned before, we argue that the increase in translation quality is the result of a massive increase in the overall-knowledge in the 5\% pruned model. After that, the translation quality drastically decreases as we increase the pruning any further from 5\% to 10\%.

\begin{table}[h!]
  \begin{center}
    \caption{Pruned Positive, Negative, and Overall neuron-knowledge for the least activated neurons in English-French translation models}
    \label{tab:en2fr_least}
    \begin{tabular}{l|c|c|r} 
     & \textbf{Positive} & \textbf{Negative} & \textbf{Overall} \\ 
     & \textbf{knowledge} & \textbf{knowledge} & \textbf{knowledge} \\ 
      \hline
     Sequential & 1.242264e+07 & -1.384221e+07 & -1.419572e+06 \\  
     1\%-least pruned & 1.657625e+07 & -1.755683e+07 & -9.805835e+05\\ 
     5\%-least pruned & 1.692873e+07 & -1.585295e+07 & 1.075784e+06 \\  
     10\%-least pruned & 2.011076e+07 & -2.003129e+07 & 7.947459e+04
    \end{tabular}
  \end{center}
\end{table}

\begin{figure}[H]
\centering
\includegraphics[width=\textwidth, keepaspectratio]{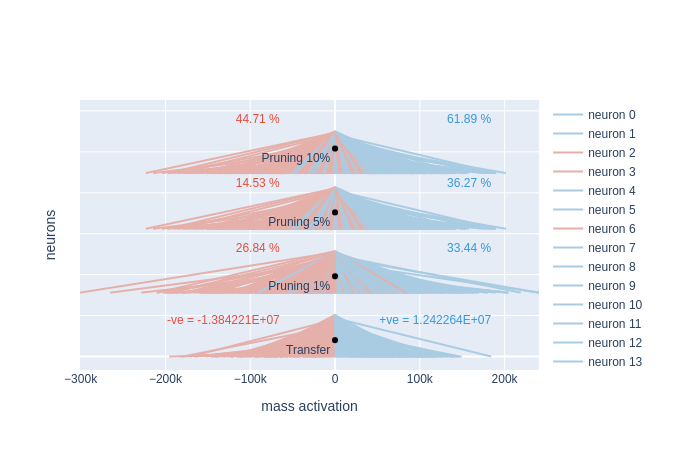}
\caption{\textbf{Positive and Negative knowledge-abstractions for English to French translations, after pruning of the least-n activated neuron-knowledge}: Positive (blue) and Negative (red) knowledge-abstractions for English to French translations in Sequential, 1\%, 5\%,and 10\%-least pruned networks.}\label{fig:en2fr_least}
\end{figure}

\begin{figure}[h]
\centering
\includegraphics[width=\textwidth, keepaspectratio]{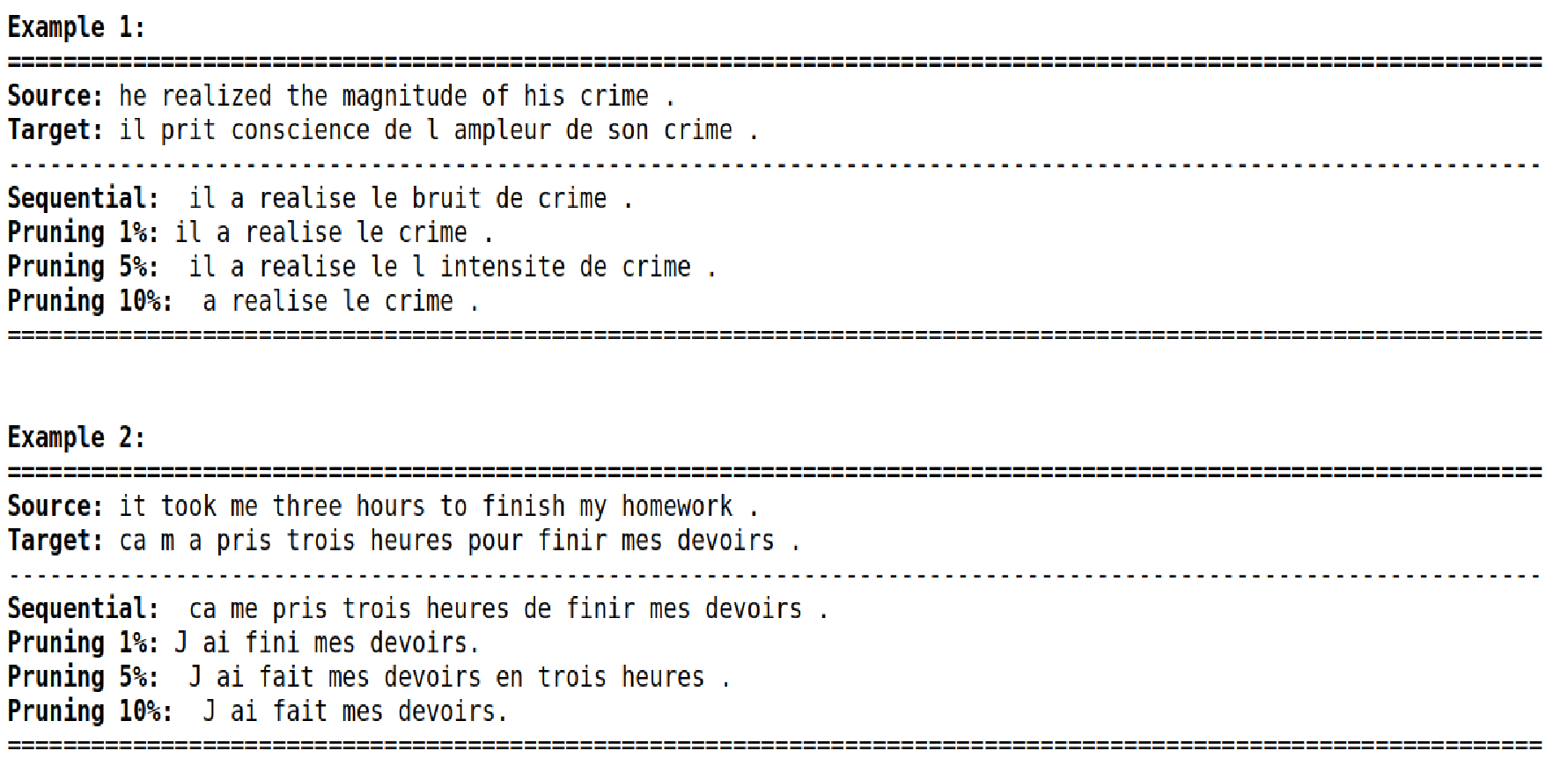}
\caption{English to French translations for Sequential, 1\%-least pruned, 5\%-least pruned,and 10\%-least pruned networks.}\label{fig:en2fr_least_translations}
\end{figure}

Similarly, figure \ref{fig:en2fr_least} and table \ref{tab:en2fr_least} shows the impact of pruning neuron-knowledge from the least-$n$ activated neurons, on positive and negative-knowledge for English to French translations. As can be seen, there is an increase in positive and negative-knowledge for all the pruned models. As can be seen in table \ref{tab:leastn_activation_results}, we obtain the best results with the 5\% pruned model. As can be seen, the overall knowledge for the 5\% pruned network increases from -1.419572e+06 to 1.075784e+06. \newline

Figure \ref{fig:en2fr_least_translations} shows the impact of pruning neuron-knowledge from the least-$n$ activated neurons, on the English to French translations. While there is no drastic change in translation quality with the increase in pruning \%, we obtain the best translations for the 5\% pruned network. However, none of the translations beat our Sequential transfer baseline. Also, we noticed that as we increase the pruning from 5\% to 10\%, the translation quality starts to decrease once again.


\subsubsection{Fr-Es Transfer}

\begin{figure}[h!]
\centering
\includegraphics[width=\textwidth, keepaspectratio]{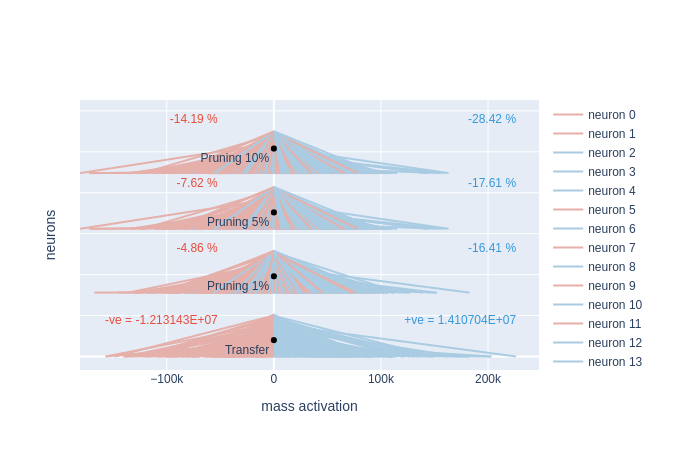}
\caption{\textbf{Positive and Negative knowledge-abstractions for English to Spanish translations, after pruning of the most-n activated neuron-knowledge}: Positive (blue) and Negative (red) knowledge-abstractions for English to Spanish translations in Sequential, 1\%, 5\%, and 10\%-most pruned networks.}\label{fig:en2es_most}
\end{figure}

\begin{table}[h!]
  \begin{center}
    \caption{Pruned Positive, Negative, and Overall neuron-knowledge for the most-$n$ activated neurons in English-Spanish translation models}
    \label{tab:en2es_most}
    \begin{tabular}{l|c|c|r} 
     & \textbf{Positive} & \textbf{Negative} & \textbf{Overall} \\ 
     & \textbf{knowledge} & \textbf{knowledge} & \textbf{knowledge} \\ 
      \hline
     Sequential & 1.410704e+07 & -1.213143e+07 & 1.975610e+06 \\  
     1\%-most pruned & 1.179226e+07 & -1.154243e+07 & 2.498330e+05\\ 
     5\%-most pruned & 1.162343e+07 & -1.120743e+07 & 4.159987e+05 \\  
     10\%-most pruned & 1.009849e+07 & -1.040980e+07 & -3.113079e+05
    \end{tabular}
  \end{center}
\end{table}

Figure \ref{fig:en2es_most} and table \ref{tab:en2es_most} shows the impact of pruning neuron-knowledge from the most-$n$ activated neurons, on positive and negative knowledge-abstractions, for English to Spanish translations. As can be seen, there is a consistent drop in positive and negative-knowledge as we increase the pruning from 1\% to 10\%. As mentioned before, we argue that continuously pruning the most active neuron-knowledge from English-English, English-German, and English-French translation models erased most of the essential information. \newline

\begin{figure}[H]
\centering
\includegraphics[width=\textwidth, keepaspectratio]{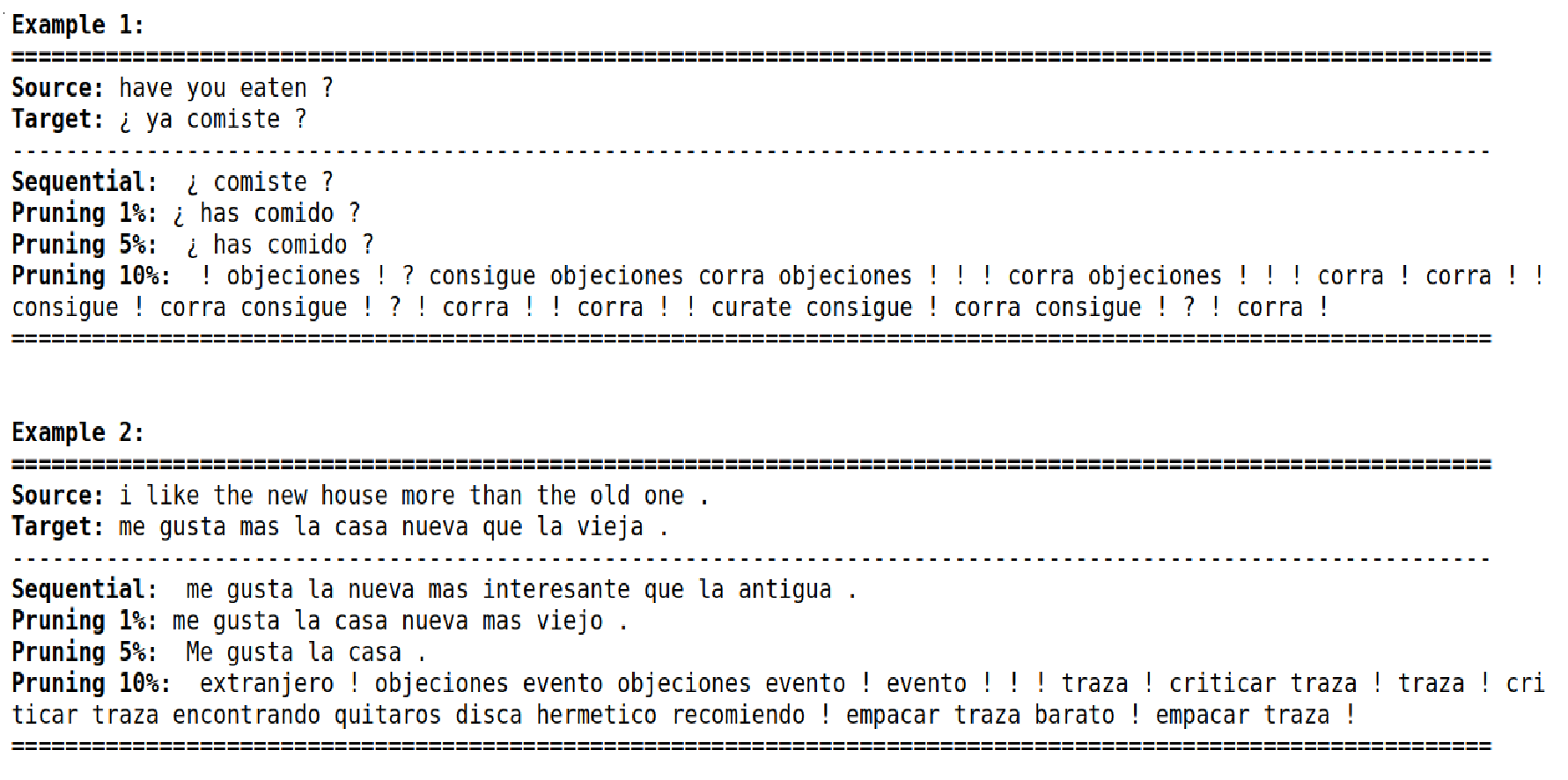}
\caption{English to Spanish translations for Sequential, 1\%-most pruned, 5\%-most pruned,and 10\%-most pruned networks.}\label{fig:en2es_most_translations}
\end{figure}

It is interesting to observe once again, that the best performing, i.e. 5\% pruned model, has the largest overall-knowledge. Figure \ref{fig:en2es_most_translations} shows the impact of pruning neuron-knowledge from the most activated neurons, on the English to Spanish translation quality. Once again, none of the pruned translations was better than our sequential-transfer baseline, and we obtain the best translations from the 5\% pruned-network.\newline

\begin{figure}[H]
\centering
\includegraphics[width=\textwidth, keepaspectratio]{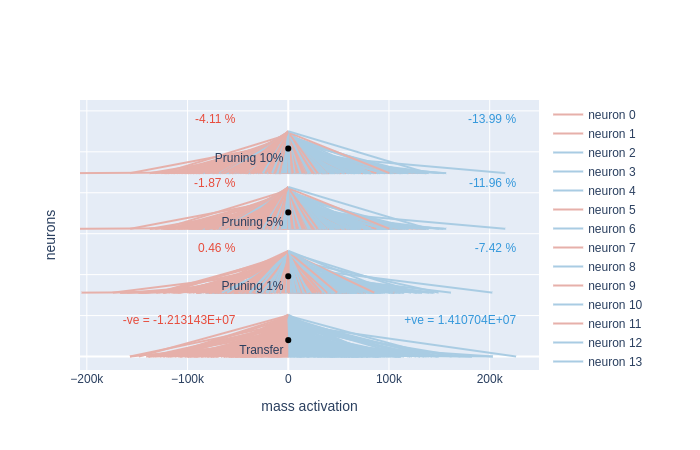}
\caption{\textbf{Positive and Negative knowledge-abstractions for English to Spanish translations, after pruning of the least-$n$ activated neuron-knowledge}: Positive (blue) and Negative (red) knowledge-abstractions for English to Spanish translations in Sequential, 1\%, 5\%, and 10\%-least pruned networks.}\label{fig:en2es_least}
\end{figure}

Similarly, figure \ref{fig:en2es_least} and table \ref{tab:en2es_least} shows the impact of pruning neuron-knowledge from the least-$n$ activated neurons, on positive and negative knowledge abstractions for English to Spanish translations. As can be seen, for the 1\% pruned model, there is a 7.42\% decrease in positive knowledge and 0.46\% increase in negative knowledge. However, for both the 5\% and 10\% pruned model, there is a decrease in both the positive and negative knowledge. Unsurprisingly, the overall-knowledge decreases as we increase the pruning from 1\% to 10\%.\newline

\begin{table}[h!]
  \begin{center}
    \caption{Pruned Positive, Negative, and Overall neuron-knowledge for the least activated neurons in English-Spanish translation models}
    \label{tab:en2es_least}
    \begin{tabular}{l|c|c|r} 
     & \textbf{Positive} & \textbf{Negative} & \textbf{Overall} \\ 
     & \textbf{knowledge} & \textbf{knowledge} & \textbf{knowledge} \\ 
      \hline
     Sequential & 1.410704e+07 & -1.213143e+07 & 1.975610e+06 \\  
     1\%-least pruned & 1.306030e+07 & -1.218670e+07 & 8.736050e+05\\ 
     5\%-least pruned & 1.241915e+07 & -1.190429e+07 & 5.148629e+05 \\  
     10\%-least pruned & 1.213277e+07 & -1.163301e+07 & 4.997626e+05
    \end{tabular}
  \end{center}
\end{table}

\begin{figure}[h]
\centering
\includegraphics[width=\textwidth, keepaspectratio]{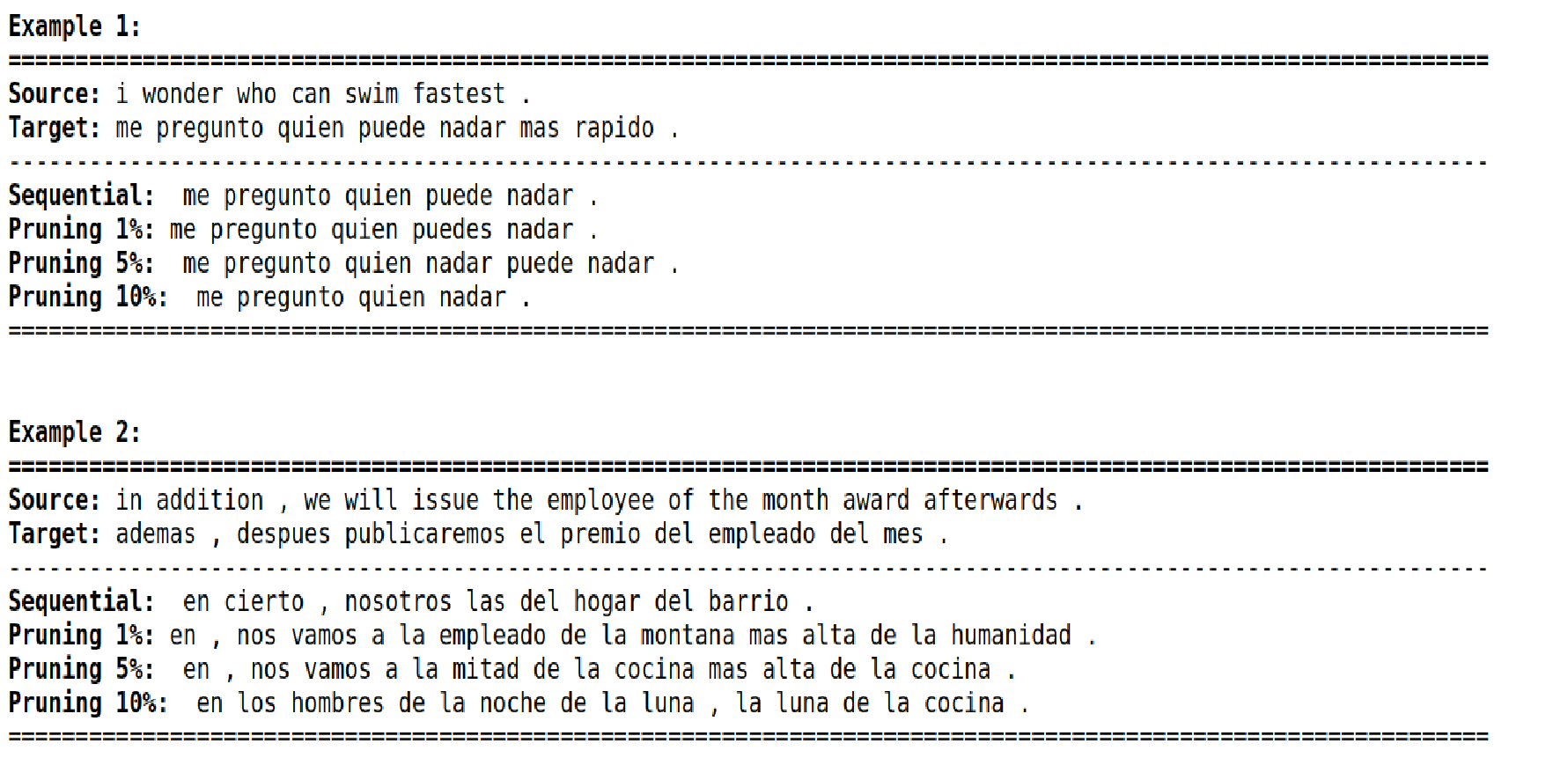}
\caption{English to Spanish translations for Sequential, 1\%-least pruned, 5\%-least pruned,and 10\%-least pruned networks.}\label{fig:en2es_least_translations}
\end{figure}

Figure \ref{fig:en2es_least_translations} shows the impact of pruning neuron-knowledge from the least-$n$ activated neurons, on the English to Spanish translation quality. Once again, none of the pruned translations was better than our sequential-transfer baseline. As can be seen, the translations only get worse as we increase the pruning from 1\% to 10\%. \newpage

\section{RQ3(B) : Interpreting knowledge-transfer through POS-distributions in the sequential-transfer network}
\label{sec:neuron-feature-distributions}
So far, we've talked about the impact of knowledge-transfer and pruning techniques on our system's performance and translation quality. Inspired by the POS-token-distribution visualizations in [\textbf{\cite{rethmeier2019txray}}], that showed the knowledge-transfer from Unsupervised and Zero-shot to the Supervised domain, we demonstrate the knowledge-transfer from one language to another, in our Sequential-transfer learning setup. \footnote{Please note that due to the computation reasons, we only produce plots for our best performing setup, i.e. the sequential-transfer network. We encourage our readers to take inspiration from these POS-token-distributions and explore different transfer-settings.} \newline

In our original work [\textbf{\cite{rethmeier2019txray}}], we passed the input token through our LSTM Encoder to produce a hidden vector, aka the activation matrix. Despite utilizing the entire activation matrix, we tagged each input token to a max-activation and its associated neuron. Using max-activation not only reduced the computation of our analysis but also helped us in calculating the Hellinger Distance (the symmetric version of the Kullback Leibler divergence), between neurons of two different setups. This allowed us to identify the neurons that changed the most and the ones that changed least, after application of zero-shot and supervision. However, in this research, we argue that since Multi-lingual Neural Machine Translation is an N-class classification problem, forcing the network to create sparse connections only decreases our system's performance. Nevertheless, methods like Hellinger Distance can only compute the distance between two probability distributions. In this research, however, we focus on interpreting knowledge-transfer using the entire-activation matrix, i.e. with positive and negative values. Therefore, we use the mass-activation matrix as the measure to quantify the differences between two discrete feature-preference distributions. To calculate the change in neuron-knowledge, we index wise-subtract the mass-activation matrix, before and after transfer. We call this new matrix as the change-in-mass-activation matrix.
In this section, we only show neurons that have the most and least activation potential in the change-in-mass-activation matrix.  \newline

On the Y-axis, we represent the normalized-activation potentials of the activated input(token)-features, whereas, on the X-axis, we describe the pos-classes. To avoid repetition of similar input-features in our POS-token-distribution, we take the average of all duplicate tokens and merge them into one. Therefore, we assign each unique input-feature with an average-activation potential and a POS-class category. For visibility reasons, we only show the most-5 activated input features for any described neuron. \footnote{Note that we use Plotly as a framework to generate our plots, and therefore, we zoom and investigate each input-feature distinctly. However, due to the limited scope of these plots in the thesis, we only show the most-5 activated input-features.} The bar plots represent the POS-density from all the positively-activated input-features in a specific neuron. The scatter plot, on the other hand, shows the unique input-features belonging to a particular POS-class with a distinct normalized-activation potential on Y-axis. \newline

\subsection{En-De Transfer}
\begin{figure}[h]
\centering
\includegraphics[width=\textwidth, keepaspectratio]{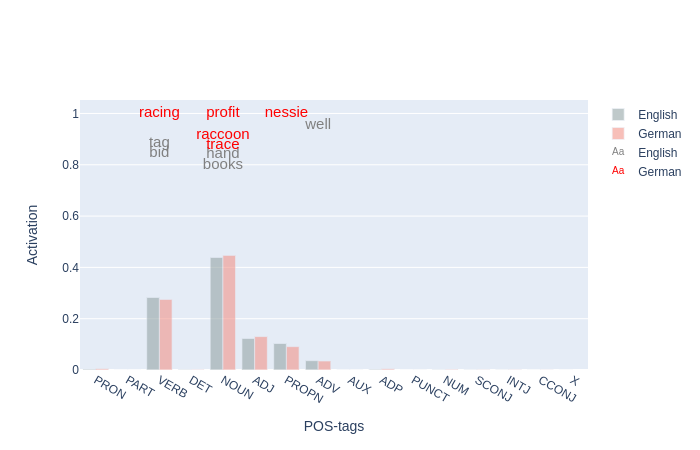}
    \caption{\textbf{Neuron-322: English (En) to German (De) Transfer} - The figure shows the pos-token-distribution for neuron:322 with the most change in neuron-knowledge between English (Grey color) to German (red color) transfer.}\label{fig:en2de_most_322}
\end{figure}


\begin{table}[h!]
  \begin{center}
    \caption{Change in neuron-knowledge for neuron-322, after English to German transfer}
    \label{tab:en2de_322}
    \begin{tabular}{l|c|c|r} 
     & \textbf{Input-} & \textbf{POS-} & \textbf{Mass-} \\ 
     & \textbf{features} & \textbf{class} & \textbf{activation} \\ 
      \hline
     Before transfer & 9579 & 16 & 1128.8164 \\  
     After transfer & 4750 & 15 & -3.4117 \\ 
    \end{tabular}
  \end{center}
\end{table}

Figure \ref{fig:en2de_most_322} shows the pos-token-distribution for neurons-322 that changed the most, during English to German transfer. It is interesting to observe how neuron-knowledge from the English language (tag, bid, hand, books, and well) in the neuron-322, got replaced with the new German knowledge (racing, profit, raccoon, trace, and nessie). Table \ref{tab:en2de_322}, on the other hand, shows the change in neuron-knowledge of neuron-322 for English to German transfer, in terms of activated input-features, POS-classes, and mass-activation. As can be seen, there is a decrease in the number of activated input-features and POS-class after the transfer, which suggests an increase in specialized knowledge for neuron-322. There is a huge drop in the mass-activation of the neuron implies a decrease in knowledge-transfer and an increase in catastrophic forgetting. \newline

\begin{figure}[h]
\centering
\includegraphics[width=\textwidth, keepaspectratio]{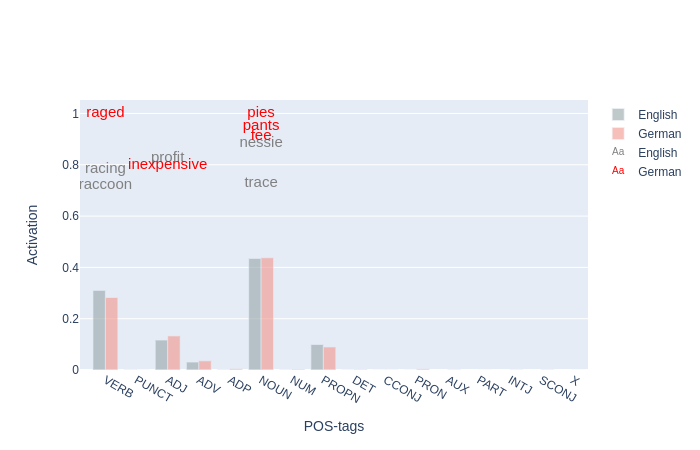}
    \caption{\textbf{Neuron-65: English (En) to German (De) Transfer} - The figure shows the pos-token-distribution for neuron:65 with the least change in neuron-knowledge between English (Grey color) to German (red color) transfer.}\label{fig:en2de_least_65}
\end{figure}


\begin{table}[h!]
  \begin{center}
    \caption{Change in neuron-knowledge for neuron-65, after English to German transfer}
    \label{tab:en2de_65}
    \begin{tabular}{l|c|c|r} 
     & \textbf{Input-} & \textbf{POS-} & \textbf{Mass-} \\ 
     & \textbf{features} & \textbf{class} & \textbf{activation} \\ 
      \hline
     English & 1901 & 11 & -3465.3454 \\  
     German & 8979 & 16 & 3854.2759 \\ 
    \end{tabular}
  \end{center}
\end{table}

Figure \ref{fig:en2de_least_65} shows the pos-token-distribution for neurons-65 that changed the least, during English to German transfer. Once again, we see that the neuron-knowledge from English language(racing, racoon, profit, trace, and nessie) in the neuron-65, got replaced with the new German knowledge (raged, inexpensive, pies, pants, and fee). Table \ref{tab:en2de_65}, on the other hand, shows the change in neuron-knowledge of neuron-65 for English to German transfer, in terms of activated input-features, POS-classes, and mass-activation. As can be seen, there is a massive increase in the number of activated input-features and POS-class after the transfer, which suggests an increase in generalized knowledge for neuron-65. The enormous increase in the mass-activation of the neuron implies an increase in knowledge-transfer for neuron-65. \newline

\subsection{De-Fr Transfer}
\begin{figure}[h]
\centering
\includegraphics[width=\textwidth, keepaspectratio]{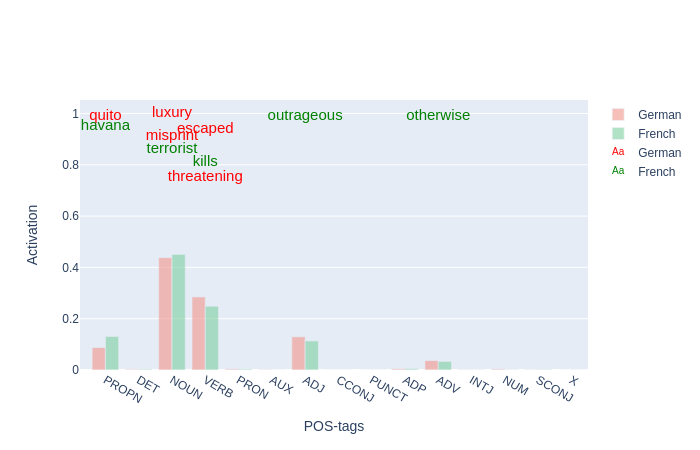}
    \caption{\textbf{Neuron-135: German (De) to French (Fr) Transfer} - The figure shows the pos-token-distribution for neuron:65 with the most change in neuron-knowledge between German (Red color) to French (Green color) transfer.}\label{fig:de2fr_most_135}
\end{figure}


\begin{table}[h!]
  \begin{center}
    \caption{Change in neuron-knowledge for neuron-135, after German to French transfer}
    \label{tab:de2fr_135}
    \begin{tabular}{l|c|c|r} 
     & \textbf{Input-} & \textbf{POS-} & \textbf{Mass-} \\ 
     & \textbf{features} & \textbf{class} & \textbf{activation} \\ 
      \hline
     German & 5240 & 15 & 588.2327 \\  
     French & 4317 & 15 & 0.04956 \\ 
    \end{tabular}
  \end{center}
\end{table}

Figure \ref{fig:de2fr_most_135} shows the pos-token-distribution for neurons-135 that changed the most, during German to French transfer. We see that the neuron-knowledge from the German language(qutio, luxury, misprint, escaped, and threatening) in the neuron-135, is replaced with the new French knowledge (havana, terrorit, kills, outrage, and otherwise). It is interesting to see how network by itself is collecting input-features related to crime, in a single neuron. Table \ref{tab:de2fr_135} shows the change in neuron-knowledge of neuron-135 for German to French transfer, in terms of activated input-features, POS-classes, and mass-activation. As can be seen, there is a decrease in the number of activated input-features after the transfer. However, the number of activated POS-class remains the same. This indicates an increase in specialized knowledge for neuron-135. Nevertheless, the massive decrease in the mass-activation indicates a decrease in knowledge-transfer for neuron-135.\newline

\begin{figure}[H]
\centering
\includegraphics[width=\textwidth, keepaspectratio]{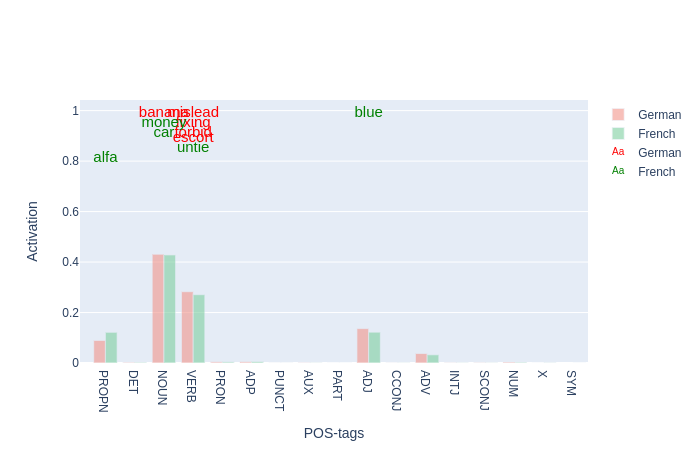}
    \caption{\textbf{Neuron-449: German (De) to French (Fr) Transfer} - The figure shows the pos-token-distribution for neuron:449 with the least change in neuron-knowledge between German (Red color) to French (Green color) transfer.}\label{fig:de2fr_least_449}
\end{figure}


\begin{table}[h!]
  \begin{center}
    \caption{Change in neuron-knowledge for neuron-449, after German to French transfer}
    \label{tab:de2fr_449}
    \begin{tabular}{l|c|c|r} 
     & \textbf{Input-} & \textbf{POS-} & \textbf{Mass-} \\ 
     & \textbf{features} & \textbf{class} & \textbf{activation} \\ 
      \hline
     German & 8337 & 16 & 3965.5332 \\  
     French & 6950 & 17 & 2287.1006 \\ 
    \end{tabular}
  \end{center}
\end{table}

Similarly, figure \ref{fig:de2fr_least_449} shows the pos-token-distribution for neurons-449 that changed the least, during German to French transfer. We see that the neuron-knowledge from the German language(banana, mislead, fixing, forbid, and escort) in the neuron-449, is replaced with the new French knowledge (alfa, money, car, untie, and blue). Table \ref{tab:de2fr_449} shows the change in neuron-knowledge of neuron-135 for German to French transfer, in terms of activated input-features, POS-classes, and mass-activation. As can be seen, there is a decrease in the number of activated input-features after the transfer. However, there is an increase in the number of activated POS-classes. This indicates an increase in generalized knowledge for neuron-449. Additionally, the decrease in the mass-activation indicates a decrease in knowledge-transfer for neuron-449.

\subsection{Fr-Es Transfer}

\begin{figure}[H]
\centering
\includegraphics[width=\textwidth, keepaspectratio]{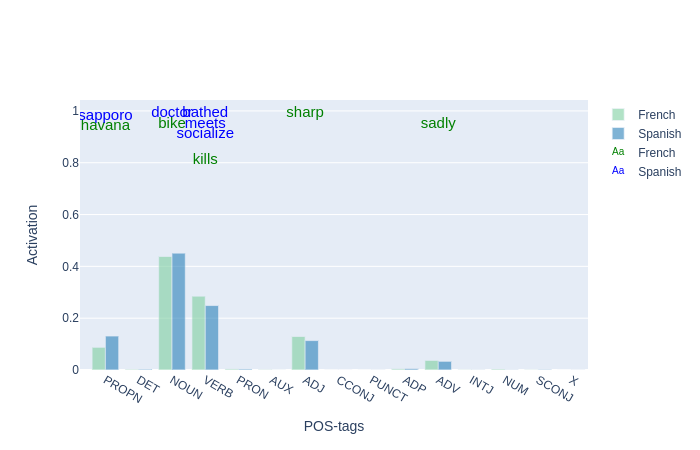}
    \caption{\textbf{Neuron-135: French (Fr) to Spanish (Es) Transfer} - The figure shows the pos-token-distribution for neuron:135 with the most change in neuron-knowledge between French (Green color) to Spanish (Blue color) transfer.}\label{fig:fr2es_most_135}
\end{figure}

Figure \ref{fig:fr2es_most_135} shows the pos-token-distribution for neurons-135 that changed the most, during French to Spanish transfer. Once again, we see that the neuron-knowledge from the French language(havana, bike, kills, sharp, and sadly) in the neuron-135, is replaced with the new Spanish knowledge (sapporo, doctor, bathed, meet, and socialize). Table \ref{tab:fr2es_135} shows the change in neuron-knowledge of neuron-135 for French to Spanish transfer, in terms of activated input-features, POS-classes, and mass-activation. As can be seen, there is a decrease in the number of activated input-features after the transfer. However, the number of activated POS-class remains the same. This indicates an increase in specialized knowledge for neuron-135. The decrease in mass-activation indicates a decrease in knowledge-transfer for neuron-135. \newline


\begin{table}[h!]
  \begin{center}
    \caption{Change in neuron-knowledge for neuron-135, after French to Spanish transfer}
    \label{tab:fr2es_135}
    \begin{tabular}{l|c|c|r} 
     & \textbf{Input-} & \textbf{POS-} & \textbf{Mass-} \\ 
     & \textbf{features} & \textbf{class} & \textbf{activation} \\ 
      \hline
     French & 4540 & 15 & 272.6493 \\  
     Spanish & 4317 & 15 & 0.0495 \\ 
    \end{tabular}
  \end{center}
\end{table}

\begin{figure}[H]
\centering
\includegraphics[width=\textwidth]{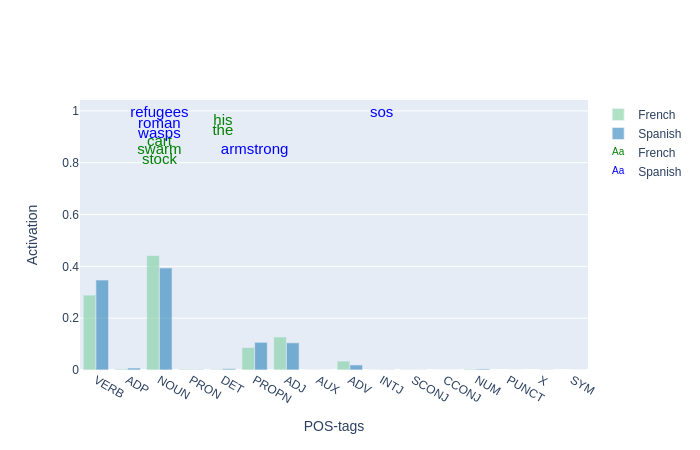}
    \caption{\textbf{Neuron-168: French (Fr) to Spanish (Es) Transfer} - The figure shows the pos-token-distribution for neuron:168 with the least change in neuron-knowledge between French (Green color) to Spanish (Blue color) transfer.}\label{fig:fr2es_least_168}
\end{figure}

Similarly, figure \ref{fig:fr2es_least_168} shows the pos-token-distribution for neurons-168 that changed the least, during French to Spanish transfer. Once again, we see that the neuron-knowledge from the French language(cart, swarm, stock, he, and the) in the neuron-168, is replaced with the new Spanish knowledge (refugees, roman, wasps, armstrong, and sos). Table \ref{tab:fr2es_168} shows the change in neuron-knowledge of neuron-168 for French to Spanish transfer, in terms of activated input-features, POS-classes, and mass-activation. As can be seen, there is a decrease in the number of activated input-features. However, the increase in the number of activated POS-classes indicates an increase in generalized knowledge. Once again, there is a massive drop in the mass-activation, which suggests a decrease in knowledge-transfer and an increase in catastrophic forgetting. \footnote{Note that we only comment about catastrophic forgetting when there a drastic change in mass-activation and the number of input-features. We argue that the drop in mass activation for other cases, is a mere result of knowledge-shift from one neuron to another.} \newline


\begin{table}[h!]
  \begin{center}
    \caption{Change in neuron-knowledge for neuron-168, after French to Spanish transfer}
    \label{tab:fr2es_168}
    \begin{tabular}{l|c|c|r} 
     & \textbf{Input-} & \textbf{POS-} & \textbf{Mass-} \\ 
     & \textbf{features} & \textbf{class} & \textbf{activation} \\ 
      \hline
     French & 7547 & 15 & 2893.7721 \\  
     Spanish & 1934 & 16 & -2198.6153 \\ 
    \end{tabular}
  \end{center}
\end{table}
 

\chapter{Conclusion} 

\label{Chapter7} 
Through this research, we investigated questions such as :

\begin{itemize}
    \item \textbf{RQ1: How does knowledge-transfer affect Multi-lingual Neural Machine Translation, in an extremely low-resource setting?} \newline

Through this research question, we explored how different languages improves knowledge-transfer in a Multi-lingual Neural Machine Translation system. We discovered that while sequentially training on languages from different roots decreases the knowledge-transfer performance, languages from similar roots improves it. Additionally, we also show how multi-task and sequential-transfer affect the model's performance in such an extremely low-resource setting. We discovered that freezing weights at every step of the training not only helps against catastrophic forgetting but also helps in improving knowledge-transfer over time. Not only our sequential network outperforms the end-to-end baselines, but it also shows improvement over the direct 1-hop transfer.  

    \item \textbf{RQ2: How does selectively pruning neuron-knowledge affects the model's generalization, robustness, and catastrophic forgetting?}
    
Through this research question, we explored the effects of pruning dead neurons from the max-activation technique. We discovered that unlike the binary text-classification problem encountered in [\textbf{\cite{rethmeier2019txray}}], Neural Machine Translation is an N-class classification problem. By forcing our network to generate sparse connections, we only pruned the sequential information that was needed for a better transfer. In other words, pruning dead-neurons based on max mass-activation technique does not apply to an N-class classification problem. \newline

Additionally, we also investigate the impact of pruning neurons with most-n and least-n neuron-knowledge. We discovered that while pruning neuron-knowledge from the most and least activated neurons initially increases the robustness and generalization of the system, continuously pruning eventually destroys the system performance. Therefore, we conclude that sequentially pruning in such an extremely low-resource setting environment, only erases the essential information needed for a better transfer.

\item \textbf{RQ3: How can we use interpretability as a tool to understand knowledge abstractions and knowledge-transfer in a trained model?}
    
Through this research question, we produced various visualizations techniques to understand the effects of positive and negative-activations (knowledge) on the system's performance. We also examined the effects of different pruning techniques on these positive and negative-activations. Considering these activations as the knowledge-abstractions, we found that the system's performance decreases whenever negative-knowledge increases more than the positive-knowledge. These knowledge-representations also helped us to understand the increase or decrease in knowledge-transfer at every step of our sequential training. Additionally, we also generate pos-token-distributions to visualize the knowledge-content in each of the trained-neurons. Not only these pos-token-distributions helped us to understand the impact of knowledge-transfer on neuron-knowledge, but they also allowed us to identify whether a neuron got specialized or generalized after the application of transfer. We argue that together, these visualizations provide a gateway to understand how knowledge-transfer changes model-learning and neuron-distribution at different stages of our sequential training.  
\end{itemize}


\chapter{Future Research} 

\label{Chapter8} 
While pruning neuron-knowledge did not improve the performance of Multi-lingual Neural Machine Translation in our extremely low-resource setting, we argue that the effects of catastrophic forgetting would be less severe for a multi-layered neural network. Therefore, in future, we would like to investigate the effects of different pruning strategies on a deeper neural network, with more hidden units and larger embedding size. Additionally, instead of training our pre-trained model on Tatoeba Dataset [\textbf{\cite{TiedemannN04}}], we would like to pre-train our English-English translation model on Wikitext-103 English sentences [\textbf{\cite{wikitext}}]. This will increase the understanding of source language in our model. While we conclude that the performance of our system improves as we sequentially train on different languages, it will also be interesting to observe the effects of knowledge-transfer on Arabic and Sanskrit based languages, i.e. with entirely different language-roots.  \newline 

Additionally, we would also like to examine the effects of noise on our model's performance. In future, we plan to introduce the noise through the embeddings and test the robustness of our Multi-lingual Neural Machine Translation system. Similar to our pruning experiments, we can also identify neurons with most and least noisy neuron-knowledge. It would be interesting to investigate how does pruning these noisy neurons affects catastrophic forgetting and robustness of our system. \newline

While TX-Ray is capable of explaining knowledge-transfer, it is computationally expensive. Unlike many conventional XAI methods, where the interpretations and explanations can be obtained simultaneously to the network training, our approach is a post-hoc technique. This implies that our analysis can only be utilised once the models are trained. Depending upon the complexity of a network, generating these knowledge-representations from the trained model is not only computationally expensive and time-consuming but also requires substantial memory-resources.  In future, we would like to focus on XAI techniques that can be applied to heavier architectures, on the get-go.  Lastly, conditioning on POS-features restricts our method by capturing only low-level features. In future, we would like to represent knowledge-transfer both on POS(low) and subject-object(high) level-features, using Spacy's POS-tagger and dependency parser.


\appendix 



\chapter{Appendix} 

\section{Evolution of Positive and Negative knowledge-abstractions during training}

\label{AppendixA} 
\begin{figure}[h]
\centering
\includegraphics[width=\linewidth, height=12cm]{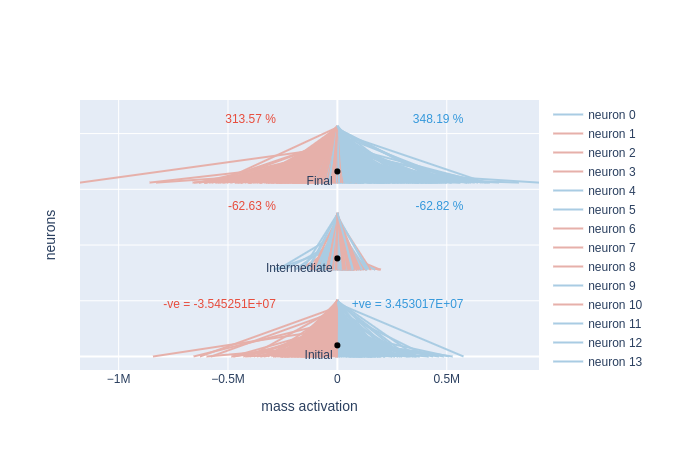}
\caption{\textbf{Positive and Negative knowledge-abstractions for English to English translations in end-to-end network} : Evolution of Positive (blue) and Negative (red) knowledge-abstractions, for English to English translations from first to final epoch, in end-to-end network. }\label{fig:en2en_end2end}
\end{figure}

\begin{figure}[h]
\centering
\includegraphics[width=\linewidth, height=12cm]{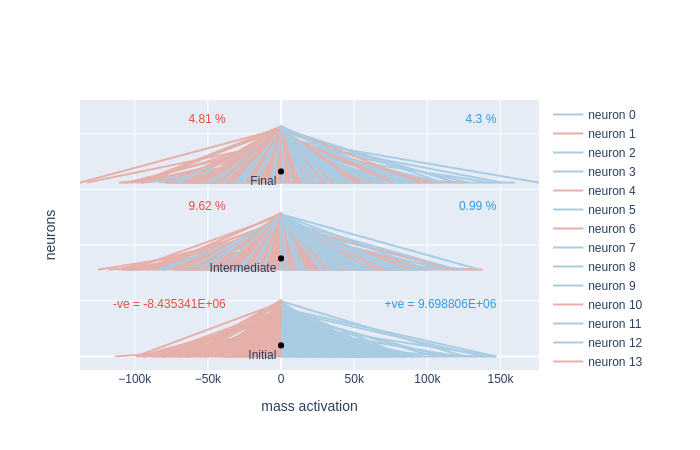}
\caption{\textbf{Positive and Negative knowledge-abstractions for English to German translations in end-to-end network} : Evolution of Positive (blue) and Negative (red) knowledge-abstractions, for English to German translations from first to final epoch, in end-to-end network. }\label{fig:en2de_end2end}
\end{figure}

\begin{figure}[h]
\centering
\includegraphics[width=\linewidth, height=12cm]{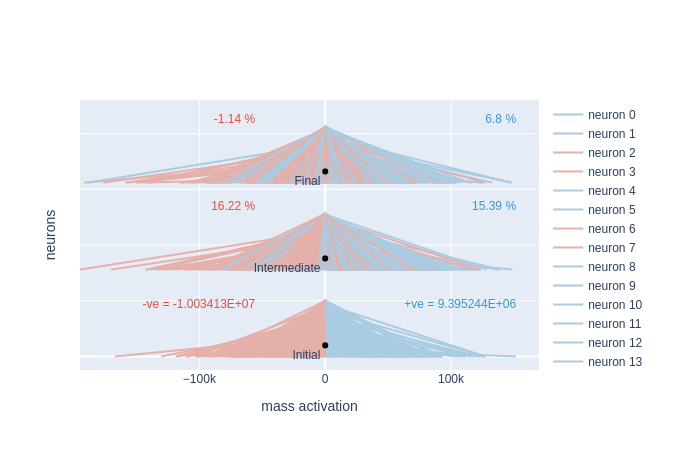}
\caption{\textbf{Positive and Negative knowledge-abstractions for English to French translations in end-to-end network} : Evolution of Positive (blue) and Negative (red) knowledge-abstractions, for English to French translations from first to final epoch, in end-to-end network. }\label{fig:en2fr_end2end}
\end{figure}

\begin{figure}[h]
\centering
\includegraphics[width=\linewidth, height=12cm]{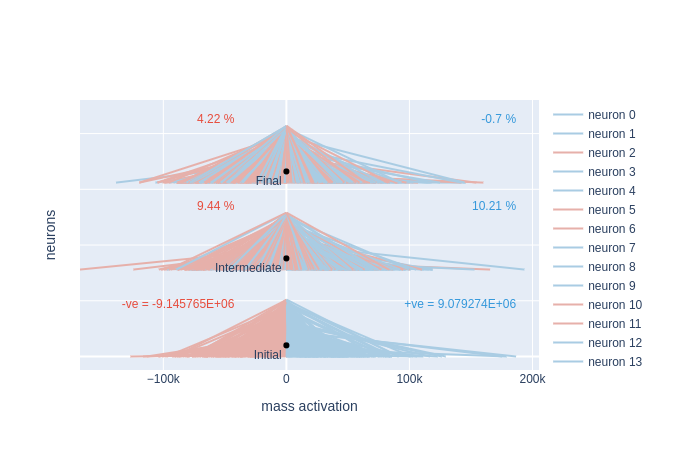}
\caption{\textbf{Positive and Negative knowledge-abstractions for English to Spanish translations in end-to-end network} : Evolution of Positive (blue) and Negative (red) knowledge-abstractions, for English to Spanish translations from first to final epoch, in end-to-end network. }\label{fig:en2es_end2end}
\end{figure}

\begin{figure}[h]
\centering
\includegraphics[width=\linewidth, height=12cm]{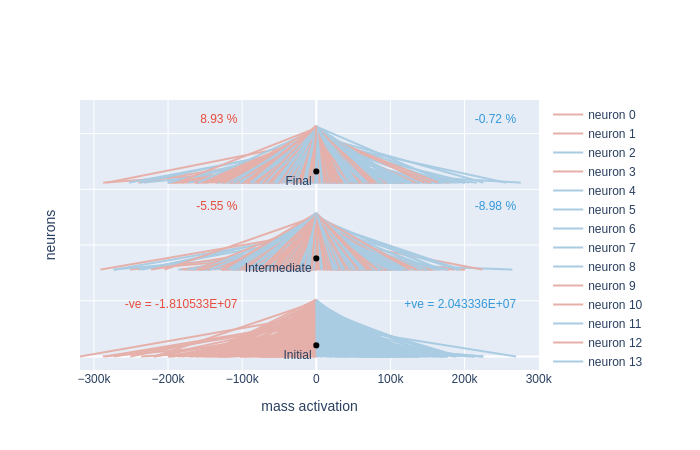}
\caption{\textbf{Positive and Negative knowledge-abstractions for English to German translations in multi-task transfer network} : Evolution of Positive (blue) and Negative (red) knowledge-abstractions, for English to German translations from first to final epoch, in multi-task transfer network. }\label{fig:en2de_multi}
\end{figure}

\begin{figure}[h]
\centering
\includegraphics[width=\linewidth, height=12cm]{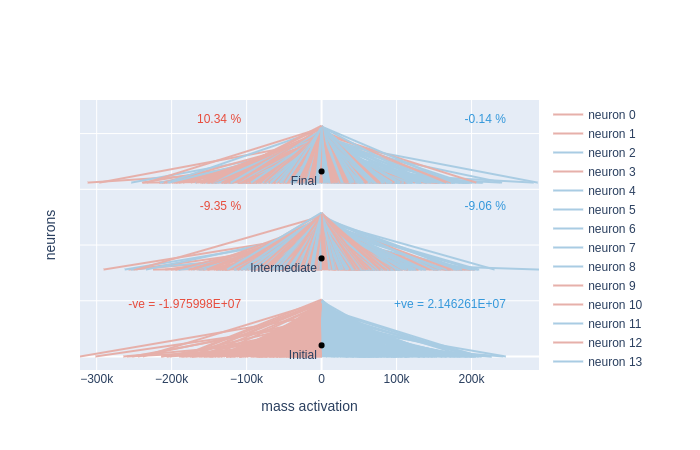}
\caption{\textbf{Positive and Negative knowledge-abstractions for English to French translations in multi-task transfer network} : Evolution of Positive (blue) and Negative (red) knowledge-abstractions, for English to French translations from first to final epoch, in multi-task transfer network. }\label{fig:en2fr_multi}
\end{figure}

\begin{figure}[h]
\centering
\includegraphics[width=\linewidth, height=12cm]{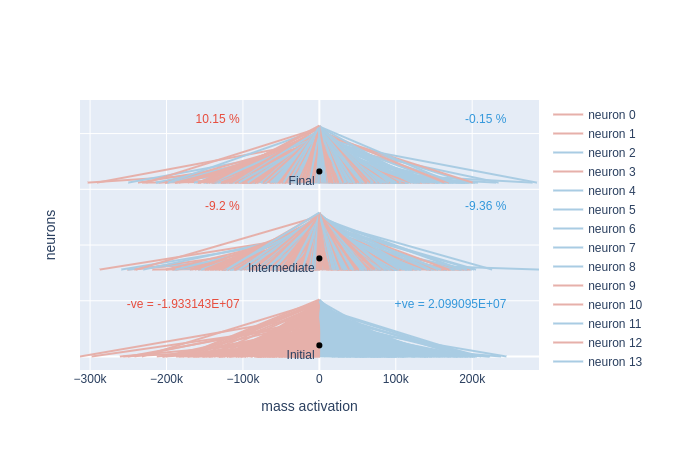}
\caption{\textbf{Positive and Negative knowledge-abstractions for English to Spanish translations in multi-task transfer network} : Evolution of Positive (blue) and Negative (red) knowledge-abstractions, for English to Spanish translations from first to final epoch, in sequential transfer network. }\label{fig:en2es_multi}
\end{figure}

\begin{figure}[h]
\centering
\includegraphics[width=\linewidth, height=12cm]{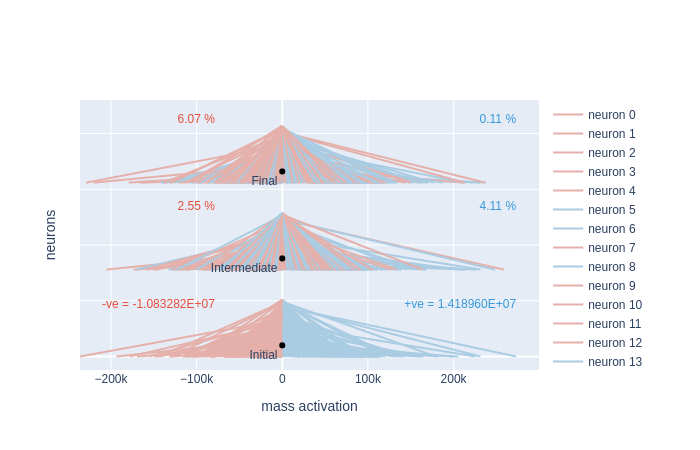}
\caption{\textbf{Positive and Negative knowledge-abstractions for English to German translations in sequential transfer network} : Evolution of Positive (blue) and Negative (red) knowledge-abstractions, for English to German translations from first to final epoch, in sequential transfer network. }\label{fig:en2de_sequential}
\end{figure}

\begin{figure}[h]
\centering
\includegraphics[width=\linewidth, height=12cm]{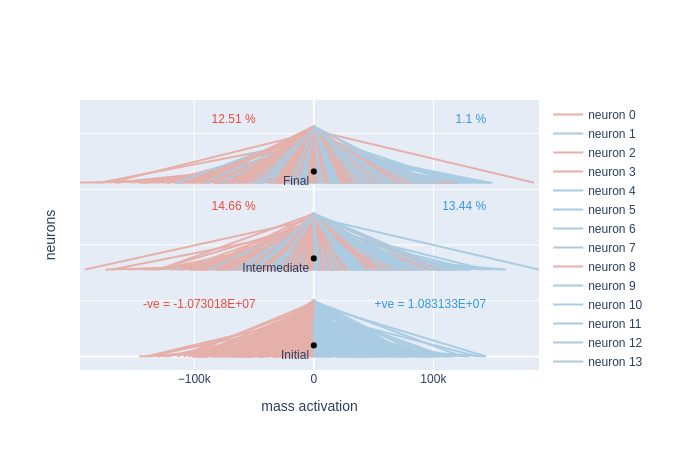}
\caption{\textbf{Positive and Negative knowledge-abstractions for English to French translations in sequential transfer network} : Evolution of Positive (blue) and Negative (red) knowledge-abstractions, for English to French translations from first to final epoch, in sequential transfer network. }\label{fig:en2fr_sequential}
\end{figure}

\begin{figure}[h]
\centering
\includegraphics[width=\linewidth, height=12cm]{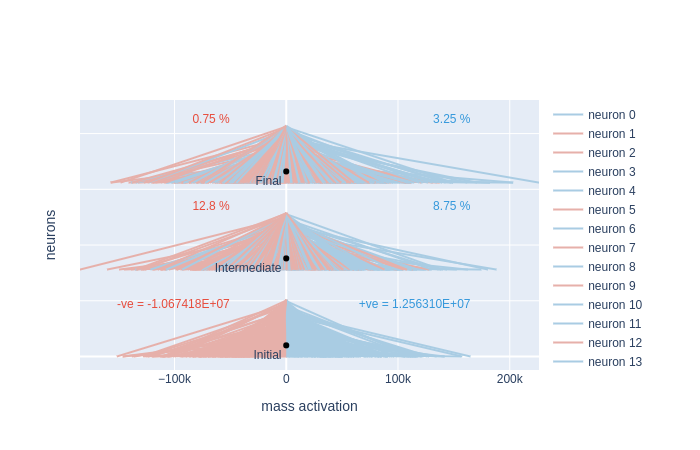}
\caption{\textbf{Positive and Negative knowledge-abstractions for English to Spanish translations in sequential transfer network} : Evolution of Positive (blue) and Negative (red) knowledge-abstractions, for English to Spanish translations from first to final epoch, in sequential transfer network. }\label{fig:en2es_sequential}
\end{figure}


\printbibliography

@misc{clinchant2019use,
    title={On the use of BERT for Neural Machine Translation},
    author={Stéphane Clinchant and Kweon Woo Jung and Vassilina Nikoulina},
    year={2019},
    eprint={1909.12744},
    archivePrefix={arXiv},
    primaryClass={cs.CL}
}

@misc{sutskever2014sequence,
    title={Sequence to Sequence Learning with Neural Networks},
    author={Ilya Sutskever and Oriol Vinyals and Quoc V. Le},
    year={2014},
    eprint={1409.3215},
    archivePrefix={arXiv},
    primaryClass={cs.CL}
}

@misc{zhu2020incorporating,
    title={Incorporating BERT into Neural Machine Translation},
    author={Jinhua Zhu and Yingce Xia and Lijun Wu and Di He and Tao Qin and Wengang Zhou and Houqiang Li and Tie-Yan Liu},
    year={2020},
    eprint={2002.06823},
    archivePrefix={arXiv},
    primaryClass={cs.CL}
}

@misc{klein2017opennmt,
    title={OpenNMT: Open-Source Toolkit for Neural Machine Translation},
    author={Guillaume Klein and Yoon Kim and Yuntian Deng and Jean Senellart and Alexander M. Rush},
    year={2017},
    eprint={1701.02810},
    archivePrefix={arXiv},
    primaryClass={cs.CL}
}

@misc{rethmeier2019txray,
    title={TX-Ray: Quantifying and Explaining Model-Knowledge Transfer in (Un-)Supervised NLP},
    author={Nils Rethmeier and Vageesh Kumar Saxena and Isabelle Augenstein},
    year={2019},
    eprint={1912.00982},
    archivePrefix={arXiv},
    primaryClass={cs.LG}
}

@unpublished{spacy2,
    AUTHOR = {Honnibal, Matthew and Montani, Ines},
    TITLE  = {{spaCy 2}: Natural language understanding with {B}loom embeddings, convolutional neural networks and incremental parsing},
    YEAR   = {2017},
    Note   = {To appear}
}

@misc{aharoni2019massively,
    title={Massively Multilingual Neural Machine Translation},
    author={Roee Aharoni and Melvin Johnson and Orhan Firat},
    year={2019},
    eprint={1903.00089},
    archivePrefix={arXiv},
    primaryClass={cs.CL}
}

@misc{frankle2018lottery,
    title={The Lottery Ticket Hypothesis: Finding Sparse, Trainable Neural Networks},
    author={Jonathan Frankle and Michael Carbin},
    year={2018},
    eprint={1803.03635},
    archivePrefix={arXiv},
    primaryClass={cs.LG}
}

@misc{lakew2019multilingual,
    title={Multilingual Neural Machine Translation for Zero-Resource Languages},
    author={Surafel M. Lakew and Marcello Federico and Matteo Negri and Marco Turchi},
    year={2019},
    eprint={1909.07342},
    archivePrefix={arXiv},
    primaryClass={cs.CL}
}

@misc{gu2018universal,
    title={Universal Neural Machine Translation for Extremely Low Resource Languages},
    author={Jiatao Gu and Hany Hassan and Jacob Devlin and Victor O. K. Li},
    year={2018},
    eprint={1802.05368},
    archivePrefix={arXiv},
    primaryClass={cs.CL}
}

@article{johnson-etal-2017-googles,
    title = "{G}oogle{'}s Multilingual Neural Machine Translation System: Enabling Zero-Shot Translation",
    author = "Johnson, Melvin  and
      Schuster, Mike  and
      Le, Quoc V.  and
      Krikun, Maxim  and
      Wu, Yonghui  and
      Chen, Zhifeng  and
      Thorat, Nikhil  and
      Vi{\'e}gas, Fernanda  and
      Wattenberg, Martin  and
      Corrado, Greg  and
      Hughes, Macduff  and
      Dean, Jeffrey",
    journal = "Transactions of the Association for Computational Linguistics",
    volume = "5",
    year = "2017",
    url = "https://www.aclweb.org/anthology/Q17-1024",
    doi = "10.1162/tacl_a_00065",
    pages = "339--351",
}

@article{Dwarampudi2019EffectsOP,
  title={Effects of padding on LSTMs and CNNs},
  author={Mahidhar Dwarampudi and N. V. Subba Reddy},
  journal={ArXiv},
  year={2019},
  volume={abs/1903.07288}
}

@article{johnson2017google,
  title={Google’s multilingual neural machine translation system: Enabling zero-shot translation},
  author={Johnson, Melvin and Schuster, Mike and Le, Quoc V and Krikun, Maxim and Wu, Yonghui and Chen, Zhifeng and Thorat, Nikhil and Vi{\'e}gas, Fernanda and Wattenberg, Martin and Corrado, Greg and others},
  journal={Transactions of the Association for Computational Linguistics},
  volume={5},
  pages={339--351},
  year={2017},
  publisher={MIT Press}
}

@misc{nguyen2020dissecting,
    title={Dissecting Catastrophic Forgetting in Continual Learning by Deep Visualization},
    author={Giang Nguyen and Shuan Chen and Thao Do and Tae Joon Jun and Ho-Jin Choi and Daeyoung Kim},
    year={2020},
    eprint={2001.01578},
    archivePrefix={arXiv},
    primaryClass={cs.LG}
}

@inproceedings{lee2017overcoming,
  title={Overcoming catastrophic forgetting by incremental moment matching},
  author={Lee, Sang-Woo and Kim, Jin-Hwa and Jun, Jaehyun and Ha, Jung-Woo and Zhang, Byoung-Tak},
  booktitle={Advances in neural information processing systems},
  pages={4652--4662},
  year={2017}
}

@misc{cho2014learning,
    title={Learning Phrase Representations using RNN Encoder-Decoder for Statistical Machine Translation},
    author={Kyunghyun Cho and Bart van Merrienboer and Caglar Gulcehre and Dzmitry Bahdanau and Fethi Bougares and Holger Schwenk and Yoshua Bengio},
    year={2014},
    eprint={1406.1078},
    archivePrefix={arXiv},
    primaryClass={cs.CL}
}

@misc{bahdanau2014neural,
    title={Neural Machine Translation by Jointly Learning to Align and Translate},
    author={Dzmitry Bahdanau and Kyunghyun Cho and Yoshua Bengio},
    year={2014},
    eprint={1409.0473},
    archivePrefix={arXiv},
    primaryClass={cs.CL}
}

@misc{chung2014empirical,
    title={Empirical Evaluation of Gated Recurrent Neural Networks on Sequence Modeling},
    author={Junyoung Chung and Caglar Gulcehre and KyungHyun Cho and Yoshua Bengio},
    year={2014},
    eprint={1412.3555},
    archivePrefix={arXiv},
    primaryClass={cs.NE}
}

@inproceedings{TiedemannN04,
  author = {Tiedemann, Jörg and Nygaard, Lars},
  booktitle = {LREC},
  crossref = {conf/lrec/2004},
  ee = {http://www.lrec-conf.org/proceedings/lrec2004/summaries/320.htm},
  publisher = {European Language Resources Association},
  title = {The OPUS Corpus - Parallel and Free: http: //logos.uio.no/opus.},
  url = {http://dblp.uni-trier.de/db/conf/lrec/lrec2004.html#TiedemannN04},
  year = 2004
}

@inproceedings{TiedemannThottingal:2020,
  title        = {{OPUS-MT} – {B}uilding open translation services for the World},
  author       = {Tiedemann, J{\"o}rg and Thottingal, Santhosh},
  booktitle    = {Proceedings of the 22nd Annual Conference of the European Association for Machine Translation (EAMT)},
  year         = {2020},
  isbn         = {978-989-33-0589-8},
  url          = {http://eamt2020.inesc-id.pt/proceedings-eamt2020.pdf},
  organization = {European Association for Machine Translation}
}

@article{golkar2019continual,
  title={Continual learning via neural pruning},
  author={Golkar, Siavash and Kagan, Michael and Cho, Kyunghyun},
  journal={arXiv preprint arXiv:1903.04476},
  year={2019}
}

@article{wikitext,
title= {Wikitext-103},
journal= {},
author= {Stephen Merity, 2016},
year= {2016},
url= {https://arxiv.org/abs/1609.07843},
license= {},
superseded= {},
terms= {}
}

@article{soseliafreezing,
  title={Freezing Networks: Weight Preservation Procedure for Continual Learning},
  author={Soselia, Davit and Shugliashvili, Levan and Amashukeli, Shota and Koberidze, Irakli and Chelidze, Giorgi},
  year={2018}
}

@article{kirkpatrick2017overcoming,
  title={Overcoming catastrophic forgetting in neural networks},
  author={Kirkpatrick, James and Pascanu, Razvan and Rabinowitz, Neil and Veness, Joel and Desjardins, Guillaume and Rusu, Andrei A and Milan, Kieran and Quan, John and Ramalho, Tiago and Grabska-Barwinska, Agnieszka and others},
  journal={Proceedings of the national academy of sciences},
  volume={114},
  number={13},
  pages={3521--3526},
  year={2017},
  publisher={National Acad Sciences}
}

@phdthesis{ruder2019neural,
  title={Neural transfer learning for natural language processing},
  author={Ruder, Sebastian},
  year={2019},
  school={NUI Galway}
}

@article{carter2019activation,
  title={Activation atlas},
  author={Carter, Shan and Armstrong, Zan and Schubert, Ludwig and Johnson, Ian and Olah, Chris},
  journal={Distill},
  volume={4},
  number={3},
  pages={e15},
  year={2019}
}

@article{journals/corr/YosinskiCNFL15,
  author = {Yosinski, Jason and Clune, Jeff and Nguyen, Anh Mai and Fuchs, Thomas J. and Lipson, Hod},
  ee = {http://arxiv.org/abs/1506.06579},
  journal = {CoRR},
  title = {Understanding Neural Networks Through Deep Visualization.},
  url = {http://dblp.uni-trier.de/db/journals/corr/corr1506.html#YosinskiCNFL15},
  volume = {abs/1506.06579},
  year = 2015
}

@article{hohman2019summit,
  title={Summit: Scaling Deep Learning Interpretability by Visualizing Activation and Attribution Summarizations},
  author={Hohman, Fred and Park, Haekyu and Robinson, Caleb and Chau, Duen Horng},
  journal={IEEE Transactions on Visualization and Computer Graphics (TVCG)},
  year={2020},
  publisher={IEEE},
  url={https://fredhohman.com/papers/19-summit-vast.pdf}
}

@article{belinkov-glass-2019-analysis,
    title = "Analysis Methods in Neural Language Processing: A Survey",
    author = "Belinkov, Yonatan  and
      Glass, James",
    journal = "Transactions of the Association for Computational Linguistics",
    volume = "7",
    month = apr,
    year = "2019",
    url = "https://www.aclweb.org/anthology/Q19-1004",
    doi = "10.1162/tacl_a_00254",
    pages = "49--72",
}

@inproceedings{papineni2002bleu,
  title={BLEU: a method for automatic evaluation of machine translation},
  author={Papineni, Kishore and Roukos, Salim and Ward, Todd and Zhu, Wei-Jing},
  booktitle={Proceedings of the 40th annual meeting of the Association for Computational Linguistics},
  pages={311--318},
  year={2002}
}

@article{olah2018the,
  author = {Olah, Chris and Satyanarayan, Arvind and Johnson, Ian and Carter, Shan and Schubert, Ludwig and Ye, Katherine and Mordvintsev, Alexander},
  title = {The Building Blocks of Interpretability},
  journal = {Distill},
  year = {2018},
  url = {https://distill.pub/2018/building-blocks},
  doi = {10.23915/distill.00010}
}

@inproceedings{arras-etal-2017-explaining,
    title = "Explaining Recurrent Neural Network Predictions in Sentiment Analysis",
    author = {Arras, Leila  and
      Montavon, Gr{\'e}goire  and
      M{\"u}ller, Klaus-Robert  and
      Samek, Wojciech},
    booktitle = "Proceedings of the 8th Workshop on Computational Approaches to Subjectivity, Sentiment and Social Media Analysis",
    month = sep,
    year = "2017",
    address = "Copenhagen, Denmark",
    publisher = "Association for Computational Linguistics",
    url = "https://www.aclweb.org/anthology/W17-5221",
    doi = "10.18653/v1/W17-5221",
    pages = "159--168",
}

@inproceedings{Senteval,
  author    = {Alexis Conneau and
               Douwe Kiela},
  title     = {SentEval: An Evaluation Toolkit for Universal Sentence Representations},
  booktitle = {Proceedings of the Eleventh International Conference on Language Resources
               and Evaluation, {LREC} 2018, Miyazaki, Japan, May 7-12, 2018.},
  year      = {2018},
  url       = {http://www.lrec-conf.org/proceedings/lrec2018/summaries/757.html},
  bibsource = {dblp computer science bibliography, https://dblp.org}
}

@inproceedings{Glue,
  author    = {Alex Wang and
               Amanpreet Singh and
               Julian Michael and
               Felix Hill and
               Omer Levy and
               Samuel R. Bowman},
  title     = {{GLUE:} {A} Multi-Task Benchmark and Analysis Platform for Natural
               Language Understanding},
  booktitle = {7th International Conference on Learning Representations, {ICLR} 2019,
               New Orleans, LA, USA, May 6-9, 2019},
  year      = {2019},
  url       = {https://openreview.net/forum?id=rJ4km2R5t7},
  bibsource = {dblp computer science bibliography, https://dblp.org}
}

@inproceedings{2019-errudite,
    title = "{E}rrudite: Scalable, Reproducible, and Testable Error Analysis",
    author = "Wu, Tongshuang  and
      Ribeiro, Marco Tulio  and
      Heer, Jeffrey  and
      Weld, Daniel",
    booktitle = "Proceedings of the 57th Annual Meeting of the Association for Computational Linguistics",
    month = jul,
    year = "2019",
    address = "Florence, Italy",
    publisher = "Association for Computational Linguistics",
    url = "https://www.aclweb.org/anthology/P19-1073",
    pages = "747--763",
}

@article{DeepEyes18,
  author    = {Nicola Pezzotti and
               Thomas H{\"{o}}llt and
               Jan C. van Gemert and
               Boudewijn P. F. Lelieveldt and
               Elmar Eisemann and
               Anna Vilanova},
  title     = {DeepEyes: Progressive Visual Analytics for Designing Deep Neural Networks},
  journal   = {{IEEE} Transactions on Visualization and Computer Graphics},
  volume    = {24},
  number    = {1},
  pages     = {98--108},
  year      = {2018},
  url       = {https://doi.org/10.1109/TVCG.2017.2744358},
  doi       = {10.1109/TVCG.2017.2744358},
  bibsource = {dblp computer science bibliography, https://dblp.org}
}

@ARTICLE{CSI19,
author={Sebastian Gehrmann and
              Hendrik Strobelt and
              Robert Kr{\"{u}}ger and
              Hanspeter Pfister and
              Alexander M. Rush},
journal={IEEE Transactions on Visualization and Computer Graphics},
title={Visual Interaction with Deep Learning Models through Collaborative Semantic Inference},
year={2019},
volume={},
number={},
pages={1-1},
url= {http://arxiv.org/abs/1907.10739},
doi={10.1109/TVCG.2019.2934595},
ISSN={},
month={},
}

@inproceedings{arrasACL19,
    title = "Evaluating Recurrent Neural Network Explanations",
    author = {Arras, Leila  and
      Osman, Ahmed  and
      M{\"u}ller, Klaus-Robert  and
      Samek, Wojciech},
    booktitle = "Proceedings of the 2019 ACL Workshop BlackboxNLP: Analyzing and Interpreting Neural Networks for NLP",
    month = aug,
    year = "2019",
    address = "Florence, Italy",
    publisher = "Association for Computational Linguistics",
    url = "https://www.aclweb.org/anthology/W19-4813",
    pages = "113--126",
}

@article{Decathlon,
  author    = {Bryan McCann and
               Nitish Shirish Keskar and
               Caiming Xiong and
               Richard Socher},
  title     = {The Natural Language Decathlon: Multitask Learning as Question Answering},
  journal   = {CoRR},
  volume    = {abs/1806.08730},
  year      = {2018},
  url       = {http://arxiv.org/abs/1806.08730},
  archivePrefix = {arXiv},
  eprint    = {1806.08730},
  bibsource = {dblp computer science bibliography, https://dblp.org}
}

@inproceedings{DiagnosticClassifiers,
  author    = {Mario Giulianelli and
               Jack Harding and
               Florian Mohnert and
               Dieuwke Hupkes and
               Willem H. Zuidema},
  title     = {Under the Hood: Using Diagnostic Classifiers to Investigate and Improve
               how Language Models Track Agreement Information},
  booktitle = {Proceedings of the Workshop: Analyzing and Interpreting Neural Networks
               for NLP, BlackboxNLP@EMNLP 2018, Brussels, Belgium, November 1, 2018},
  pages     = {240--248},
  year      = {2018},
  url       = {https://www.aclweb.org/anthology/W18-5426/},
  bibsource = {dblp computer science bibliography, https://dblp.org}
}

@inproceedings{ExplainingExplanations,
  author    = {Leilani H. Gilpin and
               David Bau and
               Ben Z. Yuan and
               Ayesha Bajwa and
               Michael Specter and
               Lalana Kagal},
  title     = {Explaining Explanations: An Overview of Interpretability of Machine
               Learning},
  booktitle = {5th {IEEE} International Conference on Data Science and Advanced Analytics,
               {DSAA} 2018, Turin, Italy, October 1-3, 2018},
  pages     = {80--89},
  year      = {2018},
  url       = {https://doi.org/10.1109/DSAA.2018.00018},
  doi       = {10.1109/DSAA.2018.00018},
  bibsource = {dblp computer science bibliography, https://dblp.org}
}

@article{SEQ2SEQVIS,
  author    = {Hendrik Strobelt and
               Sebastian Gehrmann and
               Michael Behrisch and
               Adam Perer and
               Hanspeter Pfister and
               Alexander M. Rush},
  title     = {Seq2seq-Vis: {A} Visual Debugging Tool for Sequence-to-Sequence Models},
  journal   = {{IEEE} Transactions on Visualization and Computer Graphics},
  volume    = {25},
  number    = {1},
  pages     = {353--363},
  year      = {2019},
  url       = {https://doi.org/10.1109/TVCG.2018.2865044},
  doi       = {10.1109/TVCG.2018.2865044},
  bibsource = {dblp computer science bibliography, https://dblp.org}
}

@article{RetainVis,
  author    = {Bum Chul Kwon and
               Min{-}Je Choi and
               Joanne Taery Kim and
               Edward Choi and
               Young Bin Kim and
               Soonwook Kwon and
               Jimeng Sun and
               Jaegul Choo},
  title     = {RetainVis: Visual Analytics with Interpretable and Interactive Recurrent
               Neural Networks on Electronic Medical Records},
  journal   = {{IEEE} Transactions on Visualization and Computer Graphics},
  volume    = {25},
  number    = {1},
  pages     = {299--309},
  year      = {2019},
  url       = {https://doi.org/10.1109/TVCG.2018.2865027},
  doi       = {10.1109/TVCG.2018.2865027},
  bibsource = {dblp computer science bibliography, https://dblp.org}
}

@book{10.5555/1207109,
author = {Priddy, Kevin L. and Keller, Paul E.},
title = {Artificial Neural Networks: An Introduction (SPIE Tutorial Texts in Optical Engineering, Vol. TT68)},
year = {2005},
isbn = {0819459879},
publisher = {SPIE- International Society for Optical Engineering}
}

@misc{nguyen2019understanding,
    title={Toward Understanding Catastrophic Forgetting in Continual Learning},
    author={Cuong V. Nguyen and Alessandro Achille and Michael Lam and Tal Hassner and Vijay Mahadevan and Stefano Soatto},
    year={2019},
    eprint={1908.01091},
    archivePrefix={arXiv},
    primaryClass={cs.LG}
}

\end{document}